%% file: black.revision01.tex
\documentclass[12pt]{elsarticle}

\usepackage{amsmath}
\usepackage{amssymb}
\usepackage{amsthm}
\usepackage{mathtools}
\usepackage{url}
\usepackage{multirow}
\usepackage[justification=centering]{caption}
\usepackage{tikz}
\usepackage{tikz-cd}
\usetikzlibrary{matrix,calc,arrows,arrows.meta}
\usepackage{arydshln}

\usepackage{xcolor}
\usepackage[normalem]{ulem}

\usepackage{rotating}

\journal{Artificial Intelligence}

\begin{document}

\input{commands}

\begin{frontmatter}



\title{Black-box Testing of First-Order Logic Ontologies \\ Using \WORDNET{}
}


\author[EHU]{Javier \'{A}lvez}
\author[EHU]{Paqui Lucio}
\author[EHU]{German Rigau}

\address[EHU]{Department of Computer Languages and Systems \\ University of the Basque Country UPV/EHU}

\begin{abstract}
Artificial Intelligence aims to provide computer programs with commonsense knowledge to reason about our world. This paper offers a new practical approach towards automated commonsense reasoning with first-order logic (FOL) ontologies.  We propose a new black-box testing methodology of FOL \SUMO{}-based ontologies by exploiting \WORDNET{} and its mapping into \SUMO{}. Our proposal includes a method for the (semi-)automatic creation of a very large benchmark of competency questions and a procedure for its automated evaluation by using automated theorem provers (ATPs). Applying different quality criteria, our testing proposal enables a successful evaluation of a) the competency of several translations of \SUMO{} into FOL and b) the performance of various automated ATPs. Finally, we also provide a fine-grained and complete analysis of the commonsense reasoning competency of current FOL \SUMO{}-based ontologies.
\end{abstract}


\end{frontmatter}


\newtheorem{lemma}{Lemma}

\newtheorem{theorem}{Theorem}

\section{Introduction} \label{section:introduction}


Recently, Artificial Intelligence has shown great advances in many varied research areas, but there is one critical area where limited progress has been shown: commonsense knowledge representation and commonsense reasoning \cite{mccarthy1989artificial,BBK01,BlB05,minsky2007emotion,DaM15}. The work introduced in this paper proposes to advance a step forward in this research line by providing a new black-box testing methodology of first-order logic (FOL) \SUMO{}-based ontologies \cite{Niles+Pease'01} that exploits \WORDNET{} \cite{Fellbaum'98} and its mapping into \SUMO{} \cite{Niles+Pease'03}.

Formal ontology development is a discipline whose goal is to derive explicit formal specifications of the concepts in a domain and relations among them \cite{noy2001ontology,Gru09,StS09,ALR12}. As with other software artifacts, ontologies typically have to fulfill some previously specified requirements. Usually both the creation of ontologies and the verification of its requirements are manual tasks that require a significant amount of human effort. In the literature, some methodologies exist that collect the experience in ontology development \cite{GFC04} and, more specifically, in ontology verification \cite{GCC06}.

Roughly speaking, the methodologies for validating functional requirements of ontologies are based on the use of {\it competency questions} (CQs) \cite{GrF95}. That is, according to the requirements of a given ontology, its {\it competency} is described by means of a set of goals or problems that the ontology is expected to answer. Thus, testing an ontology consists in checking whether its set of CQs is effectively answered by the ontology. In this sense, these methods can be classified as {\it black-box} testing \cite{MSB12} according to the classical definition in software engineering, since the definition of questions does not depend on the particular specification of knowledge proposed by the ontology. 
Black-box testing strategies have some disadvantages. For example, it is difficult to determine the coverage level of a set of tests, since different black-box tests can repeatedly check the same portions of software. Further, the process of obtaining CQs is not automatic but creative \cite{FGS13}. Depending on the size and complexity of the ontology, creating a suitable set of CQs is by itself a very challenging and costly task.

\begin{figure}[t]
\centering
\begin{tikzpicture}[>=triangle 60]
\matrix[matrix of math nodes,column sep={180pt,between origins},row sep={40pt,between origins},nodes={asymmetrical rectangle}] (s)
{
|[name=verb]| [ \subsumptionMappingTikZ{Planning} ] : \langle \synsetTikZ{schedule}{2}{v} \rangle & |[name=noun]| \langle \synsetTikZ{schedule}{1}{n} \rangle : [ \subsumptionMappingTikZ{Plan} ] \\
};
\draw[-latex]	(verb) -- node[auto] {\(\langle result \rangle\)} (noun);
\end{tikzpicture}
\caption{Creation of competency questions using \WORDNET{}}
\label{fig:introduction}
\end{figure}

In this paper, we propose a new method for the (semi-)automatic creation of CQs that enables the evaluation of the {\it competency} of \SUMO{}-based ontologies in the sense proposed in \cite{GrF95}. Our proposal for the construction of CQs is based on several predefined question patterns that yield a large set of conjectures by using information from \WORDNET{} and its mapping into \SUMO{}. A preliminary version of our method for the automatic creation of CQs has already been presented in \cite{ALR15}, where we also proposed an adaptation of the methodology for the evaluation of ontologies introduced in \cite{GrF95} to be automatically applied using automated theorem provers (ATPs). As far as we know, our proposals are the first attempts to exploit \WORDNET{} for the evaluation of \SUMO{} and, in general, for the evaluation of knowledge-based resources of this kind. We illustrate our proposal for the creation of CQs using \WORDNET{} by means of the next example: the synsets (sets of synonyms) \synset{schedule}{2}{v} and \synset{schedule}{1}{n} ---which refer to the second sense of the verb {\it schedule} and the first sense of the noun {\it schedule} respectively (see Subsection \ref{subsection:WordNet})--- are related by the semantic relation \textPredicate{result} in \WORDNET{}, as depicted in Figure \ref{fig:introduction}.\footnote{We denote \WORDNET{} synsets and relations between chevrons (angle brackets). In addition, we denote the mapping information of each synset into \SUMO{} separated by colon (:), where \SUMO{} concepts are denoted between square brackets.} In the same figure, we also provide the mapping of \synset{schedule}{2}{v} and \synset{schedule}{1}{n} into \SUMO{}: \synset{schedule}{2}{v} is connected to \subsumptionMapping{Planning} and \synset{schedule}{1}{n} is connected to \subsumptionMapping{Plan}, where the symbol $+$ refers to the \subsumptionMappingRelation{} mapping relation (see Subsection \ref{subsection:WordNet}). Roughly speaking, the mapping states that the semantics of the synsets \synset{schedule}{2}{v} and \synset{schedule}{1}{n} is more specific than the semantics of \textConstant{Planning} and \textConstant{Plan} ---i.e., \textConstant{Planning} and \textConstant{Plan} are more general concepts than \synset{schedule}{2}{v} and \synset{schedule}{1}{n}. Using the above information, we obtain a new conjecture by stating the same fact in terms of \SUMO{}: that is, {\it ``\textConstant{Plan} is \textPredicate{result} of a process of \textConstant{Planning}''}. Indeed, we can propose two different conjectures (CQs) on the basis of the knowledge in Figure \ref{fig:introduction}. In the first one, the statement is assumed to be true in the ontology:\footnote{Assuming that the knowledge in the ontology and \WORDNET{} is correct, and also that the mapping from \WORDNET{} to the ontology is correct, we consider that statement (\ref{goal:PlanPlanning}) is true according to our commonsense knowledge interpretation.}

\vspace{-\baselineskip}
\begin{footnotesize}
\begin{flalign}
%
%
\doubletab & ( \connective{exists} \; ( \variable{X} \; \variable{Y} ) & \label{goal:PlanPlanning} \\
 & \tab ( \connective{and} & \nonumber \\
 & \tab \tab ( \predicate{\$instance} \; \variable{X} \; \constant{Planning} ) & \nonumber \\
 & \tab \tab ( \predicate{\$instance} \; \variable{Y} \; \constant{Plan} ) & \nonumber \\
 & \tab \tab ( \predicate{result} \; \variable{X} \; \variable{Y} ) ) ) & \nonumber
\end{flalign}
\end{footnotesize}
\hspace{-5pt}In the second one, which is obtained by the negation of (\ref{goal:PlanPlanning}), we assume that the statement is false: that is, that {\it ``\textConstant{Plan} is not \textPredicate{result} of any process of \textConstant{Planning}''}. By proceeding in this way, we obtain around 7,500 pairs of CQs on the basis of the information of \WORDNET{} using additional \WORDNET{} relations and question patterns.

The contributions of this paper are manifold. First, we present an evolved version of our methodology for the evaluation of FOL ontologies using ATPs. As introduced in \cite{ALR15}, our proposal is an adaptation of the methodology described in \cite{GrF95} for the design and evaluation of ontologies. Second, we propose a novel method for the (semi-)automatic creation of CQs that relies on a small set of question patterns. The proposed set of CQs enables the evaluation of a) the competency of ontologies derived from \SUMO{}, b) the mapping between \WORDNET{} and \SUMO{}, c) the knowledge in \WORDNET{}, and d) ATPs and other tools for automated reasoning. To the best of our knowledge, our proposal is the first attempt to exploit the information in \WORDNET{} and its mapping into \SUMO{} for the automatic evaluation of knowledge-based resources using FOL ATPs. Third, we summarize the results of an automatic evaluation of the competency of several translations of \SUMO{} into first-order logic (FOL) and the performance of various FOL ATPs by means of the adapted evaluation method proposed in \cite{ALR15}. Fourth, we report on the evaluation of the set of resulting CQs according to different quality criteria. On one hand, we automatically check its level of coverage with respect to the evaluated ontologies by parsing the proofs provided by ATPs. On the other hand, we perform a manual evaluation of a sample of the CQs and analyze in detail their results by considering the quality of proposed conjectures, the mapping information of the involved synsets and the knowledge in the ontology.

{\it Outline of the paper}. In order to make the paper self-contained, in the following section we review the state-of-the-art in automatic evaluation of \SUMO{}-based ontologies using CQs. Our revision includes the existing translations of \SUMO{} into FOL, the most successful FOL ATPs and the previously proposed CQs. In Section \ref{section:methodology}, we describe our methodology for the automatic evaluation of ontologies using ATPs. Next, in Section \ref{section:CQs} we introduce our proposal for the (semi-)automatic creation of CQs by exploiting the knowledge in \WORDNET{} and its mapping into \SUMO{}, with the purpose of evaluating \SUMO{}-based ontologies. The different question patterns proposed for the creation of CQs are described in Sections \ref{section:MultipleMappingPattern} to \ref{section:ProcessPatterns}. Then, we report on our experimental evaluation of the competency of some FOL translations of \SUMO{}, the performance of FOL ATPs and the quality of the proposed CQs in Section \ref{section:experimentation}. Finally, we provide some conclusions and discuss future work in Section \ref{section:conclusions}.

\section{State of the art} \label{section:art}

In this section, we review the state-of-the-art in automatic evaluation of \SUMO{}-based ontologies. For this purpose, we focus on the description of the resources that have been proposed and used in the literature for the evaluation of \SUMO{}-based ontologies using CQs. First, we introduce \SUMO{} and its transformations into FOL in the following subsection. Next, we describe the most successful state-of-the-art FOL ATPs in Subsection \ref{subsection:ATPs}. Finally, we review the CQs that have been previously proposed for the evaluation of \SUMO{}-based ontologies in Subsection \ref{subsection:CQs}.

\subsection{\SUMO{} and its Transformations into FOL} \label{subsection:SUMO}

\SUMO{}\footnote{\url{http://www.ontologyportal.org}} \cite{Niles+Pease'01} has its origins in the nineties, when a group of engineers from the IEEE Standard Upper Ontology Working Group pushed for a formal ontology standard. Their goal was to develop a standard upper ontology to promote data interoperability, information search and retrieval, automated inference and natural language processing.

\SUMO{} is expressed in SUO-KIF (Standard Upper Ontology Knowledge Interchange Format \cite{Pea09}), which is a dialect of KIF (Knowledge Interchange Format \cite{Richard+'92}). Both KIF and SUO-KIF can be used to write FOL formulas, but their syntax goes beyond FOL. Consequently, \SUMO{} cannot be directly used by FOL ATPs without a suitable transformation \cite{ALR12}. With respect to higher-order aspects of \SUMO{}, an additional translation is required for enabling the use of \SUMO{} by means of pure higher-order theorem provers \cite{PeB13}.

Several different proposals for converting large portions of \SUMO{} into a FOL ontology exist. In \cite{PeS07}, the authors report some preliminary experimental results evaluating the query timeout for different options when translating \SUMO{} into FOL. Evolved versions of the translation described in \cite{PeS07} can be found in the {\it Thousands of Problems for Theorem Provers} (TPTP) problem library\footnote{\url{http://www.tptp.org}} \cite{Sut09} (hereinafter \TPTPSUMO{}), but is no longer maintained since TPTP problem library version v5.4.0 (the current TPTP version is v7.0.0). Following the approach of \cite{HoV06}, in \cite{ALR12} we use ATPs for reengineering around 88\% of \SUMO{}, obtaining \ADIMENSUMO{} (v2.2). We are continuously evolving and improving \ADIMENSUMO{} by correcting some of the defects presented in \SUMO{}. As result of this process, we have corrected more than 100 defective axioms in the current version of \ADIMENSUMO{} (v2.6). Both \TPTPSUMO{} and \ADIMENSUMO{} inherits information from the top and the middle levels of \SUMO{} (from now on, the {\it core} of \SUMO{}), thus not considering the information from the domain ontologies.

The knowledge in \SUMO{} is organized around the notions of {\it object} and {\it class} ---the main \SUMO{} concepts. These concepts are respectively defined in \ADIMENSUMO{} by means of the {\it meta}-predicates \textPredicate{\$instance} and \textPredicate{\$subclass}. \SUMO{} objects and classes are not disjoint, since every \SUMO{} class is defined to be instance of \textConstant{class}, and thus every \SUMO{} class is also a \SUMO{} object. Additionally, \SUMO{} also differentiates between {\it relations} and {\it attributes}. In particular, \SUMO{} distinguishes between {\it individual} relation and attributes ---that is, instances of the \SUMO{} classes \textConstant{Relation} and \textConstant{Attribute} respectively--- and {\it classes} of relations and attributes ---that is, subclasses of the \SUMO{} classes \textConstant{Relation} and \textConstant{Attribute} respectively. \SUMO{} provides specific predicates for dealing with relations and attributes. Amongst others, we currently use the next ones in \ADIMENSUMO{}:
\begin{itemize}
\item \textPredicate{subrelation}, which relates two individual \SUMO{} relations (that is, two instances of the \SUMO{} class \textConstant{Relation}). For example, the following \SUMO{} axiom states that \textConstant{member} is subrelation of \textConstant{part}: 

\vspace{-\baselineskip}
\begin{footnotesize}
\begin{flalign}
\doubletab & ( \predicate{subrelation} \; \constant{member} \; \constant{part} ) & \label{axiom:part}
\end{flalign}
\end{footnotesize}
\item \textPredicate{subAttribute}, which relates two individual \SUMO{} attributes (that is, two instances of the \SUMO{} class \textConstant{Attribute}). For example, the following \SUMO{} axiom states that \textConstant{Headache} is subattribute of \textConstant{Pain}:

\vspace{-\baselineskip}
\begin{footnotesize}
\begin{flalign}
\doubletab & ( \predicate{subAttribute} \; \constant{Headache} \; \constant{Pain} ) & \label{axiom:Pain}
\end{flalign}
\end{footnotesize}
\item \textPredicate{holds$^k$}, which relates an individual \SUMO{} relation (that is, an instance of the \SUMO{} class \textConstant{Relation}) with a $k$-tuple of \SUMO{} concepts. For example, the following \ADIMENSUMO{} formula is inherited from the \SUMO{} axiom that characterizes transitive relations: 

\vspace{-\baselineskip}
\begin{footnotesize}
\begin{flalign}
\doubletab & ( \connective{forall} \; ( \variable{REL} ) & \label{axiom:TransitiveRelation} \\
 & \tab ( \connective{<=>} & \nonumber \\
 & \tab \tab ( \predicate{\$instance} \; \variable{REL} \; \constant{TransitiveRelation} ) & \nonumber \\
 & \tab \tab ( \connective{forall} \; ( \variable{INST1} \; \variable{INST2} \; \variable{INST3} ) & \nonumber \\
 & \tab \tab \tab ( \connective{=>} & \nonumber \\
 & \tab \tab \tab \tab ( \connective{and} & \nonumber \\
 & \tab \tab \tab \tab \tab ( \predicate{\$holds3} \; \variable{REL} \; \variable{INST1} \; \variable{INST2} ) & \nonumber \\
 & \tab \tab \tab \tab \tab ( \predicate{\$holds3} \; \variable{REL} \; \variable{INST2} \; \variable{INST3} ) ) & \nonumber \\
 & \tab \tab \tab \tab ( \predicate{\$holds3} \; \variable{REL} \; \variable{INST1} \; \variable{INST3} ) ) ) ) & \nonumber
\end{flalign}
\end{footnotesize}
\item \textPredicate{attribute}, which relates a \SUMO{} object with an individual \SUMO{} attribute (that is, an instance of the \SUMO{} class \textConstant{Attribute}). For example, in the next \SUMO{} axiom the predicate \textPredicate{attribute} is used for the characterization of \textPredicate{subAttribute}:

\vspace{-\baselineskip}
\begin{footnotesize}
\begin{flalign}
\doubletab & ( \connective{forall} \; ( \variable{ATTR1} \; \variable{ATTR2} ) & \label{axiom:subAttribute} \\
 & \tab ( \connective{=>} & \nonumber \\
 & \tab \tab ( \predicate{subAttribute} \; \variable{ATTR1} \; \variable{ATTR2} ) & \nonumber \\
 & \tab \tab ( \connective{forall} \; ( \variable{OBJ} ) & \nonumber \\
 & \tab \tab \tab ( \connective{=>} & \nonumber \\
 & \tab \tab \tab \tab ( \predicate{attribute} \; \variable{OBJ} \; \variable{ATTR1} ) & \nonumber \\
 & \tab \tab \tab \tab ( \predicate{attribute} \; \variable{OBJ} \; \variable{ATTR2} ) ) ) ) & \nonumber
\end{flalign}
\end{footnotesize}
\end{itemize}
For simplicity, from now on we denote the nature of \SUMO{} concepts by adding as subscript the symbols $\SUMOObjectSymbol{}$ (\SUMO{} objects that are neither classes nor individual relations nor individual attributes), $\SUMOClassSymbol{}$ (\SUMO{} classes that are neither classes of relations nor classes of attributes), $\SUMOIndividualRelationSymbol{}$ (individual \SUMO{} relations), $\SUMOIndividualAttributeSymbol{}$ (individual \SUMO{} attributes), $\SUMOClassOfRelationsSymbol{}$ (classes of \SUMO{} relations) and $\SUMOClassOfAttributesSymbol{}$ (classes of \SUMO{} attributes). For example: \SUMOObject{YearDuration}, \SUMOClass{Artifact}, \SUMOIndividualRelation{customer}, \SUMOIndividualAttribute{HotTemperature}, \SUMOClassOfRelations{TranstiveRelation} and \SUMOClassOfAttributes{BreakabilityAttribute}.

\begin{table}[t]
\centering
\begin{tabular} {lrrrrrr}
\hline \\[-10pt]
\multirow{2}{*}{} & \multirow{2}{*}{\hspace{10pt} {\bf \SUMO{}}} & \multicolumn{3}{c}{{\bf \TPTPSUMO{}}} & \multicolumn{2}{r}{{\bf \ADIMENSUMO{}}} \\
\multirow{2}{*}{} & \multirow{2}{*}{} & \multicolumn{3}{c}{v5.3.0} & v2.2 & v2.6 \\
\hline \\[-10pt]
Objects & 20,168 & & 2,920 & & 940 & 1,007 \\
Classes & 5,595 & \hspace{14pt} & 2,086 & & 2,093 & 2,120\\
Relations & 369 & & 208 & & 207 & 207 \\
Attributes & 2,181 & & 68 & & 67 & 66 \\
\hline
\end{tabular}
\caption{\label{table:SUMOFigures} Some figures about \SUMO{}, \TPTPSUMO{} and \ADIMENSUMO{}}
\end{table}

In Table \ref{table:SUMOFigures} we provide some figures comparing the explicit content of \SUMO{}, \TPTPSUMO{} and \ADIMENSUMO{}. In particular, the number of objects, classes, relations (both individual relations and classes of relations) and attributes (both individual attributes and classes of attributes) that are explicitly defined. The most significant difference between \TPTPSUMO{} and \ADIMENSUMO{} is the number of explicitly defined objects, which is due to the fact that during the FOL transformation many objects that are implicitly defined in the core of \SUMO{} are explicitly introduced in \TPTPSUMO{}. On the contrary, the translation from \SUMO{} into \ADIMENSUMO{} is based on a small set of axioms, which provide the axiomatization of \SUMO{} {\it meta}-predicates. Apart from \SUMOIndividualRelation{\$instance} and \SUMOIndividualRelation{\$subclass} for the definition of objects and classes, some of these {\it meta}-predicates are \SUMOIndividualRelation{\$disjoint} and \SUMOIndividualRelation{\$partition}. The axiomatization of these {\it meta}-predicates, which is essential for the transformation of \SUMO{} knowledge into FOL formulas, cannot be directly inherited from \SUMO{} (see \cite{ALR12}). The transformation also adds new axioms for a suitable characterization of \SUMO{} types, variable-arity relations and \HoldsIndividualRelation{\$holds}{k} predicates, which simulate the use of variable-predicates in FOL formulas. 

Nevertheless, \ADIMENSUMO{} (and also \TPTPSUMO{}) does not include most of the instances defined in \SUMO{} since domain ontologies are not translated. To overcome this problem, we include the following axiom in \ADIMENSUMO{} v2.4:

\vspace{-\baselineskip}
\begin{footnotesize}
\begin{flalign}
\doubletab & ( \connective{forall} \; ( \variable{CLASS} ) & \label{axiom:nonEmptyClasses} \\
 & \tab ( \connective{=>} & \nonumber \\
 & \tab \tab ( \predicate{\$subclass} \; \variable{CLASS} \; \constant{Entity} ) & \nonumber \\
 & \tab \tab ( \connective{exists} \; ( \variable{THING} ) & \nonumber \\
 & \tab \tab \tab ( \predicate{\$instance} \; \variable{THING} \; \variable{CLASS} ) ) ) ) & \nonumber
\end{flalign}
\end{footnotesize}
In this fashion, we ensure the existence in \ADIMENSUMO{} (v2.4 or newer) of some instance of every \SUMO{} class although domain ontologies are not translated.

\subsection{\WORDNET{} and its Mapping to \SUMO{}} \label{subsection:WordNet}

\WORDNET{} \cite{Fellbaum'98} is a large lexical database where nouns, verbs, adjectives and adverbs are grouped into sets of synonyms ({\it synsets}), each expressing a distinct concept. Each synset refers to a word sense using the following format: \synset{word}{s}{p}, where $s$ is the sense number and $p$ is the part-of-speech ($n$ for nouns, $v$ for verbs, $a$ for adjectives and $s$ for satellites).

\begin{figure}[t]
\centering
\begin{tikzpicture}
\node at (-3, 2) {\synset{blistering}{2}{s}~~};
\node at (-3, 1) {\synset{warming}{2}{s}~~};
\node at (-3, 0) {\synset{torrid}{3}{s}~~};
\node at (-3,-1) {\synset{heated}{1}{s}~~};
\node at (-3,-2) {\synset{tropical}{4}{s}~~};
\node at ( 0, 0) {\synset{hot}{1}{a}};
\node at ( 1, 0) {/};
\node at ( 2, 0) {\synset{cold}{1}{a}};
\node at ( 5, 2) {~~\synset{gelid}{1}{s}};
\node at ( 5, 1) {~~\synset{frosty}{3}{s}};
\node at ( 5, 0) {~~\synset{heatless}{1}{s}};
\node at ( 5,-1) {~~\synset{refrigerated}{1}{s}};
\node at ( 5,-2) {~~\synset{shivery}{1}{s}};
\draw [-latex] (-2, 2) -- (-0.5, 0.5);
\draw [-latex] (-2, 1) -- (-0.6, 0.25);
\draw [-latex] (-2, 0) -- (-0.75, 0);
\draw [-latex] (-2,-1) -- (-0.6,-0.25);
\draw [-latex] (-2,-2) -- (-0.5,-0.5);
\draw [latex-latex] ( 0.5, 0) -- ( 1.5, 0);
\draw [-latex] (4, 2) -- (2.5, 0.5);
\draw [-latex] (4, 1) -- (2.6, 0.25);
\draw [-latex] (4, 0) -- (2.75, 0);
\draw [-latex] (4,-1) -- (2.6,-0.25);
\draw [-latex] (4,-2) -- (2.5,-0.5);
\end{tikzpicture}
\caption{Antonym-pairs}
\label{fig:antonymPairs}
\end{figure}
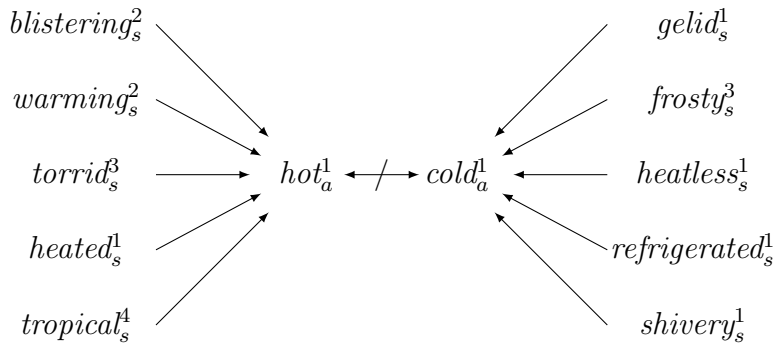

Although superficially resembling a thesaurus, \WORDNET{} interlinks not just word forms but specific senses of words. Thus, the main relation in \WORDNET{} is synonymy, but synsets are interlinked by means of many conceptual-semantic and lexical relations such as the super- and subordinate relations hyperonymy and hyponymy. Amongst them, in this paper we focus on the following ones:
\begin{itemize}

\item {\it Morphosemantic Links} \cite{FOC09}, which are semantic relations between morphologically related verbs and nouns provided in the morphosemantic database.\footnote{Available at \url{http://wordnetcode.princeton.edu/standoff-files/morphosemantic-links.xls}.} Among the 14 proposed semantic relations, one can find {\it agent}, {\it instrument}, {\it result} and {\it event}. The first three ones relate a process (verb) with its corresponding agent/instrument/result (noun), while {\it event} relates nouns and verbs referring to the same process. For example, the synsets \synset{patent}{1}{v} and \synset{patentee}{1}{n} are related by {\it agent}, \synset{cool}{1}{v} and \synset{cooler}{1}{n} are related by {\it instrument}, \synset{schedule}{2}{v} and \synset{schedule}{1}{n} are related by {\it result} (see Figure \ref{fig:introduction}), and the synsets \synset{kill}{10}{v} and \synset{killing}{2}{n} are related by {\it event}.

\item {\it antonymy} and {\it similarity} relations, which are used to organize adjectives as follows: {\it antonymy} connects pairs of adjectives with opposite semantics, and each of these adjectives in turn is linked to semantically comparable adjectives ---called {\it satellites}--- by {\it similarity}. For example, the adjectives \synset{hot}{1}{a} and \synset{cold}{1}{a} are related by {\it antonymy}, and the adjectives \synset{blistering}{2}{s}, \synset{warming}{2}{s}, \synset{torrid}{3}{s}, \synset{heated}{1}{s} and \synset{tropical}{4}{s} are satellites of \synset{hot}{1}{a} (see Figure \ref{fig:antonymPairs}). In addition, {\it antonymy} is inherited by {\it similarity}, which enables the extension of the set of pairs of adjectives related by {\it antonymy}. In the above example, each satellite of \synset{hot}{1}{a} (resp. \synset{cold}{1}{a}) is antonym of \synset{cold}{1}{a} (resp. \synset{hot}{1}{a}) and, furthermore, is also an antonym of each satellite of \synset{cold}{1}{a} (resp. \synset{hot}{1}{a}), thus obtaining a set of 36 antonym-pairs from the information in Figure \ref{fig:antonymPairs}. In addition, {\it antonymy} also relates nouns or verbs with opposite semantics. For example, \synset{natural\_object}{1}{n} and \synset{artifact}{1}{n} are related by the semantic relation {\it antonymy}.

\end{itemize}

\WORDNET{} is linked with \SUMO{} by means of the mapping described in \cite{Niles+Pease'03}. This mapping connects \WORDNET{} synsets to terms in \SUMO{} using three relations: \equivalenceMappingRelation{}, \subsumptionMappingRelation{} and \instanceMappingRelation{}. Additionally, the mapping also uses the complementaries of \equivalenceMappingRelation{} and \instanceMappingRelation{}. We denote mapping relations by concatenating the symbols `$\equivalenceMappingSymbol$' (\equivalenceMappingRelation{}), `$\subsumptionMappingSymbol$' (\subsumptionMappingRelation), `$\instanceMappingSymbol$' (\instanceMappingRelation), `$\negatedEquivalenceMappingSymbol$' (complementary of \equivalenceMappingRelation) and `$\negatedSubsumptionMappingSymbol$' (complementary of \subsumptionMappingRelation) to the corresponding \SUMO{} concept. For example, the synsets \synset{horse}{1}{n}, \synset{education}{4}{n}, \synset{zero}{1}{a}, \synset{natural\_object}{1}{n} and \synset{dark}{1}{a} are connected to \equivalenceMapping{\SUMOClass{Horse}}, \subsumptionMapping{\SUMOClass{EducationalProcess}}, \instanceMapping{\SUMOClass{Integer}}, \negatedEquivalenceMapping{\SUMOClass{Artifact}} and \negatedSubsumptionMapping{\SUMOClass{RadiatingLight}} respectively. \equivalenceMappingRelation{} denotes that the related \WORDNET{} synset and \SUMO{} concept are equivalent in meaning, whereas \subsumptionMappingRelation{} and \instanceMappingRelation{} indicate that the semantics of the \WORDNET{} synset is less general than the semantics of the \SUMO{} concept. In particular, \instanceMappingRelation{} is used when the semantics of the \WORDNET{} synsets refers to a particular member of the class to which the semantics of the \SUMO{} concept is referred.\footnote{Note that \instanceMappingRelation{} denotes the relation that is used in the mapping between \WORDNET{} and \SUMO{} (for example, in \instanceMapping{Integer}), while \SUMOIndividualRelation{\$instance} denotes the meta-predicate that is used in the axiomatization of \ADIMENSUMO{}.} From now on, we say that a \WORDNET{} synset is {\it less general} than the \SUMO{} concepts to which the synset is connected using \subsumptionMappingRelation{} or \instanceMappingRelation{}. 

\WORDNET{} v3.0 consists of 117,659 synsets: 82,115 nouns, 13,767 verbs, 18,156 adjectives and 3,621 adverbs. From the 82,115 noun synsets, 576 synsets are connected to more than one \SUMO{} concept. Furthermore, 1,560 adjective synsets and 179 adverb synsets are not connected to any \SUMO{} concept. All the remaining synsets are connected to a single \SUMO{} concept.

\subsection{FOL Automated Theorem Provers} \label{subsection:ATPs}

The automatic application of methodologies based on CQs requires the use of ATPs. State-of-the-art ATPs for FOL are highly sophisticated systems that have been demonstrated to provide advanced reasoning support to expressive ontologies. Since 1993, many researchers have used the {\it Thousands of Problems for Theorem Provers} (TPTP) problem library as an appropriate and convenient basis for ATP system evaluation \cite{Sut09}, and TPTP has become the {\it de facto} standard set of test problems for classical FOL ATP systems. The performance of ATP systems is evaluated every year in the {\it CADE ATP System Competition} (CASC) \cite{PSS02,SuS06} in the context of a set of problems chosen from the TPTP problem library and applying a specified time limit for each individual problem. Among the systems that have ever participated in CASC, we have selected the ones that are of special interest for reasoning with FOL ontologies, which are Vampire \cite{RiV02} and E \cite{Sch02}. Next, we describe those systems and justify our selection.

The first one is Vampire\footnote{\url{http://www.vprover.org}} \cite{RiV02}, an ATP system for first-order classical logic which has been the winner of the FOF\footnote{First-Order Form non-propositional theorems (axioms with a provable conjecture).} and LTB\footnote{First-order form theorems from Large Theories, presented in Batches.} divisions in CASC during several years. Vampire implements the calculi of ordered binary resolution and superposition for handling equality, and it also implements the Inst-gen calculus. Vampire uses various standard redundancy criteria and implements several simplification techniques for pruning the search space, such as subsumption, tautology deletion, subsumption resolution and rewriting by ordered unit equalities. The reduction ordering is the Knuth-Bendix Ordering. In this paper, we consider four different versions of Vampire that have participated in CASC since 2012: v2.6, v3.0, v4.0 and v4.1. Vampire v2.6 is the CASC-J6 (2012), CASC-24 (2013) and CASC-J7 (2014) FOF division winner, and the CASC-J6 (2012) LTB division winner. Vampire v3.0 obtained 2$^{nd}$ place in the CASC-24 (2013) FOF division, but performed better than the winner (Vampire v2.6), and was used for the experimentation reported in \cite{ALR15}. Vampire v4.0 is the CASC-25 (2015), CASC-J8 (2016) and CASC-26 (2017) LTB division winner, the CASC-25 (2015) and CASC-J8 (2016) FOF division winner, and 
the CASC-25 (2015) FNT\footnote{First-order form non-propositional Non-Theorems.} and EPR\footnote{Effectively PRopositional clause normal form theorems and non-theorems.} divisions winner. In addition, Vampire v4.0 obtained $2^{nd}$ place in the CASC-26 (2017) FOF division. Finally, Vampire v4.1 is the CASC-J8 (2016) and CASC-26 (2017) FNT and TFT\footnote{Typed First-order Theorems.} divisions winner, and also achieved the 2$^{nd}$ place in the CASC-J8 (2016) FOF and LTB divisions.

The second system that we have selected is E \cite{Sch02}, a theorem prover for full FOL with equality which consists of a (optional) clausifier for pre-processing full first-order formulae into clausal form, and a saturation algorithm implementing an instance of the superposition calculus with negative literal selection and a number of redundancy elimination techniques. Among other awards, E has been one of the top three ATP systems in the FOF division of CASC since 2012. E has also been used as a subcomponent by some other competitors in CASC. For its evaluation, we use E v2.0, which is available at \url{http://www.eprover.org}.

\subsection{Available Competency Questions for \SUMO{}} \label{subsection:CQs}

In this subsection, we review the CQs that have been proposed in the literature for the evaluation of \SUMO{}-based ontologies. We classify those CQs into 2 sets, depending on the nature of their creation method.

On one hand, the first set consists of only 64 CQs that have been manually created ({\it creative} CQs). This set includes the 33 CQs belonging to the {\it Commonsense Reasoning} (CSR) domain of the TPTP problem library that is based on \SUMO{}. For example, the following conjecture that belongs to the CSR domain of the TPTP problem library

\vspace{-\baselineskip}
\begin{footnotesize}
\begin{flalign}
\doubletab & ( \connective{forall} \; ( \variable{ORG1} \; \variable{ORG2} \; \variable{ORG3} ) & \label{goal:SiblingMother} \\
 & \tab ( \connective{=>} & \nonumber \\
 & \tab \tab ( \connective{and} & \nonumber \\
 & \tab \tab \tab ( \predicate{mother} \; \variable{ORG1} \; \variable{ORG2} ) & \nonumber \\
 & \tab \tab \tab ( \predicate{sibling} \; \variable{ORG1} \; \variable{ORG3} ) ) & \nonumber \\
 & \tab \tab ( \predicate{mother} \; \variable{ORG3} \; \variable{ORG2} ) ) ) & \nonumber
\end{flalign}
\end{footnotesize}
\hspace{-5pt}states that {\it ``Siblings have the same mother''} as follows: the mother of an organism \textVariable{ORG3} is \textVariable{ORG2} whenever \textVariable{ORG2} is mother of some other organism \textVariable{ORG1} such that \textVariable{ORG1} and \textVariable{ORG3} are sibling. In the past, the CSR domain was part of the set of eligible problems for the LTB division in CASC, but is not currently used. In addition, we have proposed 5 creative CQs in \cite{ALR12} and 26 creative CQs in \cite{ALR15}. For example, the conjectures {\it ``Plants do not suffer from headache''} \cite{ALR12} and {\it ``Herbivores eat animals''} \cite{ALR15}:

\vspace{-\baselineskip}
\begin{footnotesize}
\begin{flalign}
\doubletab & ( \connective{=>} & \label{goal:HeadachePlant} \\
 & \tab ( \predicate{attribute} \; \variable{OBJ} \; \constant{Headache} ) & \nonumber \\
 & \tab ( \connective{not} & \nonumber \\
 & \tab \tab ( \predicate{\$instance} \; \variable{OBJ} \; \constant{Plant} ) ) ) & \nonumber \\[5pt]
 & ( \connective{exists} \; ( \variable{HERBIVORE} \; \variable{ANIMAL} \; \variable{EATING} ) & \label{goal:HerbivoreFalsityTest} \\
 & \tab ( \connective{and} & \nonumber \\
 & \tab \tab ( \predicate{\$instance} \; \variable{HERBIVORE} \; \constant{Herbivore} ) & \nonumber \\
 & \tab \tab ( \predicate{\$instance} \; \variable{ANIMAL} \; \constant{Animal} ) & \nonumber \\
 & \tab \tab ( \predicate{\$instance} \; \variable{EATING} \; \constant{Eating} ) & \nonumber \\
 & \tab \tab ( \predicate{agent} \; \variable{EATING} \; \variable{HERBIVORE} ) & \nonumber \\
 & \tab \tab ( \predicate{patient} \; \variable{EATING} \; \variable{ANIMAL} ) ) ) & \nonumber
\end{flalign}
\end{footnotesize}
\hspace{-5pt}Obviously, conjecture (\ref{goal:HeadachePlant}) is assumed to be true and conjecture (\ref{goal:HerbivoreFalsityTest}) is assumed to be false according to commonsense knowledge.

On the other hand, the second set consists of the CQs that have been obtained by following a (semi-)automatic process ({\it automatically generated} CQs). To the best of our knowledge, the first proposal for the (semi-)automatic creation of CQs is described in \cite{ALR15}, where we introduced a preliminary version of the method described in this paper for the exploitation of \WORDNET{} and its mapping into \SUMO{}. Among other restrictions, we focused on synsets connected to \SUMO{} classes, and thus we discarded much of the mapping information. The resulting set of 7,112 CQs have been used for the automatic evaluation of ATP systems reported in \cite{ALR16}. We provide more details about this preliminary version of our proposal in Section \ref{section:CQs}. In addition, we have applied the same methodology for the creation of CQs on the basis of the meronymy relations of \WORDNET{}, as described in \cite{AlR18,AGR18}. The resulting benchmark consists of 4,290 CQs.

\section{Automatic Evaluation of FOL Ontologies using CQs} \label{section:methodology}

In this section, we summarize our adaptation of the methodology for the design and evaluation of ontologies introduced in \cite{GrF95} to be automatically applied using state-of-the-art ATPs, as initially proposed in \cite{ALR15}.

In \cite{GrF95}, the authors propose to evaluate the expressiveness of an ontology by proving completeness theorems w.r.t. a set of CQs: that is, the conditions under which the solutions to the CQs are complete. The proof of completeness theorems requires checking whether a given CQ is entailed by the ontology or not: that is, given an ontology $\Phi$ and a conjecture $\phi$, we must decide if $\Phi \models \phi$. For this purpose, in \cite{ALR15} we propose to use ATPs such as Vampire \cite{RiV02} and E \cite{Sch02} that work by refutation\footnote{The proof that a conjecture is entailed by an ontology consists in demonstrating that the formula resulting from the conjunction of the ontology and the negation of the conjecture is unsatisfiable.} within some given execution-time and memory limits. Theoretically, if the conjecture is entailed by the ontology, then ATPs will eventually find a refutation given enough time (and space). However, theorem proving in FOL is a very hard problem, so it is not reasonable to expect ATPs to find a proof for every entailed conjecture \cite{KoV13}. Thus, if ATPs can find a proof for a conjecture $\phi$ in an ontology $\Phi$, then we can be sure that the corresponding CQ is entailed by $\Phi$: that is, $\Phi \models \phi$. On the contrary, if ATPs cannot find a proof, we do not know if (a) the conjecture is not entailed by the ontology ($\Phi \not\models^? \phi$) or (b) although the conjecture is entailed, ATPs have not been able to find the proof within the provided execution-time and memory limits ($\Phi \models^? \phi$). Due to the semi-decidability problem of FOL, increasing the execution-time and memory limits is not a solution for conjectures that are not entailed. For the same reason, using other systems that do not work by refutation (for example, by model generation) is not a general solution.

Furthermore, we also propose the division of the set of CQs into two classes: {\it truth-tests} and {\it falsity-tests}, depending on whether we expect the conjecture to be entailed by the ontology or not. An example of truth-test is conjecture (\ref{goal:SiblingMother}) ---{\it ``Siblings have the same mother''}---, which belongs to the CSR domain of the TPTP problem library, because it is expected to be entailed. On the contrary, conjecture (\ref{goal:HerbivoreFalsityTest}) ---{\it ``Herbivores eat animals''}---, which belongs to the set of CQs proposed in \cite{ALR15}, is a falsity-test since it is not expected to be entailed by the ontology.

In order to overcome the problem of deciding whether CQs are entailed or not by the ontology using ATPs, we propose the classification of CQs as either (i) {\it passing}, (ii) {\it non-passing} or (iii) {\it unknown} using the following criteria:
\begin{itemize}
\item If ATPs find a proof, then {\it truth-tests} are classified as {\it passing} since the corresponding conjectures are expected to be entailed, while {\it falsity-tests} are classified as {\it non-passing}, because the corresponding conjectures are expected not to be entailed. For example, ATPs easily prove that conjecture (\ref{goal:SiblingMother}) is entailed by \ADIMENSUMO{} v2.6, thus the truth-test is classified as {\it passing}.
\item Otherwise, if no proof is found, then we classify both {\it truth-} and {\it falsity-tests} as {\it unknown} because we do not know whether the corresponding conjectures are entailed or not. For example, conjecture (\ref{goal:HerbivoreFalsityTest}) is classified as {\it unknown} according to \ADIMENSUMO{} v2.6.
\end{itemize}

\setlength\dashlinedash{0.2pt}
\setlength\dashlinegap{5pt}
\setlength\arrayrulewidth{0.3pt}


\begin{table}[t]
\centering
\resizebox{\textwidth}{!}{
\begin{tabular} {ll;{2.5pt/2.5pt}ll;{2.5pt/2.5pt}ll;{2.5pt/2.5pt}l}
\hline 
\multicolumn{2}{c;{2.5pt/2.5pt}}{\bf Problem} & \multicolumn{4}{c}{\bf Condition} & \multicolumn{1}{;{2.5pt/2.5pt}c}{\multirow{2}{*}{\bf Assessment}} \\
\multicolumn{2}{c;{2.5pt/2.5pt}}{\bf classification} & \multicolumn{2}{c}{\bf Truth-test} & \multicolumn{2}{c}{\bf Falsity-test} & \multicolumn{1}{;{2.5pt/2.5pt}c}{\multirow{2}{*}{}} \\
\hline
\multirow{4}{*}{Solved} & \multirow{2}{*}{Entailed} & \multirow{2}{*}{Passing} & \multirow{2}{*}{($\Phi \models \TT{\phi}$)} & \multirow{2}{*}{Unknown} & ($\Phi \models^? \FT{\phi}$) & $\phi$ is redundant knowledge \\ 
\multirow{4}{*}{} & \multirow{2}{*}{} & \multirow{2}{*}{} & \multirow{2}{*}{} & \multirow{2}{*}{} & ($\Phi \not\models^? \FT{\phi}$) & $\Phi$ is validated against $\phi$ \\
\cdashline{3-7}[2.5pt/2.5pt]
\multirow{4}{*}{} & \multirow{2}{*}{Incompatible} & \multirow{2}{*}{Unknown} & ($\Phi \models^? \TT{\phi}$) & \multirow{2}{*}{Non-passing} & \multirow{2}{*}{($\Phi \models \FT{\phi}$)} & $\Phi$ and $\phi$ are incompatible \\ 
\multirow{4}{*}{} & \multirow{2}{*}{} & \multirow{2}{*}{} & ($\Phi \not\models^? \TT{\phi}$) & \multirow{2}{*}{} & \multirow{2}{*}{} & There is a defect in $\Phi$ \\

\hdashline[2.5pt/2.5pt]

\multicolumn{2}{l;{2.5pt/2.5pt}}{} & Passing & ($\Phi \models \TT{\phi}$) & Non-passing & ($\Phi \models \FT{\phi}$) & $\Phi$ is inconsistent \\

\cdashline{3-7}[2.5pt/2.5pt]

\multicolumn{2}{l;{2.5pt/2.5pt}}{\multirow{3}{*}{Unsolved}} & \multirow{3}{*}{Unknown} & \multirow{2}{*}{($\Phi \models^? \TT{\phi}$)} & \multirow{3}{*}{Unknown} & \multirow{2}{*}{($\Phi \models^? \FT{\phi}$)} & Is $\phi$ new knowledge? \\ 

\multicolumn{2}{l;{2.5pt/2.5pt}}{\multirow{3}{*}{}} & \multirow{3}{*}{} & \multirow{2}{*}{($\Phi \not\models^? \TT{\phi}$)} & \multirow{3}{*}{} & \multirow{2}{*}{($\Phi \not\models^? \FT{\phi}$)} & Is $\phi$ redundant? \\ 

\multicolumn{2}{l;{2.5pt/2.5pt}}{\multirow{3}{*}{}} & \multirow{3}{*}{} & \multirow{3}{*}{} & \multirow{3}{*}{} & \multirow{3}{*}{} & Is there any defect in $\Phi$? \\

\hline
\end{tabular}
}
\caption{\label{table:Methodology} Evaluating FOL Ontologies Using ATPs}
\end{table}

As discussed for the example in Figure \ref{fig:introduction}, truth- and falsity-tests can be interpreted as complementary conjectures. That is, given a truth-test $\phi$, one can propose its negation $\neg \phi$ as falsity-test, and {\it vice versa}. 
For example, the following truth-test ---{\it ``Herbivores do not eat animals''}--- is obtained by the negation of (\ref{goal:HerbivoreFalsityTest}):

\vspace{-\baselineskip}
\begin{footnotesize}
\begin{flalign}
\doubletab & ( \connective{forall} \; ( \variable{HERBIVORE} \; \variable{ANIMAL} \; \variable{EATING} ) & \label{goal:HerbivoreTruthTest} \\
 & \tab ( \connective{=>} & \nonumber \\
 & \tab \tab ( \connective{and} & \nonumber \\
 & \tab \tab \tab ( \predicate{\$instance} \; \variable{HERBIVORE} \; \constant{Herbivore} ) & \nonumber \\
 & \tab \tab \tab ( \predicate{\$instance} \; \variable{ANIMAL} \; \constant{Animal} ) & \nonumber \\
 & \tab \tab \tab ( \predicate{\$instance} \; \variable{EATING} \; \constant{Eating} ) ) & \nonumber \\
 & \tab \tab ( \connective{not} & \nonumber \\
 & \tab \tab \tab ( \connective{and} & \nonumber \\
 & \tab \tab \tab \tab ( \predicate{agent} \; \variable{EATING} \; \variable{HERBIVORE} ) & \nonumber \\
 & \tab \tab \tab \tab ( \predicate{patient} \; \variable{EATING} \; \variable{ANIMAL} ) ) ) ) ) & \nonumber
\end{flalign}
\end{footnotesize}
\hspace{-5pt}Conjecture (\ref{goal:HerbivoreTruthTest}) is classified as {\it passing} according to \ADIMENSUMO{} v2.6. In the same way, we obtain a new falsity-test by negating conjecture (\ref{goal:SiblingMother}). Hence, in general we can assume that any set of CQs that is used for the evaluation of FOL ontologies consists of complementary truth- and falsity-tests. Furthermore, from now on we consider a truth-test $\phi$ and its negative counterpart $\neg \phi$ as a single {\it problem} consisting of two conjectures. For the sake of simplicity, we denote each problem by its truth-test. Thus, the truth-test of a problem $\phi$ is $\phi$ itself, and the falsity-test of a problem $\phi$ is $\neg \phi$.

In Table \ref{table:Methodology}, we describe the evaluation of a FOL ontology $\Phi$ on the basis of a set of problems that are assumed to be true using ATPs. For each problem, we distinguish four cases. In the first two cases, a problem $\phi$ is decided to be {\it solved} because ATPs find a proof for either its truth-test $\phi$ or its falsity-test $\neg \phi$. If ATPs prove only $\Phi \models \phi$ (that is, $\Phi \models^? \neg \phi$ and $\Phi \not\models^? \neg \phi$), then we know that the knowledge in $\phi$ is already included in the ontology and, consequently, we say that the problem $\phi$ is {\it entailed} by (also {\it compatible} with) the ontology $\Phi$. Otherwise, when ATPs prove only $\Phi \models \neg \phi$ (that is, $\Phi \models^? \phi$ and $\Phi \not\models^? \phi$), this reveals the existence of a defect in the ontology since we assume that $\phi$ is true. Therefore, we can say that the problem $\phi$ is {\it incompatible} with the ontology. In the last two cases, the problem $\phi$ remains {\it unsolved}. On one hand, if $\Phi$ is inconsistent then ATPs find a proof for its truth- and falsity-test, which are classified as passing and non-passing respectively. Since falsity-tests are obtained by the negation of truth-tests and a consistent formula cannot entail a formula and its negation, then we can be certain that $\Phi$ is inconsistent in this case. On the other hand, both the truth- and the falsity-test of a problem $\phi$ are classified as unknown because ATPs do not find any proof before running out of resources. Hence, we have no information for the evaluation of $\Phi$ according to the problem $\phi$ and, more specifically, we do not know whether:
\begin{itemize}
\item $\phi$ is new knowledge that could be included in $\Phi$ for improving the knowledge in the ontology.
\item $\phi$ is either redundant ---that is, $\Phi$ already entails $\phi$--- or incompatible with $\Phi$ ---that is, $\Phi \models \neg \phi$---, since ATPs cannot find a proof within the given resources of time and memory.
\end{itemize}



\section{Automatic Creation of CQs Using \WORDNET{}} \label{section:CQs}

In this section, we introduce our proposal for the creation of problems by exploiting \WORDNET{} and its mapping into \SUMO{}, as introduced with the example in Figure \ref{fig:introduction}. Our proposal is a substantially evolved version of the method presented in \cite{ALR15}. Amongst other improvements, we now make use of the mapping relations between \WORDNET{} and \SUMO{}, that were equally addressed in \cite{ALR15}, and we are now able to exploit additional \WORDNET{} information. In addition, we have also improved the process of obtaining a mapping between \WORDNET{} and the core of \SUMO{}. Therefore, the set of CQs introduced in this work ---which is different from the one introduced in \cite{ALR15}--- enables richer exploitation of the knowledge in \WORDNET{} and its mapping into \SUMO{}. In the following subsections, we first describe the method for obtaining a mapping from \WORDNET{} into \ADIMENSUMO{} (Subsection \ref{subsection:AdimenSUMOMapping}). Then, we introduce the method for the translation of \WORDNET{} knowledge into \ADIMENSUMO{} statements in Subsection \ref{subsection:AdimenSUMOStatements}. Finally, we focus on the description of the \WORDNET{} knowledge and the hypothesis that are the basis of our proposal in Subsection \ref{subsection:Exploiting}.

\subsection{Obtaining a mapping between \WORDNET{} and the core of \SUMO{}} \label{subsection:AdimenSUMOMapping}

The mapping between \WORDNET{} and \SUMO{} uses terms from the core ---top and middle levels--- of \SUMO{}, but also from the domain ontologies. However, both \TPTPSUMO{} and \ADIMENSUMO{} use only axioms from the core of \SUMO{}.

A full mapping between \WORDNET{} and the core of \SUMO{} is obtained by means of the structural relations of \SUMO{}: \SUMOIndividualRelation{\$instance}, \SUMOIndividualRelation{\$subclass}, \SUMOIndividualRelation{subrelation} and \SUMOIndividualRelation{subAttribute}. Since \SUMOIndividualRelation{\$subclass}, \SUMOIndividualRelation{subrelation} and \SUMOIndividualRelation{subAttribute} are transitive and, additionally, the relations \SUMOIndividualRelation{\$instance}, \SUMOIndividualRelation{subrelation} and \SUMOIndividualRelation{subAttribute} are inherited through \SUMOIndividualRelation{\$subclass}, it is not difficult to obtain the super-concepts of each \SUMO{} concept. By proceeding in this way, for each \SUMO{} concept that is not defined in the core of \SUMO{} we have obtained its set of most-specific super-concepts that are defined in the core of \SUMO{}. If a \SUMO{} concept is already defined in the core of \SUMO{}, then its set of most-specific \SUMO{} concepts defined in the core of \SUMO{} exclusively consists of itself. Additionally, we have manually corrected some minor and typographical errors affecting 293 \SUMO{} concepts. To summarize, 24,906 \SUMO{} concepts not defined in the core of \SUMO{} are used in the \WORDNET{}-\SUMO{} mapping, from which 14,472 concepts are related with several (more than one) super-concepts belonging to the core of \SUMO{}, whereas 10,434 concepts are related with a single super-concept. 

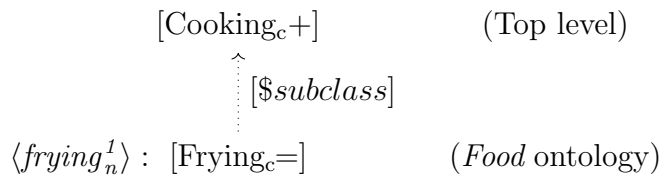
\begin{figure}
\centering
\begin{tikzpicture}[>=triangle 60]
\matrix[matrix of math nodes,column sep={60pt,between origins},row sep={50pt,between origins},nodes={asymmetrical rectangle}] (s)
{
 & |[name=Cooking]| [ \subsumptionMappingTikZ{\SUMOClassTikZ{Cooking}} ] & & |[name=TopLevel]| \mbox{(Top level)} \\
|[name=synset]| \langle \synsetTikZ{frying}{1}{n} \rangle : & |[name=Frying]| [ \equivalenceMappingTikZ{\SUMOClassTikZ{Frying}} ] & & |[name=FoodOntology]| \mbox{({\it Food} ontology)} \\
};
\draw[-To,dotted](Frying) -- node[right] {\([\$subclass]\)} (Cooking);
\end{tikzpicture}
\caption{Obtaining a mapping between \WORDNET{} and \ADIMENSUMO{}}
\label{fig:AdimenSUMOMapping}
\end{figure}

Using the sets of most-specific super-concepts as described above, we obtain the mapping between each synset $ws$ of \WORDNET{} and the core of \SUMO{} as follows: if $ws$ is already mapped into a concept in the core of \SUMO{}, we simply keep the current mapping of $ws$; otherwise, if $ws$ is connected to a concept $C$ that is not defined in the core of \SUMO{}, then we map $ws$ to each element of its set of most-specific super-concepts of $C$ in the core of \SUMO{}. Additionally, in the latter case, the \equivalenceMappingRelation{} mapping relation is replaced with \subsumptionMappingRelation{}, since the super-concepts of $C$ are more general than $C$. For example, the synset \synset{frying}{1}{n} is connected to \equivalenceMapping{\SUMOClass{Frying}}, which belongs to the domain ontology {\it Food}. In the same domain ontology, \SUMOClass{Frying} is defined to be a subclass of \SUMOClass{Cooking}, which is defined in the top level of \SUMO{}. That is, \SUMOClass{Frying} is not defined in the core of \SUMO{}, but \SUMOClass{Cooking} is. Thus, we decide to connect \synset{frying}{1}{n} to \SUMOClass{Cooking} in the mapping from \WORDNET{} to the core of \SUMO{}. However, instead of \equivalenceMappingRelation{}, we connect \synset{frying}{1}{n} to \SUMOClass{Cooking} using the \subsumptionMappingRelation{} mapping relation: that is, \subsumptionMapping{\SUMOClass{Cooking}} (see Figure \ref{fig:AdimenSUMOMapping}). It is worth noting that the complementaries of the relations \equivalenceMappingRelation{} and \subsumptionMappingRelation{} are only used with concepts belonging to the core of \SUMO{} in the \WORDNET{}-\SUMO{} mapping.

As result of this process, we obtain a mapping between all \WORDNET{} synsets and the core of \SUMO{} except for 822 nouns, 24 verbs, 3,634 adjectives and 260 adverbs. In addition to the synsets that are not connected to any concept, this process also reveals the existence of synsets connected to concepts that were defined in older versions of \SUMO{} but that are no longer available in the current version. For example, the synsets \synset{salmon}{1}{n} and \synset{architect}{2}{n} are connected to \equivalenceMapping{\SUMOClass{Salmon}} and \equivalenceMapping{\SUMOClass{Architect}}, which do not appear in recent versions of \SUMO{}. In total, 113 concepts that are used in the \WORDNET{}-\SUMO{} mapping are not currently defined in the ontology. In order to obtain a complete mapping into the core of \SUMO{}, all synsets without a suitable mapping (around 4,700 synsets) are connected to the \SUMO{} top-concept \SUMOClass{Entity} using \subsumptionMappingRelation{}: that is, \subsumptionMapping{\SUMOClass{Entity}}. In the resulting mapping, 1,104 noun synsets and 2 verb synsets are connected to multiple \SUMO{} concepts ---the mapping of those synsets is used in the {\it Multiple mapping} category for the creation of CQs (see Section \ref{section:MultipleMappingPattern})---, whereas the remainder are connected to a single concept.

\subsection{Translating the mapping information into the language of \ADIMENSUMO{}} \label{subsection:AdimenSUMOStatements}

In order to use the \WORDNET{}-\SUMO{} mapping to obtain CQs, we have to characterize the mapping information using statements in the language of \ADIMENSUMO{}.

%
%

%
%

%
%

%
%

%
%

As described in Subsection \ref{subsection:AdimenSUMOMapping}, each \WORDNET{} synset is connect to \SUMO{} concepts using \equivalenceMappingRelation{}, \subsumptionMappingRelation{} (or its complementaries) or \instanceMappingRelation{}. For example, the synsets \synset{horse}{1}{n}, \synset{pony}{1}{n} and \synset{Secretariat}{2}{n} are connected to \equivalenceMapping{\SUMOClass{Horse}}, \subsumptionMapping{\SUMOClass{Horse}} and \instanceMapping{\SUMOClass{Horse}}. Thus, in a literal (or strict) interpretation of the \WORDNET{}-\SUMO{} mapping, \synset{horse}{1}{n} is exactly equivalent to the \SUMO{} concept \SUMOClass{Horse}, while \synset{pony}{1}{n} is less general than \SUMOClass{Horse} and \synset{Secretariat}{2}{n} is an instance of \SUMOClass{Horse}. In order to translate the above interpretation of the mapping information into statements in the language of \ADIMENSUMO{}, we might simply use {\it equality} in the case of the synset \synset{horse}{1}{n}. With respect to the last two synsets, we might use the meta-predicates \SUMOIndividualRelation{\$subclass} and \SUMOIndividualRelation{\$instance} respectively. Likewise, since \synset{male\_horse}{1}{n} is connected to both \subsumptionMappingOfConcept{\SUMOIndividualAttribute{Male}} and \subsumptionMappingOfConcept{\SUMOClass{Horse}}, we have that \synset{male\_horse}{1}{n} is less general than both \SUMOIndividualAttribute{Male} and \SUMOClass{Horse}. Hence, by following the same literal interpretation of the mapping information, \synset{male\_horse}{1}{n} should be translated as both subclass of \SUMOClass{Horse} ---by means of \SUMOIndividualRelation{\$subclass}--- and subattribute of \SUMOIndividualAttribute{Male} ---by means of \SUMOIndividualRelation{subAttribute}. However, this literal interpretation of the mapping information would lead to inconsistent \ADIMENSUMO{} statements: on one hand, \SUMOIndividualRelation{subAttribute} relates two individual \SUMO{} attributes, which are therefore restricted to be instance of \SUMOClass{Attribute}; on the other hand, \SUMOIndividualRelation{\$subclass} relates two \SUMO{} classes, which are defined to be instance of \SUMOClass{class}. Since the \SUMO{} classes \SUMOClass{Attribute} and \SUMOClass{class} are disjoint, it is inconsistent to state that any \SUMO{} concept is both a subclass of \SUMOClass{Horse} and subattribute of \SUMOIndividualAttribute{Male}.

Unlike its literal interpretation, one can propose several suitable translations of the mapping information that do not yield inconsistent \ADIMENSUMO{} statements. Amongst the existing options, in this work we use two different translations of the mapping information on the basis of the following criteria. First, our main purpose is to exploit as much information as possible, to obtain the maximum amount of problems. Second, our intention is also to propose the strongest possible candidate truth-tests. It is worth noting that these two criteria are sometimes contradictory, so we need to find a trade-off between them.

Next, we introduce two different proposals for the translation of the mapping information into \ADIMENSUMO{} statements, where the second proposal produces stronger statements than the first. The purpose of our first proposal is to relate \WORDNET{} synsets with sets of \SUMO{} objects, while the purpose of the second one is to relate \WORDNET{} synsets with \SUMO{} classes. For these purposes, we consider the nature of the \SUMO{} concept to which a synset is connected in order to choose the most suitable \ADIMENSUMO{} predicate: either \SUMOIndividualRelation{equal}, \SUMOIndividualRelation{\$instance}, \SUMOIndividualRelation{\$subclass} or \SUMOIndividualRelation{attribute}.\footnote{In this work, we do not translate the mapping information of synsets connected to \SUMO{} relations. This information should be translated using \HoldsIndividualRelation{\$holds}{k}. However, \HoldsIndividualRelation{\$holds}{k} does not enable the definition of the set of \SUMO{} concepts that is related with a synset. This is due to the fact that the arity of \SUMO{} relations is greater than 1. Consequently, \HoldsIndividualRelation{\$holds}{k} relates \SUMO{} relations with a set of tuples of 2 or more \SUMO{} concepts, instead of a set of (single) \SUMO{} concepts.}

\paragraph{First proposal}

In order to restrict the set of single \SUMO{} objects that can be related with a given synset, we make use of a lenient interpretation of the \WORDNET{}-\SUMO{} mapping. In the proposed \ADIMENSUMO{} statements, we use the predicate \SUMOIndividualRelation{equal} with synsets connected to \SUMO{} objects, the predicate \SUMOIndividualRelation{\$instance} with synsets connected to \SUMO{} classes and \SUMOIndividualRelation{\$attribute} with synsets connected to \SUMO{} attributes. We introduce a new variable in the \ADIMENSUMO{} statement proposed for each individual synset. The quantification of the introduced variables is determined by question patterns and the mapping relation that is used for connecting the given synset (see Sections \ref{section:MultipleMappingPattern} and \ref{section:AntonymPatterns}-\ref{section:ProcessPatterns}). Next, we formalize our proposal for the translation of the mapping information of synsets connected to a single \SUMO{} concept:

\begin{itemize}

\item If the given synset is connected to a {\it \SUMO{} object}, then we simply use {\it equality} to state that the synset is exactly related with that \SUMO{} object. For example, the synset \synset{yearlong}{1}{s} is connected to the \SUMO{} object \SUMOObject{YearDuration}, thus the statement

\vspace{-\baselineskip}
\begin{footnotesize}
\begin{flalign} \label{subCQ:yearlong}
\doubletab & ( \predicate{equal} \; \variable{X} \; \constant{YearDuration} ) & 
\end{flalign}
\end{footnotesize}
\hspace{-5pt}represents that the values of \textVariable{X} related with \synset{yearlong}{1}{s} have to be equal to \SUMOObject{YearDuration}.
%
%

\item If the synset is connected to a {\it \SUMO{} class}, then we use the \ADIMENSUMO{} predicate \SUMOIndividualRelation{\$instance}. For example, \synset{artifact}{1}{n} is connected to the \SUMO{} class \SUMOClass{Artifact}, hence

\vspace{-\baselineskip}
\begin{footnotesize}
\begin{flalign} \label{subCQ:artifact}
\doubletab & ( \predicate{\$instance} \; \variable{X} \; \constant{Artifact} ) & 
\end{flalign}
\end{footnotesize}
\hspace{-5pt}states that the values of \textVariable{X} related with \synset{artifact}{1}{n} must be an instance of \SUMOClass{Artifact}. 

\item If the given synset is connected to an {\it individual \SUMO{} attribute}, we can establish the properties of the \SUMO{} objects related to that synset using the \ADIMENSUMO{} predicate \textPredicate{attribute}.\footnote{Due to the restrictions on arguments of predicates provided by \SUMO{} {\it domain} axioms, we use the \SUMO{} predicate \SUMOIndividualRelation{property} instead of \SUMOIndividualRelation{attribute} when convenient.} 
For example, \synset{goddess}{1}{n} is connected to \SUMOIndividualAttribute{Female} as stated before, therefore the statement

\vspace{-\baselineskip}
\begin{footnotesize}
\begin{flalign} \label{subCQ:femaleAttribute}
\doubletab & ( \predicate{attribute} \; \variable{X} \; \constant{Female} ) &
\end{flalign}
\end{footnotesize}
\hspace{-5pt}states that the values of \textVariable{X} related with \synset{goddess}{1}{n} have \SUMOIndividualAttribute{Female} as a property. 

\item Finally, if the synset is connected to a {\it class of \SUMO{} attributes}, then we have to conveniently combine the \SUMO{} predicates \textPredicate{attribute} and \textPredicate{\$instance}. For example, the synset \synset{breakableness}{1}{n} is connected to \SUMOClassOfAttributes{BreakabilityAttribute}, which denotes a class of \SUMO{} attributes. Hence, the statement

\vspace{-\baselineskip}
\begin{footnotesize}
\begin{flalign}
\doubletab & ( \connective{exists} ( \variable{Z} ) & \label{subCQ:breakablenessAttribute} \\
 & \tab ( \connective{and} & \nonumber \\
 & \tab \tab ( \predicate{\$instance} \; \variable{Z} \; \constant{BreakabilityAttribute} ) & \nonumber \\
 & \tab \tab ( \predicate{attribute} \; \variable{X} \; \variable{Z} ) ) ) ) & \nonumber
\end{flalign}
\end{footnotesize}
\hspace{-5pt}states that the values of \textVariable{X} related with \synset{breakableness}{1}{a} have some instance of \SUMOClassOfAttributes{BreakabilityAttribute} as property.
\end{itemize}

Regardless of the nature of the \SUMO{} concept to which a synset is connected, we negate the statements obtained for synsets connected using the complementary of the \equivalenceMappingRelation{} or the \subsumptionMappingRelation{} mapping relations. For example, the synset \synset{natural\_object}{1}{n} is connected to \negatedEquivalenceMappingOfConcept{\SUMOClass{Artifact}}. By proceeding as described above, we would obtain statement (\ref{subCQ:artifact}). Hence, we negate statement (\ref{subCQ:artifact}) and obtain

\vspace{-\baselineskip}
\begin{footnotesize}
\begin{flalign} \label{subCQ:natural_object}
\doubletab & ( \connective{not} & \\
 & \tab ( \predicate{\$instance} \; \variable{X} \; \constant{Artifact} ) ) & \nonumber
\end{flalign}
\end{footnotesize}
\hspace{-5pt}which states that the values of \textVariable{X} related to \synset{natural\_object}{1}{n} cannot be an instance of \SUMOClass{Artifact}.

In addition, for the translation of the mapping information of synsets connected to more than one \SUMO{} concept, we conveniently combine the statements obtained for each single \SUMO{} concept as previously stated with conjunction. In this way, 
the mapping information of \synset{male\_horse}{1}{n}, which is connected to both \subsumptionMappingOfConcept{\SUMOIndividualAttribute{Male}} and \subsumptionMappingOfConcept{\SUMOClass{Horse}}, is translated as follows:

\vspace{-\baselineskip}
\begin{footnotesize}
\begin{flalign} \label{subCQ:male_horse}
\doubletab & ( \connective{and} & \\
 & \tab ( \predicate{attribute} \; \variable{X} \; \constant{Male} ) & \nonumber \\
 & \tab ( \predicate{\$instance} \; \variable{X} \; \constant{Horse} ) ) & \nonumber
\end{flalign}
\end{footnotesize}
\vspace{-\baselineskip}

\paragraph{Second proposal}

In this proposal for the translation of the mapping information, we obtain stronger statements by restricting the \SUMO{} class ---instead of the \SUMO{} object--- that is related with a given synset. Thus, we consider exclusively those synsets connected to \SUMO{} concepts that are classes and discard the remainder. In the proposed \ADIMENSUMO{} statements, we simply use the predicates \SUMOIndividualRelation{equal} ---for synsets connected by \equivalenceMappingRelation{}--- and \SUMOIndividualRelation{\$subclass} ---for synsets connected by \subsumptionMappingRelation{} or \instanceMappingRelation{}. Therefore, the mapping information of synsets connected by the complementary of \equivalenceMappingRelation{} or \subsumptionMappingRelation{} is also discarded for the moment.

In the following sections, we use the methods proposed above for the translation of the mapping information of synsets to obtain CQs according to different conceptual question patterns. By following our previously introduced criteria, we use the first proposal in Sections \ref{section:MultipleMappingPattern} and \ref{section:AntonymPatterns} to \ref{section:ProcessPatterns}, while the second is used in Section \ref{section:EventPatterns}. In those sections, we also discuss the differences between using each of the proposed translations of the mapping information.

\subsection{Exploiting \WORDNET{} and its Mapping into \SUMO{}} \label{subsection:Exploiting}

This subsection explains how the semantic knowledge of \WORDNET{} and its \SUMO{} mapping introduced in Subsection \ref{subsection:WordNet} are exploited for the construction of CQs. Our proposal is based on the hypothesis that both \WORDNET{} relation-pairs and the mapping information are correct. Under this assumption, we propose different question patterns with two different purposes: first, the validation of the mapping itself and, second, the validation of the knowledge in the ontology according to the knowledge in \WORDNET{}.

Most of the proposed question patterns are based on checking the {\it compatibility}/{\it incompatibility} of the \ADIMENSUMO{} statements obtained from the related \SUMO{} concepts as described in the above subsection. More specifically, each question pattern states the way those \ADIMENSUMO{} statements are combined and the resulting conjecture is checked to be compatible or not. For simplicity, from now on we say that two or more \SUMO{} concepts are {\it compatible}/{\it incompatible} when the \ADIMENSUMO{} statements obtained from them are entailed/incompatible with the ontology (see Table \ref{table:Methodology}).

For the validation of the mapping information, we propose the following two problem categories ({\it Mapping} categories):
\begin{itemize}
\item {\it Multiple mapping} pattern. This category of problems focuses on synsets that are connected to multiple \SUMO{} concepts. Assuming that the mapping is correct, the truth-tests of the proposed problems state that the \SUMO{} concepts connected to the same synset are compatible. Hence, their negations (falsity-tests) state that those \SUMO{} concepts are not compatible, which implies that the mapping is inherently wrong. In Section \ref{section:MultipleMappingPattern}, we describe the single question pattern from which we obtain the problems belonging to this category.
\item {\it Event} patterns. Verbs and nouns referring to the same process are related by {\it event}. Since the synsets in {\it event}-pairs are referring to the same process, we consider that both synsets should be mapped into the same \SUMO{} concept and, if not, our hypothesis is that the mapping information is not correct. Following this hypothesis, for each pair of verb and noun related by {\it event} and connected to different \SUMO{} concepts, we propose a new problem such that its truth-test states that those \SUMO{} concepts are compatible: that is, that the mapping is not necessarily wrong. Thus, the corresponding falsity-tests state that \SUMO{} concepts connected to verbs and nouns related by {\it event} are not compatible and, thus, that the mapping is wrong. This category is divided into 3 subcategories, depending on the mapping relations that are used in {\it event}-pairs. In Section \ref{section:EventPatterns}, we describe in detail the different question patterns and provide examples.
\end{itemize}
In the case of problems proposed for the validation of the knowledge in the ontology, for each \WORDNET{} relation-pair we create a problem such that its truth-test states the same affirmation in terms of \SUMO{}. Next, we describe the two main categories of problems with this purpose ({\it Competency} categories):
\begin{itemize}
\item {\it Antonym} patterns. In this category, problems are obtained from question patterns based on {\it antonymy} as follows: since {\it antonymy} relates adjectives with opposite semantics in \WORDNET{}, for each pair of antonym adjectives we create a new problem such that its truth-test states that the \SUMO{} concepts related to those adjectives are not compatible. Consequently, the corresponding falsity-tests state that the \SUMO{} concepts related to antonym adjectives are compatible. Again, we propose 3 alternative subcategories depending on the mapping relations that are used in the pairs of antonym adjectives. This category is described in Section \ref{section:AntonymPatterns}.
\item {\it Process} patterns. This category consists of question patterns that focus on verbs and nouns related by {\it agent}, {\it instrument} and {\it result}, and the truth-tests of the proposed problems state the same relation in terms of \SUMO{}. For example, conjecture (\ref{goal:PlanPlanning}) states that \synset{schedule}{2}{v} and \synset{schedule}{1}{n} are related by \SUMOIndividualRelation{result} in terms of \SUMO{}. The corresponding falsity-tests state that the \SUMO{} concepts connected to synsets in {\it agent}/{\it instrument}/{\it result}-pairs of verbs and nouns are not semantically related in the same form. We propose a subcategory of problems for each relation and, in addition, an alternative question pattern for each possible combination of mapping relations. We provide a complete description of this category of problems in Section \ref{section:ProcessPatterns}.
\end{itemize}

\section{Multiple Mapping Pattern} \label{section:MultipleMappingPattern}

In this section, we describe the problems that are obtained from synsets connected to several \SUMO{} concepts for the validation of the mapping information.

For this purpose, we assume that both \WORDNET{} relation-pairs and the mapping information of synsets are correct. Under this assumption, from each synset connected to more than one \SUMO{} concept we propose a new problem such that its truth-test states that those \SUMO{} concepts are compatible. Therefore, the corresponding falsity-tests state that the \SUMO{} concepts connected to the same synset are not compatible, which contradicts our assumption. In both cases, we follow the first proposal for the translation of the mapping information described in Subsection \ref{subsection:AdimenSUMOStatements}. This decision is based on the fact that many synsets are connected to \SUMO{} concepts that are not classes, which makes our second proposal for the translation of the mapping information unsuitable.

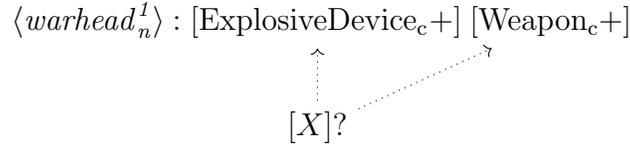
\begin{figure}[t]
\centering
\begin{tikzpicture}[>=triangle 60]
\matrix[matrix of math nodes,column sep={-5pt},row sep={50pt,between origins},nodes={asymmetrical rectangle}] (s)
{
|[name=synset]| \langle \synsetTikZ{warhead}{1}{n} \rangle & |[name=Mapping1]| : [ \subsumptionMappingTikZOfConcept{\SUMOClassTikZ{ExplosiveDevice}} ] & |[name=Mapping2]| [ \subsumptionMappingTikZOfConcept{\SUMOClassTikZ{Weapon}} ] \\[-10pt]
 & |[name=X]| [ X ]? & \\
};
\draw[-To,dotted] (X) -- (Mapping1);
\draw[-To,dotted] (X) -- (Mapping2);
\end{tikzpicture}
\caption{Multiple mapping pattern: \synset{warhead}{1}{n}}
\label{fig:MultipleMapping1}
\end{figure}

As described in Subsection \ref{subsection:AdimenSUMOMapping}, there are 1,106 synsets (1,104 nominal and 2 verbal) connected to more than one \SUMO{} concepts as result of the process of obtaining a mapping from \WORDNET{} to the core of \SUMO{}. Since \equivalenceMappingRelation{} is replaced with \subsumptionMappingRelation{} in that process, all of the synsets are connected using \subsumptionMappingRelation{} or \instanceMappingRelation{}. Hence, in this category we propose a single question pattern for the creation of problems such that its truth-tests state that the \SUMO{} concepts connected to a single synset are compatible. This simply implies that we have to consider the variable in the statement proposed for the translation of the mapping information to be existentially quantified.

For example, \synset{warhead}{1}{n} is connected to \subsumptionMappingOfConcept{\SUMOClass{ExplosiveDevice}} and \subsumptionMappingOfConcept{\SUMOClass{Weapon}} as described in Figure \ref{fig:MultipleMapping1}, from which we obtain the following truth-test that states that \SUMOClass{ExplosiveDevice} and \SUMOClass{Weapon} are compatible:

\vspace{-\baselineskip}
\begin{footnotesize}
\begin{flalign} \label{CQ:ExplosiveDeviceWeapon}
%
\doubletab & ( \connective{exists} \; ( \variable{X} ) & \\
 & \tab ( \connective{and} & \nonumber \\
 & \tab \tab ( \predicate{\$instance} \; \variable{X} \; \constant{ExplosiveDevice} ) & \nonumber \\
 & \tab \tab ( \predicate{\$instance} \; \variable{X} \; \constant{Weapon} ) ) ) & \nonumber
\end{flalign}
\end{footnotesize}
\hspace{-5pt}The corresponding falsity-test, which is obtained by negating (\ref{CQ:ExplosiveDeviceWeapon}), states that \SUMOClass{ExplosiveDevice} and \SUMOClass{Weapon} are not compatible. The mapping of \synset{warhead}{1}{n} is validated since ATPs are able to find a proof for (\ref{CQ:ExplosiveDeviceWeapon}) in \ADIMENSUMO{} v2.6, but not in \TPTPSUMO{} and \ADIMENSUMO{} v2.2. For example, ATPs are able to discover that \SUMOClass{Bomb} is a subclass of both \SUMOClass{ExplosiveDevice} and \SUMOClass{Weapon}, thus any instance of \SUMOClass{Bomb} is also an instance of \SUMOClass{ExplosiveDevice} and \SUMOClass{Weapon} simultaneously. Accordingly, the proposed problem is decided as {\it solved} and {\it entailed} in \ADIMENSUMO{} v2.6, while it is unsolved in \TPTPSUMO{} and \ADIMENSUMO{} v2.2.

\begin{figure}[t]
\centering
\begin{tikzpicture}[>=triangle 60]
\matrix[matrix of math nodes,column sep={-5pt},row sep={50pt,between origins},nodes={asymmetrical rectangle}] (s)
{
|[name=synset]| \langle \synsetTikZ{coal}{1}{n} \rangle & |[name=Mapping1]| : [ \subsumptionMappingTikZOfConcept{\SUMOClassTikZ{FossilFuel}} ] & |[name=Mapping2]| [ \subsumptionMappingTikZOfConcept{\SUMOClassTikZ{Mineral}} ] & |[name=Mapping3]| [ \subsumptionMappingTikZOfConcept{\SUMOClassTikZ{Rock}} ] \\[-10pt]
 & & |[name=X]| [ X ]? & \\
};
\draw[-To,dotted] (X) -- (Mapping1);
\draw[-To,dotted] (X) -- (Mapping2);
\draw[-To,dotted] (X) -- (Mapping3);
\end{tikzpicture}
\caption{Multiple mapping pattern: \synset{coal}{1}{n}}
\label{fig:MultipleMapping2}
\end{figure}
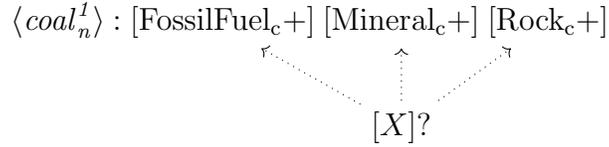

Similarly, \synset{coal}{1}{n} is connected to \subsumptionMappingOfConcept{\SUMOClass{FossilFuel}}, \subsumptionMappingOfConcept{\SUMOClass{Mineral}} and \subsumptionMappingOfConcept{\SUMOClass{Rock}} (see Figure \ref{fig:MultipleMapping2}). Hence, we create a new problem such that its truth-test states that \SUMOClass{FossilFuel}, \SUMOClass{Mineral} and \SUMOClass{Rock} are compatible:

\vspace{-\baselineskip}
\begin{footnotesize}
\begin{flalign} \label{CQ:FossilFuelMineralRock}
%
\doubletab & ( \connective{exists} \; ( \variable{X} ) & \\
 & \tab ( \connective{and} & \nonumber \\
 & \tab \tab ( \predicate{\$instance} \; \variable{X} \; \constant{FossilFuel} ) & \nonumber \\
 & \tab \tab ( \predicate{\$instance} \; \variable{X} \; \constant{Mineral} ) & \nonumber \\
 & \tab \tab ( \predicate{\$instance} \; \variable{X} \; \constant{Rock} ) ) ) & \nonumber
\end{flalign}
\end{footnotesize}
\hspace{-5pt}ATPs find a proof (as before, only in \ADIMENSUMO{} v2.6) for the corresponding falsity-test, which is obtained by negating (\ref{CQ:FossilFuelMineralRock}) and states that \SUMOClass{FossilFuel}, \SUMOClass{Mineral} and \SUMOClass{Rock} are not compatible: for example, ATPs are able to discover that every instance of \SUMOClass{FossilFuel} has \SUMOIndividualAttribute{Liquid} as attribute\footnote{Every instance of \SUMOClass{Solution}, which is a super-class of \SUMOClass{FossilFuel}, has \SUMOIndividualAttribute{Liquid} as attribute.} and every instance of \SUMOClass{Rock} has \SUMOIndividualAttribute{Solid} as attribute, 
although \SUMOIndividualAttribute{Liquid} and \SUMOIndividualAttribute{Solid} are contrary attributes. Consequently, this falsity-test enables the detection of a defect in the mapping information of \synset{coal}{1}{n} and the problem is decided to be {\it solved} and {\it incompatible} in \ADIMENSUMO{} v2.6.

By proceeding in this way, we create 151 problems from the single question pattern proposed in this category.

\section{Event Patterns} \label{section:EventPatterns}

In this section we describe the problems that are obtained from the question patterns based on the semantic relation {\it event} defined in the {\it Morphosemantic Links} database \cite{FOC09} of \WORDNET{} for the validation of synsets mapping information.

For this purpose, in addition to assuming that \WORDNET{} relation-pairs and the synsets mapping are correct, we also assume that \WORDNET{} synsets related by {\it event} should be connected to the same \SUMO{} concept since {\it event} relates verb and noun synsets that refer to the same process. Under those assumptions, for each verb and noun synsets related by {\it event} and connected to different \SUMO{} concepts, we propose a new problem such that its truth-test states that the \SUMO{} concepts linked to those synsets are compatible. Hence, the corresponding falsity-tests state that the \SUMO{} concepts connected to verb and noun synsets related by {\it event} are not compatible, which contradicts our assumptions.

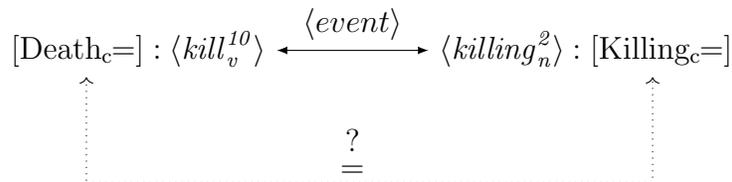
\begin{figure}[t]
\centering
\begin{tikzpicture}[>=triangle 60]
\matrix[matrix of math nodes,column sep={-5pt},row sep={50pt,between origins},nodes={asymmetrical rectangle}] (s)
{
|[name=verbMappingClass]| [ \equivalenceMappingTikZOfConcept{\SUMOClassTikZ{Death}} ] : & |[name=verb]| \langle \synsetTikZ{kill}{10}{v} \rangle & \hspace{60pt} & |[name=noun]| \langle \synsetTikZ{killing}{2}{n} \rangle & |[name=nounMappingClass]| : [ \equivalenceMappingTikZOfConcept{\SUMOClassTikZ{Killing}} ] \\[-15pt]
 & & ? & & \\[-40pt]
 & & = & & \\[-45pt]
|[name=verbMapping]| & & & & |[name=nounMapping]| \\
};
\draw[latex-latex]	(verb) -- node[auto] {\(\langle event \rangle\)} (noun);
\draw[To-,dotted] (verbMappingClass) -- (verbMapping.center);
\draw[-To,dotted] (verbMapping.center) -- (nounMapping.center) -- (nounMappingClass);
\end{tikzpicture}
\caption{Event pattern \#1}
\label{fig:EventPattern1}
\end{figure}

In the \WORDNET{} {\it Morphosemantic Links} database, there are 8,158 event-pairs of synsets where the two synsets are equally mapped to 1,991 event-pairs. In addition, the synsets are connected to different \SUMO{} concepts where at least one is not a \SUMO{} class in only 499 event-pairs. Thus, we decide to apply our second proposal for the translation of the mapping information described in Subsection \ref{subsection:AdimenSUMOStatements} in order to create problems on the basis of the remaining 5,668 event-pairs where the two synsets are connected to different \SUMO{} classes. In this manner, we obtain stronger truth-tests than using our first proposal for the translation of the mapping information. In the following subsections, we introduce different conceptual patterns of questions depending on the mapping relations used.

\subsection{Event Pattern \#1} \label{subsection:Event1}

The first question pattern is focused on the 26 event-pairs where both synsets are connected to two different \SUMO{} classes using \equivalenceMappingRelation{}. Since the mapping of those synsets denotes exactly the \SUMO{} class to which the synset is related, our question pattern states that those \SUMO{} classes are completely equivalent by using {\it equality}.

For example, the synsets \synset{kill}{10}{v} and \synset{killing}{2}{n} are related by {\it event} and connected respectively to the \SUMO{} classes \equivalenceMappingOfConcept{\SUMOClass{Death}} and \equivalenceMappingOfConcept{\SUMOClass{Killing}}, as described in Figure \ref{fig:EventPattern1}, from which we obtain the next truth-test:

\vspace{-\baselineskip}
\begin{footnotesize}
\begin{flalign} \label{CQ:DeathKilling}
%
\doubletab & ( \predicate{equal} \; \constant{Death} \; \constant{Killing} ) &
\end{flalign}
\end{footnotesize}
\hspace{-5pt}The corresponding falsity-test, which is obtained by negating (\ref{CQ:DeathKilling}), states that \SUMOClass{Death} and \SUMOClass{Killing} are different. This falsity-test is classified as non-passing only in \ADIMENSUMO{} v2.6. The proof is based on the fact that \SUMOClass{PhysiologicProcess} and \SUMOClass{PathologicProcess} are disjoint classes. On one hand, \SUMOClass{Death} is a subclass of \SUMOClass{PhysiologicProcess}. On the other hand, \SUMOClass{Killing} is a subclass of a \SUMOClass{Damaging} and every instance of \SUMOClass{Damaging} with some instance of \SUMOClass{Organism} as patient is also instance of \SUMOClass{Injuring}, which is a subclass of \SUMOClass{PathologicProcess}. Consequently, the proposed problem is decided to be {\it solved} and {\it incompatible} in \ADIMENSUMO{} v2.6, which enables the detection of an error in the mapping between \WORDNET{} and \SUMO{}. It is worth noting that in order to state the equivalence of the classes \SUMOClass{Death} and \SUMOClass{Killing} using our first proposal for the translation of the mapping information, we would have to state that the set of objects belonging to those classes are equal, which is a weaker affirmation than conjecture (\ref{CQ:DeathKilling}).

Using this first question pattern, we obtain 24 problems.

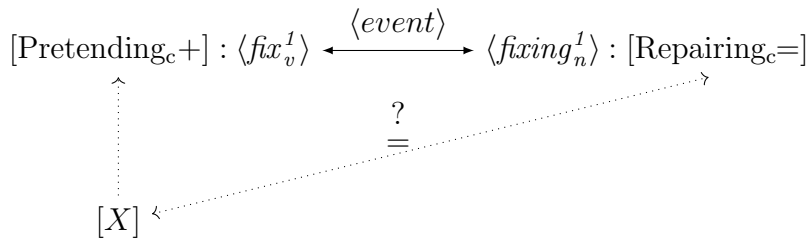
\begin{figure}[t]
\centering
\begin{tikzpicture}[>=triangle 60]
\matrix[matrix of math nodes,column sep={-5pt},row sep={50pt,between origins},nodes={asymmetrical rectangle}] (s)
{
|[name=nounMappingSuperClass1]| [ \subsumptionMappingTikZOfConcept{\SUMOClassTikZ{Pretending}} ] : & |[name=noun1]| \langle \synsetTikZ{fix}{1}{v} \rangle & \hspace{60pt} & |[name=noun2]| \langle \synsetTikZ{fixing}{1}{n} \rangle & |[name=nounMapping2]| : [ \equivalenceMappingTikZOfConcept{\SUMOClassTikZ{Repairing}} ] \\[-25pt]
 & & ? & & \\[-40pt]
 & & = & & \\[-20pt]
|[name=nounMapping1]| [ X ] & & & & \\
};
\draw[-To,dotted] (nounMapping1) -- (nounMappingSuperClass1);
\draw[latex-latex]	(noun1) -- node[auto] {\(\langle event \rangle\)} (noun2);
\draw[To-To,dotted] (nounMapping1) -- (nounMapping2.south);
\end{tikzpicture}
\caption{Event pattern \#2}
\label{fig:EventPattern2}
\end{figure}

\subsection{Event Pattern \#2} \label{subsection:Event2}

In this subsection, we describe the question pattern that focuses on the 509 event-pairs where one synset is connected using \equivalenceMappingRelation{}, while the other synset is connected using \instanceMappingRelation{} or \subsumptionMappingRelation{}. 

In this case, we know the precise \SUMO{} class to which the synset connected by \equivalenceMappingRelation{} is related, as in the previous subsection. However, for the synset connected by \subsumptionMappingRelation{} or \instanceMappingRelation{}, we only know the superclass of the \SUMO{} class to which that synset is related. That is, we know that the synset is connected to some subclass of the class provided in the mapping information. Hence, in order to prove that those \SUMO{} classes are compatible, we must demonstrate that the class related to the synset connected by \equivalenceMappingRelation{} is a subclass of the class related to the synset connected by \subsumptionMappingRelation{} or \instanceMappingRelation{}.

For example, \synset{fix}{1}{v} and \synset{fixing}{1}{n} are related by {\it event} and connected to \subsumptionMappingOfConcept{\SUMOClass{Pretending}} and \equivalenceMappingOfConcept{\SUMOClass{Repairing}} respectively, as described in Figure \ref{fig:EventPattern2}. Therefore, we create a new problem such that its truth-test states that \SUMOClass{Repairing} is a subclass of \SUMOClass{Pretending}

\vspace{-\baselineskip}
\begin{footnotesize}
\begin{flalign} \label{CQ:RepairingPretending}
%
\doubletab & ( \predicate{\$subclass} \; \constant{Repairing} \; \constant{Pretending} ) ) &
\end{flalign}
\end{footnotesize}
\hspace{-5pt}and the corresponding falsity-test states that \SUMOClass{Repairing} cannot be subclass of \SUMOClass{Pretending}. Neither conjecture 
(\ref{CQ:RepairingPretending}) nor its negation are not proved to be entailed by \TPTPSUMO{} or \ADIMENSUMO{} in our experimentation.

From this second event pattern of questions we obtain 350 problems.

\begin{figure}[t]
\centering
\begin{tikzpicture}[>=triangle 60]
\matrix[matrix of math nodes,column sep={-5pt},row sep={50pt,between origins},nodes={asymmetrical rectangle}] (s)
{
|[name=verbMapping1]| [ \subsumptionMappingTikZOfConcept{\SUMOClassTikZ{Comparing}} ] : & |[name=verb]| \langle \synsetTikZ{appraise}{1}{v} \rangle & \hspace{60pt} & |[name=noun]| \langle \synsetTikZ{appraisal}{1}{n} \rangle & |[name=nounMapping1]| : [ \subsumptionMappingTikZOfConcept{\SUMOClassTikZ{Judging}} ] \\[-15pt]
 & & ? & & \\[-40pt]
 & & = & & \\[-45pt]
|[name=verbMapping2]| [ X ] & & & & |[name=nounMapping2]| [ Y ] \\
};
\draw[-To,dotted] (verbMapping2) -- (verbMapping1);
\draw[-To,dotted] (nounMapping2) -- (nounMapping1);
\draw[latex-latex]	(verb) -- node[auto] {\(\langle event \rangle\)} (noun);
\draw[To-To,dotted] (verbMapping2) -- (nounMapping2);
\end{tikzpicture}
\caption{Event pattern \#3}
\label{fig:EventPattern3}
\end{figure}
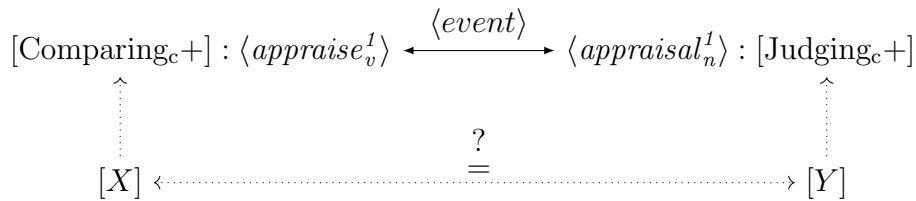

\subsection{Event Pattern \#3} \label{subsection:Event3}

Finally, we focus on the 5,130 event-pairs where both synsets are connected using \instanceMappingRelation{} or \subsumptionMappingRelation{}.

In this case, we only know the superclass of the \SUMO{} class to which each synset is related. Therefore, in order to prove that those \SUMO{} classes are compatible, we have to demonstrate that those \SUMO{} classes have a subclass in common.
 
For example, \synset{appraise}{1}{v} and \synset{appraisal}{1}{n} are related by {\it event} and respectively connected to \subsumptionMappingOfConcept{\SUMOClass{Judging}} and \subsumptionMappingOfConcept{\SUMOClass{Comparing}}, as described in Figure \ref{fig:EventPattern3}. From this event-pair, we create a new problem such that its truth-test states that \SUMOClass{Judging} and \SUMOClass{Comparing} have some common subclasses:

\vspace{-\baselineskip}
\begin{footnotesize}
\begin{flalign} \label{CQ:JudgingComparing}
%
\doubletab & ( \connective{exists} \; ( \variable{X} ) & \\
 & \tab ( \connective{and} & \nonumber \\
 & \tab \tab ( \predicate{\$subclass} \; \variable{X} \; \constant{Judging} ) & \nonumber \\
 & \tab \tab ( \predicate{\$subclass} \; \variable{X} \; \constant{Comparing} ) ) ) & \nonumber
\end{flalign}
\end{footnotesize}
\hspace{-5pt}Thus, the corresponding falsity-test states that \SUMOClass{Judging} and \SUMOClass{Comparing} do not have any common subclass. Conjecture 
(\ref{CQ:JudgingComparing}) and its negation are not shown to be entailed by \TPTPSUMO{} or \ADIMENSUMO{} in our experimentation.

Using this third question pattern based on {\it event}, we obtain 2,011 different problems.

\section{Antonym Patterns} \label{section:AntonymPatterns}

In this section, we describe the problems of the {\it Competency} categories that are obtained from the question patterns based on antonyms. For this purpose, we assume that both \WORDNET{} relation-pairs and their mappings into \SUMO{} are correct. Under those assumptions, questions patterns focus on the {\it antonymy} ---which relates words with opposite semantics--- and {\it similarity} ---that links semantically comparable words--- relations of \WORDNET{} (see Figure \ref{fig:antonymPairs}) and propose the creation of new problems such that their truth-tests state that the \SUMO{} concepts related to antonym words are not compatible. Thus, the corresponding falsity-tests state that the \SUMO{} concepts related to antonym words are compatible.

\WORDNET{} provides 7,604 antonym-pairs, from which 1,950 are noun-pairs, 1,016 are verb-pairs, 3,998 are adjective-pairs and 640 are adverb-pairs. In addition, given a synset $ws$ in an antonym-pair that is related with another synset $ws'$ via similarity, we can propose a new antonym-pair by simply replacing $ws$ with $ws'$ in the pair. In this fashion, we extend the given 7,604 antonym-pairs to a set of 121,496 antonym-pairs. Since many of the synsets in those pairs are connected to \SUMO{} concepts that are not classes, we use our first proposal for the translation of the mapping information described in Subsection \ref{subsection:AdimenSUMOStatements}. Further, in 36,934 antonym-pairs some of the synsets are mapped into \SUMO{} relations and, therefore, those pairs are not considered. In the remaining 84,562 antonym-pairs of synsets, there are:
\begin{itemize}
\item 186 antonym-pairs where both synsets are connected using \equivalenceMappingRelation{} (or its complement).
\item 2,542 antonym-pairs where \equivalenceMappingRelation{} (or its complement) is mixed with \subsumptionMappingRelation{} (or its complement) or \instanceMappingRelation{}.
\item 81,834 antonym-pairs where both synsets are connected using \subsumptionMappingRelation{} (or its complement) and \instanceMappingRelation{}.
\end{itemize}
In the following subsections, we describe 3 alternative question patterns depending on the mapping relations that are used.

\subsection{Antonym Pattern \#1} \label{subsection:Antonym1}


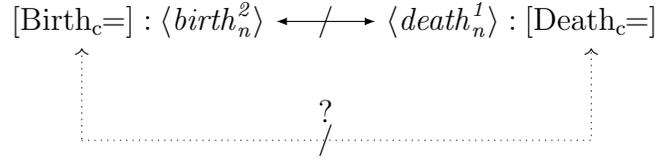
\begin{figure}[t]
\centering
\begin{tikzpicture}[>=triangle 60]
\matrix[matrix of math nodes,column sep={-5pt},row sep={50pt,between origins},nodes={asymmetrical rectangle}] (s)
{
|[name=adjectiveMappingClass1]| [ \equivalenceMappingTikZ{\SUMOClassTikZ{Birth}} ] : & |[name=adjective1]| \langle \synsetTikZ{birth}{2}{n} \rangle & \hspace{40pt} & |[name=adjective2]| \langle \synsetTikZ{death}{1}{n} \rangle : & |[name=adjectiveMappingClass2]| [ \equivalenceMappingTikZ{\SUMOClassTikZ{Death}} ] \\[-15pt]
 & & ? & & \\[-40pt]
|[name=adjectiveMapping1]| & & / & & |[name=adjectiveMapping2]| \\
};
\draw[latex-latex] (adjective1) -- node {\(/\)} (adjective2);
\draw[To-,dotted] (adjectiveMappingClass1) -- (adjectiveMapping1.center);
\draw[-To,dotted] (adjectiveMapping1.center) -- (adjectiveMapping2.center) -- (adjectiveMappingClass2);
\end{tikzpicture}
\caption{Antonym pattern \#1: \synset{birth}{2}{n} and \synset{death}{1}{n}}
\label{fig:AntonymPattern1BirthDeath}
\end{figure}

The first question pattern based on antonym is focused on the 186 antonym-pairs where both synsets are connected using \equivalenceMappingRelation{} (or its complement). In this case, we assume that all the \SUMO{} objects represented by the statement obtained from the first synset are different from all the \SUMO{} objects represented by the statement obtained from the second synset. Formally, this implies that we consider the variables used in the \ADIMENSUMO{} statements proposed for the translation of the mapping information to be universally quantified.

For example, the antonym-synsets \synset{birth}{2}{n} and \synset{death}{1}{n} are respectively connected to \equivalenceMappingOfConcept{\SUMOClass{Birth}} and \equivalenceMappingOfConcept{\SUMOClass{Death}} (see Figure \ref{fig:AntonymPattern1BirthDeath}), from which we obtain the following statements:

\vspace{-\baselineskip}
\begin{footnotesize}
\begin{flalign}
\doubletab & ( \predicate{\$instance} \; \variable{X} \; \constant{Birth} ) & \label{subCQ:birth} \\
\doubletab & ( \predicate{\$instance} \; \variable{Y} \; \constant{Death} ) & \label{subCQ:death}
\end{flalign}
\end{footnotesize}
\hspace{-5pt}By considering \textVariable{X} and \textVariable{Y} to be universally quantified, the following truth-test results from the combination of statements (\ref{subCQ:birth}) and (\ref{subCQ:death}):

\vspace{-\baselineskip}
\begin{footnotesize}
\begin{flalign}
%
\doubletab & ( \connective{forall} ( \variable{X} \; \variable{Y} ) & \label{CQ:BirthDeath} \\
 & \tab ( \connective{=>} & \nonumber \\
 & \tab \tab ( \connective{and} & \nonumber \\
 & \tab \tab \tab ( \predicate{\$instance} \; \variable{X} \; \constant{Birth} ) & \nonumber \\
 & \tab \tab \tab ( \predicate{\$instance} \; \variable{Y} \; \constant{Death} ) ) & \nonumber \\
 & \tab \tab ( \connective{not} & \nonumber \\
 & \tab \tab \tab ( \predicate{equal} \; \variable{X} \; \variable{Y} ) ) ) ) & \nonumber
\end{flalign}
\end{footnotesize}
\hspace{-5pt}The above CQ states that any two \SUMO{} objects that are instance of \SUMOClass{Birth} and \SUMOClass{Death} respectively are inevitably different. The corresponding falsity-test is obtained by negating (\ref{CQ:BirthDeath}), which states that some \SUMO{} object exists which is instance of \SUMOClass{Birth} and \SUMOClass{Death} at the same time. This problem remains unsolved in \TPTPSUMO{} and \ADIMENSUMO{} (v2.2 and v2.6) due to the lack of information in the ontology.

From this question pattern, we obtain 71 different problems.

\subsection{Antonym Pattern \#2} \label{subsection:Antonym2}

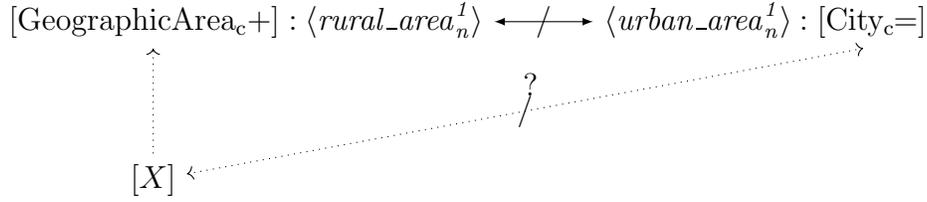
\begin{figure}[t]
\centering
\begin{tikzpicture}[>=triangle 60]
\matrix[matrix of math nodes,column sep={-5pt},row sep={50pt,between origins},nodes={asymmetrical rectangle}] (s)
{
|[name=nounMappingSuperClass1]| [ \subsumptionMappingTikZ{\SUMOClassTikZ{GeographicArea}} ] : & |[name=noun1]| \langle \synsetTikZ{rural\_area}{1}{n} \rangle & \hspace{40pt} & |[name=noun2]| \langle \synsetTikZ{urban\_area}{1}{n} \rangle & |[name=nounMapping2]| : [ \equivalenceMappingTikZ{\SUMOClassTikZ{City}} ] \\[-25pt]
 & & \hspace{-10pt} ? & & \\[-15pt]
|[name=nounMapping1]| [ X ] & & & & \\
};
\draw[-To,dotted] (nounMapping1) -- (nounMappingSuperClass1);
\draw[latex-latex]	(noun1) -- node {\(/\)} (noun2);
\draw[To-To,dotted]	(nounMapping1) -- node {\(/\)} (nounMapping2.south);
\end{tikzpicture}
\caption{Antonym pattern \#2: \synset{rural\_area}{1}{n} and \synset{urban\_area}{1}{n}}
\label{fig:AntonymPattern2RuralUrban}
\end{figure}

The second question pattern is focused on the 2,542 antonym-pairs where \equivalenceMappingRelation{} (or its complement) is mixed with \subsumptionMappingRelation{} (or its complement) or \instanceMappingRelation{}. As in the previous case, we consider the variable in the \ADIMENSUMO{} statement proposed for the translation of equivalence mapping information to be universally quantified. On the contrary, the variable in the \ADIMENSUMO{} statement translating \subsumptionMappingRelation{} or \instanceMappingRelation{} mapping information is considered to be existentially quantified because the information provided by these mapping relations is weaker than the information provided by \equivalenceMappingRelation{}. Since we are using both universally and existentially quantified variables, there are two additional options: we may nest the universally quantified statement inside the formula obtained from the existentially quantified statement, or nest the existentially quantified statement inside the formula that is derived from the universally quantified statement. From these two options, we choose the one that yields stronger truth-tests, which is the first. For example, the antonym synsets \synset{rural\_area}{1}{n} and \synset{urban\_area}{1}{n} are connected to \subsumptionMappingOfConcept{\SUMOClass{GeographicArea}} and \equivalenceMappingOfConcept{\SUMOClass{City}} respectively (see Figure \ref{fig:AntonymPattern2RuralUrban}), from which we obtain the following statements:

\vspace{-\baselineskip}
\begin{footnotesize}
\begin{flalign}
\doubletab & ( \predicate{\$instance} \; \variable{X} \; \constant{GeographicArea} ) & \label{subCQ:ruralArea} \\[5pt]
 & ( \predicate{\$instance} \; \variable{Y} \; \constant{City} ) ) & \label{subCQ:urbanArea}
\end{flalign}
\end{footnotesize}
\hspace{-5pt}Consequently, the \ADIMENSUMO{} statement that is obtained for the \synset{urban\_area}{1}{n} is nested into the \ADIMENSUMO{} statement that is obtained for \synset{rural\_area}{1}{n}:

\vspace{-\baselineskip}
\begin{footnotesize}
\begin{flalign}
%
\doubletab & ( \connective{exists} ( \variable{X} ) & \label{CQ:GeographicAreaCity} \\
 & \tab ( \connective{and} & \nonumber \\
 & \tab \tab ( \predicate{\$instance} \; \variable{X} \; \constant{GeographicArea} ) & \nonumber \\
 & \tab \tab ( \connective{forall} ( \variable{Y} ) & \nonumber \\
 & \tab \tab \tab ( \connective{=>} & \nonumber \\
 & \tab \tab \tab \tab ( \predicate{\$instance} \; \variable{Y} \; \constant{City} ) & \nonumber \\
 & \tab \tab \tab \tab ( \connective{not} & \nonumber \\
 & \tab \tab \tab \tab \tab ( \predicate{equal} \; \variable{X} \; \variable{Y} ) ) ) ) ) ) & \nonumber
\end{flalign}
\end{footnotesize}
\hspace{-5pt}The above CQ states that there exists some \SUMO{} object which is an instance of \SUMOClass{GeographicArea} such that it is different from any \SUMO{} object that is an instance of \SUMOClass{City}. It is worth noting that if the previous statement holds true, it implies that every \SUMO{} object that is an instance of \SUMOClass{City} is different from some \SUMO{} object that is an instance of \SUMOClass{GeographicArea}: in particular, all the \SUMO{} objects that are an instance of \SUMOClass{City} would be different from a single \SUMO{} object that is an instance of \SUMOClass{GeographicArea}. Hence, the truth-test in (\ref{CQ:GeographicAreaCity}) is stronger than the conjecture that results by nesting the existentially quantified statement into the formula obtained from the universally quantified statement. Although \SUMOClass{City} is a subclass of \SUMOClass{GeographicArea}, the truth-test defined in (\ref{CQ:GeographicAreaCity}) is classified as passing in only \ADIMENSUMO{} v2.6 since \SUMOClass{GeographicArea} has other subclasses that are disjoint with \SUMOClass{City}: for example, \SUMOClass{WaterArea}. Therefore, the proposed problem is {\it solved} and {\it entailed} in \ADIMENSUMO{} v2.6.

In summary, we obtain 489 problems from this second question pattern based on {\it antonymy}.

\subsection{Antonym Pattern \#3} \label{subsection:Antonym3}

\begin{figure}[t]
\centering
\begin{tikzpicture}[>=triangle 60]
\matrix[matrix of math nodes,column sep={-5pt},row sep={50pt,between origins},nodes={asymmetrical rectangle}] (s)
{
|[name=adjectiveMappingSuperClass1]| [ \subsumptionMappingTikZ{\SUMOClassTikZ{Coloring}} ] : & |[name=adjective1]| \langle \synsetTikZ{stained}{1}{a} \rangle & \hspace{40pt} & |[name=adjective2]| \langle \synsetTikZ{unstained}{1}{a} \rangle & |[name=adjectiveMappingSuperClass2]| : [ \negatedSubsumptionMappingTikZ{\SUMOClassTikZ{SurfaceChange}} ] \\[-15pt]
 & & \hspace{30pt} ? & & \\[-40pt]
|[name=adjectiveMapping1]| [ X ] & & & & |[name=adjectiveMapping2]| [ Y ] \\
};
\draw[-To,dotted] (adjectiveMapping1) -- (adjectiveMappingSuperClass1);
\draw[-To,dotted] (adjectiveMapping2) -- (adjectiveMappingSuperClass2);
\draw[latex-latex]	(adjective1) -- node {\(/\)} (adjective2);
\draw[To-To,dotted] (adjectiveMapping1) -- node {\(/\)} (adjectiveMapping2);
\end{tikzpicture}
\caption{Antonym pattern \#3: \synset{stained}{1}{a} and \synset{unstained}{1}{a}}
\label{fig:AntonymPattern3ColoringSurfaceChange}
\end{figure}
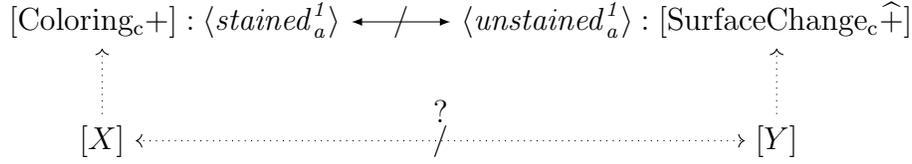

In the third question pattern based on antonym, we focus on the 81,834 antonym-pairs where both synsets are connected using \subsumptionMappingRelation{} (or its complement) or \instanceMappingRelation{}. As before, we consider the variables used in \ADIMENSUMO{} statements to be existentially quantified, which implies we consider that some of the \SUMO{} objects represented by the statements obtained from the mapping information of the antonym synsets are not equal. For example, the antonym synsets \synset{stained}{1}{a} and \synset{unstained}{1}{a} are connected respectively to \subsumptionMappingOfConcept{\SUMOClass{Coloring}} and \negatedSubsumptionMappingOfConcept{\SUMOClass{SurfaceChange}} (see Figure \ref{fig:AntonymPattern3ColoringSurfaceChange}), from which we obtain the following statements:

\vspace{-\baselineskip}
\begin{footnotesize}
\begin{flalign}
\doubletab & ( \predicate{\$instance} \; \variable{X} \; \constant{Coloring} ) & \label{subCQ:stained} \\[5pt]
 & ( \connective{not} & \label{subCQ:unstained} \\
 & \tab ( \predicate{\$instance} \; \variable{Y} \; \constant{SurfaceChanging} ) ) & \nonumber
\end{flalign}
\end{footnotesize}
\hspace{-5pt}Therefore, we propose the following truth-test

\vspace{-\baselineskip}
\begin{footnotesize}
\begin{flalign}
%
\doubletab & ( \connective{exists} ( \variable{X} \; \variable{Y} ) & \label{CQ:ColoringNotSurfaceChange} \\
 & \tab ( \connective{and} & \nonumber \\
 & \tab \tab ( \predicate{\$instance} \; \variable{X} \; \constant{Coloring} ) & \nonumber \\
 & \tab \tab ( \connective{not} & \nonumber \\
 & \tab \tab \tab ( \predicate{\$instance} \; \variable{Y} \; \constant{SurfaceChanging} ) ) & \nonumber \\
 & \tab \tab ( \connective{not} & \nonumber \\
 & \tab \tab \tab ( \predicate{equal} \; \variable{X} \; \variable{Y} ) ) ) ) & \nonumber
\end{flalign}
\end{footnotesize}
\hspace{-5pt}stating that two different \SUMO{} objects exist such that the first is an instance of \SUMOClass{Coloring} and the second one is not an instance of \SUMOClass{SurfaceChanging}. The corresponding falsity-test that is obtained by negating (\ref{CQ:ColoringNotSurfaceChange}) states that any two \SUMO{} objects such that the first one is an instance of \SUMOClass{Coloring} and the second one is not an instance of \SUMOClass{SurfaceChanging} are equal. Although \SUMOClass{Coloring} is a subclass of \SUMOClass{SurfaceChanging}, the truth-test defined in (\ref{CQ:ColoringNotSurfaceChange}) is classified as passing in only \ADIMENSUMO{} v2.6. Therefore, the proposed problem is {\it solved} and {\it entailed} in \ADIMENSUMO{} v2.6, while it is {\it unsolved} in \TPTPSUMO{} and \ADIMENSUMO{} v2.6.

Using this third question pattern, we obtain 2,444 problems.

\section{Process Patterns} \label{section:ProcessPatterns}

In this section, we describe the problems that are obtained from the {\it Morphosemantic Links} database \cite{FOC09} of \WORDNET{} for the validation of the knowledge in the ontology. As in the case of the question patterns based on antonym, we assume that both \WORDNET{} relation-pairs and their mappings into \SUMO{} are correct.

Among the 14 semantics relations between morphologically related verbs and nouns provided by the {\it Morphosemantic Links}, we concentrate on {\it agent}, {\it instrument} and {\it result}, which relate a process (verb) with its corresponding agent/instrument/result (noun). For each pair of synsets connected by the above relations, we propose the creation of a new problem such that its truth-test states the same affirmation in terms of \SUMO{}. For this purpose, we make proper use of the \SUMO{} relations \SUMOIndividualRelation{agent}, \SUMOIndividualRelation{instrument} and \SUMOIndividualRelation{result} that link \SUMO{} processes (i.e., an instance of the \SUMO{} class \textConstant{Process}) to its corresponding agent, instrument and result, which are restricted respectively to being instances of the \SUMO{} classes \SUMOClass{Agent}, \SUMOClass{Physical} and \SUMOClass{Entity}. That is, the \SUMO{} relations \SUMOIndividualRelation{agent}, \SUMOIndividualRelation{instrument} and \SUMOIndividualRelation{result} connect two \SUMO{} objects. Consequently, it is unfeasible to apply our second proposal for the translation of the mapping information described in Subsection \ref{subsection:AdimenSUMOStatements}, since the connected concepts are not \SUMO{} classes. Depending on the mapping relations that are used to relate the verb and noun synsets, we introduce 4 different question patterns by means of our first proposal for the translation of the mapping information.

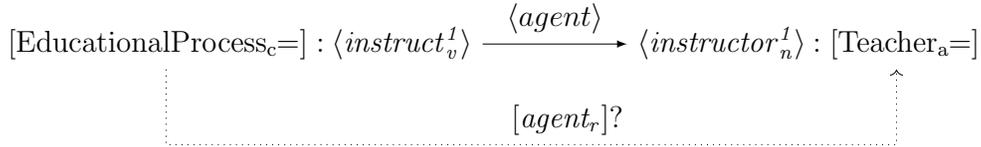
\begin{figure}[t]
\centering
\resizebox{\columnwidth}{!}{
\begin{tikzpicture}[>=triangle 60]
\matrix[matrix of math nodes,column sep={-5pt},row sep={50pt,between origins},nodes={asymmetrical rectangle},ampersand replacement=\&] (s)
{
|[name=verbMappingClass]| [ \equivalenceMappingTikZOfConcept{\SUMOClassTikZ{EducationalProcess}} ] : \& |[name=verb]| \langle \synsetTikZ{instruct}{1}{v} \rangle \& \hspace{60pt} \& |[name=noun]| \langle \synsetTikZ{instructor}{1}{n} \rangle \& |[name=nounMappingClass]| : [ \equivalenceMappingTikZOfConcept{\SUMOIndividualAttributeTikZ{Teacher}} ] \\[-20pt]
 \& \& \hspace{10pt} [\SUMOIndividualRelationTikZ{agent}]? \& \& \\[-40pt]
|[name=verbMapping]| \& \& \& \& |[name=nounMapping]| \\
};
\draw[-latex]	(verb) -- node[auto] {\(\langle agent \rangle\)} (noun);
\draw[-To,dotted] (verbMappingClass) -- (verbMapping.center) -- (nounMapping.center) -- (nounMappingClass);
\end{tikzpicture}
}
\caption{Process pattern: verb and noun synsets connected by \equivalenceMappingRelation{}}
\label{fig:ProcessPattern1}
\end{figure}
 
In the {\it Morphosemantic Links} database, there are 5.295 relation-pairs of synsets where one of the relations {\it agent}, {\it instrument} or {\it result} are used. Among those pairs, there are 5,098 relation-pairs such that none of the synsets are connected to a \SUMO{} relation and, additionally, none of the synsets is connected using the complementary of \equivalenceMappingRelation{} or \subsumptionMappingRelation{}. Therefore, we use those 5,098 relation-pairs for creating problems. For example, the synsets \synset{instruct}{1}{v} and \synset{instructor}{1}{n} are related by {\it agent} and connected respectively to \equivalenceMappingOfConcept{\SUMOClass{EducationalProcess}} and \equivalenceMappingOfConcept{\SUMOIndividualAttribute{Teacher}} (see Figure \ref{fig:ProcessPattern1}). From the mapping information of those synsets, we obtain the following statements:

\vspace{-\baselineskip}
\begin{footnotesize}
\begin{flalign}
\doubletab & ( \predicate{\$instance} \; \variable{X} \; \constant{EducationalProcess} ) & \label{subCQ:instruct} \\[5pt]
 & ( \predicate{attribute} \; \variable{Y} \; \constant{Teacher} ) & \label{subCQ:instructor}
\end{flalign}
\end{footnotesize}
\hspace{-5pt}Thus, we must combine the above statements using the \SUMO{} relation \SUMOIndividualRelation{agent} and quantify their variables in order to create a problem. However, unlike the case of {\it event} question patterns, we cannot consider both variables in \ADIMENSUMO{} statements to be universally quantified when both synsets are connected by \equivalenceMappingRelation{} (see Subsection \ref{subsection:Event1}): in our example, it is not true that all the \SUMO{} objects with \SUMOIndividualAttribute{Teacher} as an attribute are the agent of all the instances of \SUMOClass{EducationalProcess}. At most, we can state that all the instances of \SUMOClass{EducationalProcess} have a \SUMO{} object with the attribute \SUMOIndividualAttribute{Teacher} as agent and, at the same time, all the \SUMO{} objects with \SUMOIndividualAttribute{Teacher} as attribute are the agent of some instance of \SUMOClass{EducationalProcess}, as proposed in the next conjecture (truth-test):

\vspace{-\baselineskip}
\begin{footnotesize}
\begin{flalign}
%
\doubletab & ( \connective{and} & \label{CQ:AgentEducationalProcessTeacher} \\
 & \tab ( \connective{forall} ( \variable{X} ) & \nonumber \\
 & \tab \tab ( \connective{=>} & \nonumber \\
 & \tab \tab \tab ( \predicate{\$instance} \; \variable{X} \; \constant{EducationalProcess} ) & \nonumber \\
 & \tab \tab \tab ( \connective{exists} ( \variable{Y} ) & \nonumber \\
 & \tab \tab \tab \tab ( \connective{and} & \nonumber \\
 & \tab \tab \tab \tab \tab ( \predicate{attribute} \; \variable{Y} \; \constant{Teacher} ) & \nonumber \\
 & \tab \tab \tab \tab \tab ( \predicate{agent} \; \variable{X} \; \variable{Y} ) ) ) ) ) & \nonumber \\
 & \tab ( \connective{forall} ( \variable{Y} ) & \nonumber \\
 & \tab \tab ( \connective{=>} & \nonumber \\
 & \tab \tab \tab ( \predicate{attribute} \; \variable{Y} \; \constant{Teacher} ) & \nonumber \\
 & \tab \tab \tab ( \connective{exists} ( \variable{X} ) & \nonumber \\
 & \tab \tab \tab \tab ( \connective{and} & \nonumber \\
 & \tab \tab \tab \tab \tab ( \predicate{\$instance} \; \variable{X} \; \constant{EducationalProcess} ) & \nonumber \\
 & \tab \tab \tab \tab \tab ( \predicate{agent} \; \variable{X} \; \variable{Y} ) ) ) ) ) ) & \nonumber
\end{flalign}
\end{footnotesize}
\hspace{-5pt}The corresponding falsity-test states that either some instance of the \SUMO{} class \SUMOClass{EducationalProcess} exists where its agent does not have \SUMOIndividualAttribute{Teacher} as an attribute or some \SUMO{} object exists with \SUMOIndividualAttribute{Teacher} as an attribute that it is not the agent of any \SUMOClass{EducationalProcess}.

\begin{figure}[t]
\centering
\begin{tikzpicture}[>=triangle 60]
\matrix[matrix of math nodes,column sep={-5pt},row sep={50pt,between origins},nodes={asymmetrical rectangle}] (s)
{
|[name=verbMappingSuperClass]| [ \equivalenceMappingTikZOfConcept{\SUMOClassTikZ{Cooling}} ] : & |[name=verb]| \langle \synsetTikZ{cool}{1}{v} \rangle & \hspace{90pt} & |[name=noun]| \langle \synsetTikZ{cooler}{1}{n} \rangle & |[name=nounMappingSuperClass]| : [ \subsumptionMappingTikZOfConcept{\SUMOClassTikZ{Refrigerator}} ] \\[-35pt]
 & & \hspace{-10pt} \rotatebox{-11}{[$\SUMOIndividualRelationTikZ{instrument}$]?} & & \\[-10pt]
 & & & & |[name=nounMapping]| [ Y ] \\
};
\draw[-To,dotted] (nounMapping) -- (nounMappingSuperClass);
\draw[-latex]	(verb) -- node[auto] {\(\langle instrument \rangle\)} (noun);
\draw[-To,dotted] (verbMappingSuperClass.south) -- (nounMapping);
\end{tikzpicture}
\caption{Process pattern: verb and noun synsets connected by \equivalenceMappingRelation{} and \subsumptionMappingRelation{}/\instanceMappingRelation{} respectively}
\label{fig:ProcessPattern2}
\end{figure}
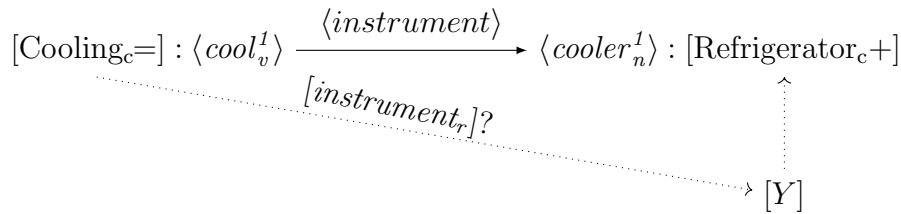

Next, we summarize the proposed question patterns and the number of problems that result from them. However, the resulting problems are organized into 3 categories ---{\it Agent}, {\it Instrument} and {\it Result}--- depending on the semantic relation with the purpose of analyzing the knowledge in \SUMO{} about each of these relations:

\begin{itemize}

\item The first question pattern focuses on relation-pairs where both synsets are connected by \equivalenceMappingRelation{}, as the example in Figure \ref{fig:ProcessPattern1}. From this pattern, we obtain 13 problems where 2 problems belong to the Agent category, 3 problems to the Instrument category and 8 problems to the Result category.

\item The second question pattern focuses on relation-pairs where the verb synset is connected by \equivalenceMappingRelation{} and the noun synset is connected by \subsumptionMappingRelation{} or \instanceMappingRelation{}, as per the example in Figure \ref{fig:ProcessPattern2}. The truth-tests of the proposed problems state that all the \SUMO{} objects that can be assigned to the verb synset have some of the \SUMO{} objects that can be assigned to the noun synset as agent/instrument/result, which corresponds to the first half of the problem proposed by the first process question pattern (see conjecture (\ref{CQ:AgentEducationalProcessTeacher})). Using this question pattern, we obtain 197 problems, from which 137, 30 and 30 problems belong respectively to the Agent, Instrument and Result categories.

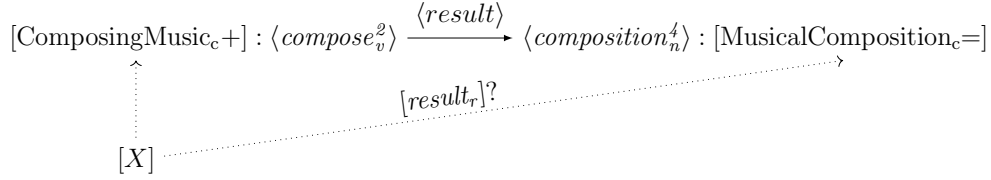
\begin{figure}[t]
\centering
\resizebox{\columnwidth}{!}{
\begin{tikzpicture}[>=triangle 60]
\matrix[matrix of math nodes,column sep={-5pt},row sep={50pt,between origins},nodes={asymmetrical rectangle},ampersand replacement=\&] (s)
{
|[name=verbMappingSuperClass]| [ \subsumptionMappingTikZOfConcept{\SUMOClassTikZ{ComposingMusic}} ] : \& |[name=verb]| \langle \synsetTikZ{compose}{2}{v} \rangle \& \hspace{50pt} \& |[name=noun]| \langle \synsetTikZ{composition}{4}{n} \rangle \& |[name=nounMappingSuperClass]| : [ \equivalenceMappingTikZOfConcept{\SUMOClassTikZ{MusicalComposition}} ] \\[-20pt]
 \& \& \hspace{-10pt} \rotatebox{7}{[$\SUMOIndividualRelationTikZ{result}$]?} \& \& \\[-25pt]
|[name=verbMapping]| [ X ] \& \& \& \& \\
};
\draw[-To,dotted] (verbMapping) -- (verbMappingSuperClass);
\draw[-latex]	(verb) -- node[auto] {\(\langle result \rangle\)} (noun);
\draw[-To,dotted] (verbMapping) -- (nounMappingSuperClass.south);
\end{tikzpicture}
}
\caption{Process pattern: verb and noun synsets connected by \subsumptionMappingRelation{}/\instanceMappingRelation{} and \equivalenceMappingRelation{} respectively}
\label{fig:ProcessPattern3}
\end{figure}

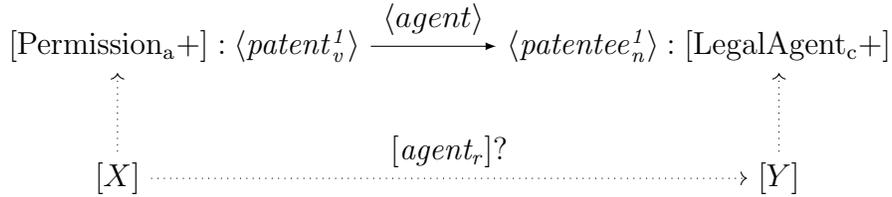
\begin{figure}[t]
\centering
\begin{tikzpicture}[>=triangle 60]
\matrix[matrix of math nodes,column sep={-5pt},row sep={70pt,between origins},nodes={asymmetrical rectangle}] (s)
{
|[name=verbMappingSuperClass]| [ \subsumptionMappingTikZOfConcept{\SUMOIndividualAttributeTikZ{Permission}} ] : & |[name=verb]| \langle \synsetTikZ{patent}{1}{v} \rangle & \hspace{50pt} & |[name=noun]| \langle \synsetTikZ{patentee}{1}{n} \rangle & |[name=nounMappingSuperClass]| : [ \subsumptionMappingTikZOfConcept{\SUMOClassTikZ{LegalAgent}} ] \\[-20pt]
|[name=verbMapping]| [ X ] & & & & |[name=nounMapping]| [ Y ] \\
};
\draw[-To,dotted] (verbMapping) -- (verbMappingSuperClass);
\draw[-To,dotted] (nounMapping) -- (nounMappingSuperClass);
\draw[-latex]	(verb) -- node[auto] {\(\langle agent \rangle\)} (noun);
\draw[-To,dotted] (verbMapping) -- node[auto] {[$\SUMOIndividualRelationTikZ{agent}$]?} (nounMapping);
\end{tikzpicture}
\caption{Process pattern: verb and noun synsets connected by \subsumptionMappingRelation{}/\instanceMappingRelation{}}
\label{fig:ProcessPattern4}
\end{figure}

\item The third question pattern is focused on relation-pairs where the verb synset is connected by \subsumptionMappingRelation{} or \instanceMappingRelation{}, while the noun synset is connected by \equivalenceMappingRelation{}, as the example in Figure \ref{fig:ProcessPattern3}. In this case, the truth-tests of the proposed problems state that all of the \SUMO{} objects that can be assigned to the noun synset are the agent/instrument/result of some \SUMO{} object that can be assigned to the verb synset, which corresponds to the second half of the problem proposed by the first process question pattern (see conjecture (\ref{CQ:AgentEducationalProcessTeacher})). Using this third question pattern, we obtain 137 problems, from which 27, 28 and 82 problems belong respectively to the Agent, Instrument and Result categories.

\item The last question pattern is focused on relation-pairs where both the verb and noun synsets are connected by \subsumptionMappingRelation{} or \instanceMappingRelation{}, such as the example in Figure \ref{fig:ProcessPattern4}. The truth-tests of the proposed problems state that some of the \SUMO{} objects that can be assigned to the verb synset have some of the \SUMO{} objects that can be assigned to the noun synset as agent/instrument/result. From this question pattern, we obtain 1,618 problems, from which 663 problems belong to the Agent category, 287 problems to the Instrument category and 668 problems to the Result category.

\end{itemize}
In total, we obtain 829 problems for the Agent category, 348 problems for the Instrument category and 788 problems for the Result category.

\begin{sidewaystable}
\centering
\resizebox{\columnwidth}{!}{
\begin{tabular}{ll;{2.5pt/2.5pt}rrrr;{2.5pt/2.5pt}rrrr;{2.5pt/2.5pt}rrrr}
\hline
\multicolumn{2}{c;{2.5pt/2.5pt}}{\multirow{2}{*}{Problem category}} & \multicolumn{4}{c;{2.5pt/2.5pt}}{{\bf \TPTPSUMO{} v5.3.0}} & \multicolumn{4}{c;{2.5pt/2.5pt}}{{\bf \ADIMENSUMO{} v2.2}} & \multicolumn{4}{c}{{\bf \ADIMENSUMO{} v2.6}} \\
\multicolumn{2}{c;{2.5pt/2.5pt}}{\multirow{2}{*}{}} & \multicolumn{1}{c}{\#} & \multicolumn{1}{c}{\%} & \multicolumn{1}{c}{T} & \multicolumn{1}{c;{2.5pt/2.5pt}}{E} & \multicolumn{1}{c}{\#} & \multicolumn{1}{c}{\%} & \multicolumn{1}{c}{T} & \multicolumn{1}{c;{2.5pt/2.5pt}}{E} & \multicolumn{1}{c}{\#} & \multicolumn{1}{c}{\%} & \multicolumn{1}{c}{T} & \multicolumn{1}{c}{E}\\
\hline
\multicolumn{2}{l;{2.5pt/2.5pt}}{{\bf Truth-tests}} & & & & & & & & & & & & \\
 & Multiple Mapping (151) & 0 & -~~ & - s. & - & 0 & -~~ & - s. & - & 23 & 15.23\% & 81.72 s. & 0.13 \\
 & Event \#1 (24) & 0 & -~~ & - s. & - & 0 & -~~ & - s. & - & 0 & -~~ & - s. & - \\
 & Event \#2 (350) & 82 & 23.43\% & 68.09 s. & 0.94 & 83 & 23.71\% & 0.66 s. & 5.46 & 83 & 23.71\% & 0.54 s. & 3.85 \\
 & Event \#3 (2,011) & 580 & 28.84\% & 20.73 s. & 0.62 & 580 & 28.84\% & 0.97 s. & 5.49 & 582 & 28.94\% & 1.67 s. & 2.76 \\
\hdashline[2.5pt/2.5pt]
\multicolumn{2}{l;{2.5pt/2.5pt}}{Mapping (2,536)} & 662 & 26.10\% & 26.60 s. & 0.66 & 663 & 26.14\% & 0.93 s. & 5.48 & 688 & 27.13\% & 4.21 s. & 2.78 \\
\hdashline[2.5pt/2.5pt]
 & Antonym \#1 (71) & 12 & 16.90\% & 4.55 s. & 1.93 & 24 & 33.80\% & 6.87 s. & 2.59 & 44 & 61.97\% & 103.42 s. & 1.16 \\
 & Antonym \#2 (489) & 66 & 13.50\% & 103.90 s. & 0.27 & 133 & 27.20\% & 22.82 s. & 0.15 & 193 & 39.47\% & 77.22 s. & 0.05 \\
 & Antonym \#3 (2,444) & 83 & 3.40\% & 125.71 s. & 0.05 & 149 & 6.10\% & 56.95 s. & 0.23 & 686 & 28.07\% & 46.36 s. & 0.08 \\
 & Agent (829) & 4 & 0.48\% & 62.22 s. & 0.24 & 7 & 0.84\% & 119.55 s. & 3.17 & 39 & 4.70\% & 6.28 s. & 0.49 \\
 & Instrument (348) & 1 & 0.29\% & 236.39 s. & 0.00 & 1 & 0.29\% & 3.60 s. & 0.28 & 61 & 17.53\% & 45.61 s. & 0.23 \\
 & Result (788) & 4 & 0.51\% & 332.67 s. & 0.01 & 11 & 1.40\% & 124.05 s. & 1.06 & 94 & 11.93\% & 11.04 s. & 0.29 \\
\hdashline[2.5pt/2.5pt]
\multicolumn{2}{l;{2.5pt/2.5pt}}{Competency (4,969)} & 170 & 3.42\% & 112.72 s. & 0.27 & 325 & 6.54\% & 42.74 s. & 0.46 & 1,117 & 22.48\% & 49.53 s. & 0.10 \\
\hdashline[2.5pt/2.5pt]
\multicolumn{2}{l;{2.5pt/2.5pt}}{Total (7,505)} & 832 & 11.09\% & 44.19 s. & 0.58 & 988 & 13.16\% & 14.68 s. & 3.83 & 1,805 & 24.05\% & 32.25 s. & 1.12 \\
\hline
\multicolumn{2}{l;{2.5pt/2.5pt}}{{\bf Falsity-tests}} & & & & & & & & & & & & \\
 & Multiple Mapping (151) & 1 & 0.66\% & 233.92 s. & 0.00 & 3 & 1.99\% & 15.08 s. & 0.07 & 2 & 1.32\% & 230.74 s. & 0.59 \\
 & Event \#1 (24) & 0 & -~~ & - s. & - & 1 & 4.17\% & 17.73 s. & 0.06 & 7 & 29.17\% & 42.40 s. & 0.02 \\
 & Event \#2 (350) & 0 & -~~ & - s. & - & 27 & 7.71\% & 33.13 s. & 0.04 & 131 & 37.43\% & 36.53 s. & 0.14 \\
 & Event \#3 (2,011) & 0 & -~~ & - s. & - & 0 & -~~ & - s. & - & 646 & 32.12\% & 22.27 s. & 0.53 \\
\hdashline[2.5pt/2.5pt]
\multicolumn{2}{l;{2.5pt/2.5pt}}{Mapping (2,536)} & 1 & 0.04\% & 233.92 s. & 0.00 & 31 & 1.22\% & 30.89 s. & 0.04 & 786 & 30.99\% & 25.36 s. & 0.46 \\
\hdashline[2.5pt/2.5pt]
 & Antonym \#1 (71) & 0 & -~~ & - s. & - & 1 & 1.41\% & 4.35 s. & 0.23 & 4 & 5.63\% & 3.66 s. & 0.28 \\
 & Antonym \#2 (489) & 25 & 5.11\% & 16.38 s. & 2.06 & 23 & 4.70\% & 25.36 s. & 5.10 & 21 & 4.29\% & 0.27 s. & 4.08 \\
 & Antonym \#3 (2,444) & 13 & 0.53\% & 23.04 s. & 0.04 & 14 & 0.57\% & 16.91 s. & 0.06 & 1 & 0.04\% & 68.91 s. & 0.01 \\
 & Agent (829) & 5 & 0.60\% & 205.67 s. & 0.01 & 2 & 0.24\% & 268.67 s. & 0.03 & 3 & 0.36\% & 402.85 s. & 0.03 \\
 & Instrument (348) & 0 & -~~ & - s. & - & 2 & 0.57\% & 15.70 s. & 0.06 & 1 & 0.29\% & 595.03 s. & 0.00 \\
 & Result (788) & 3 & 0.38\% & 249.40 s. & 0.00 & 12 & 1.52\% & 49.04 s. & 0.07 & 11 & 1.40\% & 186.29 s. & 0.28 \\
\hdashline[2.5pt/2.5pt]
\multicolumn{2}{l;{2.5pt/2.5pt}}{Competency (4,969)} & 46 & 0.93\% & 54.03 s. & 1.13 & 54 & 1.09\% & 36.70 s. & 2.21 & 41 & 0.83\% & 96.15 s. & 2.35 \\
\hdashline[2.5pt/2.5pt]
\multicolumn{2}{l;{2.5pt/2.5pt}}{Total (7,505)} & 47 & 0.63\% & 57.86 s. & 1.11 & 85 & 1.13\% & 34.58 s. & 1.42 & 827 & 11.02\% & 30.57 s. & 0.55 \\
\hline
\multicolumn{2}{l;{2.5pt/2.5pt}}{{\bf Total (15,010)}} & {\bf 879} & {\bf 5.85\%} & {\bf 44.92 s.} & {\bf 0.61} & {\bf 1,073} & {\bf 7.14\%} & {\bf 16.26 s.} & {\bf 3.64} & {\bf 2,632} & {\bf 17.53\%} & {\bf 31.19 s.} & {\bf 0.94} \\
\hline
\end{tabular}
}
\caption{\label{table:CompetencyComparison} Evaluating the competency of \SUMO{} ontologies using Vampire v3.0}
\end{sidewaystable}

\section{Experimentation} \label{section:experimentation}

Based on the set of CQs proposed in the previous sections, we have performed several experiments in order to evaluate the competency of \SUMO{} based ontologies, the mapping between \SUMO{} and \WORDNET{}, and the performance of FOL ATPs by following the methodology described in Section \ref{section:methodology}. In this experimentation, we have used an Intel\textregistered~Xeon\textregistered~CPU E5-2640v3@2.60GHz with 2GB of RAM per processor. For each CQ, we provide an ontology and the given conjecture as input to the ATP system. In the following subsections, we report on the results of these experiments.\footnote{The \ADIMENSUMO{} ontology, the set of CQs and all the execution reports are freely available at \url{http://adimen.si.ehu.es}.} Additionally, we have manually analyzed some of the tests in order to evaluate the proposed CQs, as reported in the last subsection.

\subsection{Evaluating the competency of \SUMO{} based ontologies}

In this subsection, we report on the evaluation of the competency of \TPTPSUMO{} and \ADIMENSUMO{}. In the case of \ADIMENSUMO{}, we also evaluate the improvement between two different versions: \ADIMENSUMO{} v2.2, which is the first version we proposed, and \ADIMENSUMO{} v2.6.

Table \ref{table:CompetencyComparison} summarizes some results of the ATP Vampire v3.0 when evaluating \TPTPSUMO{} and \ADIMENSUMO{} (v2.2 and v2.6) with an execution time limited to 600 seconds. The selection of Vampire v3.0 is due to the fact that it is the most successful ATP system in the experimentation reported in \cite{ALR16} when using the set of CQs proposed in \cite{ALR15} for the evaluation of ATPs. The execution time limit is set to 600 seconds since it is the longest time limit that has been ever used in CASC and we have obtained good practical results using 600 seconds as time limit in our preliminary experiments. CQs have been organized in two main divisions: {\it truth-tests} and {\it falsity-tests}. In addition, each division is organized according the two main problem categories introduced in the previous sections (and their corresponding subcategories): {\it Mapping}, for the validation of the mapping, and {\it Competency}, for the evaluation of the knowledge in the ontologies. For each ontology and each problem (sub)category (with the total number of problems between brackets), we provide the number (\# column) and percentage (\% column) of CQs that are proved together with the average run time (T column) and the efficiency measure that is used in CASC. This efficiency measure balances the time taken for each problem solved against the number of problems solved and it is calculated as the average of the inverses of the times for problems solved.

With respect to the {\it Mapping} categories, \ADIMENSUMO{} v2.6 slightly outperforms \TPTPSUMO{} and \ADIMENSUMO{} v2.2 in the truth-test division---more passing tests (688 compared to 662 and 663)--- because there is almost no difference in the Event subcategories. On the contrary, no proof is found for the truth-test Multiple Mapping category using \TPTPSUMO{} or \ADIMENSUMO{} v2.2, while 23 truth-tests can be classified as passing using \ADIMENSUMO{} v2.6. In the case of falsity-tests, the difference is clearly larger: 786 non-passing tests using \ADIMENSUMO{} v2.6 compared to 1 non-passing test using \TPTPSUMO{} and 31 non-passing tests using \ADIMENSUMO{} v2.2. This result reveals that \ADIMENSUMO{} v2.6 enables the detection of many defects in the mapping information which are not discovered using \TPTPSUMO{} or \ADIMENSUMO{} v2.2.

The results of the {\it Competency} categories shows that \ADIMENSUMO{} v2.6 is the most competent ontology. It outperforms \TPTPSUMO{} and \ADIMENSUMO{} v2.2 in terms of competency in both the truth-tests ---more passing tests (1,117 compared to 170 and 325), since conjectures are expected to be entailed--- and the falsity-test divisions ---less non-passing tests (41 compared to 46 and 54), since conjectures are expected not to be entailed. Further, \ADIMENSUMO{} v2.6 is by far the most competent ontology in all the {\it Competency} subcategories of the truth-test division. On the contrary, \TPTPSUMO{} is the less competent ontology since \ADIMENSUMO{} v2.2 clearly outperforms \TPTPSUMO{} in all the {\it Competency} subcategories of the truth-test division and the difference in the falsity-test division is not relevant.
 
\begin{sidewaystable}
\centering
\resizebox{\columnwidth}{!}{
\begin{tabular}{lrrrrrrrrrrrrrrr}
\hline
\multicolumn{1}{c}{\multirow{2}{*}{{\bf Problem category}}} & \multicolumn{3}{c}{{\bf VP v2.6}} & \multicolumn{3}{c}{{\bf VP v3.0}} & \multicolumn{3}{c}{{\bf VP v4.0}} & \multicolumn{3}{c}{{\bf VP v4.1}} & \multicolumn{3}{c}{{\bf EP v2.0}} \\
\multicolumn{1}{c}{\multirow{2}{*}{}} & \multicolumn{1}{c}{\#} & \multicolumn{1}{c}{T} & \multicolumn{1}{c}{E} & \multicolumn{1}{c}{\#} & \multicolumn{1}{c}{T} & \multicolumn{1}{c}{E} & \multicolumn{1}{c}{\#} & \multicolumn{1}{c}{T} & \multicolumn{1}{c}{E} & \multicolumn{1}{c}{\#} & \multicolumn{1}{c}{T} & \multicolumn{1}{c}{E} & \multicolumn{1}{c}{\#} & \multicolumn{1}{c}{T} & \multicolumn{1}{c}{E} \\
\hline
Antonym \#1 (71) & 44 & 68.32 s. & 2.06 & 44 & 103.42 s. & 1.16 & 37 & 68.15 s. & 0.82 & 43 & 30.07 s. & 0.27 & 40 & 35.01 s. & 0.31 \\
Antonym \#2 (489) & 204 & 86.84 s. & 0.13 & 193 & 77.22 s. & 0.05 & 54 & 130.27 s. & 0.05 & 119 & 144.33 s. & 0.11 & 150 & 145.10 s. & 0.03 \\
Antonym \#3 (2,444) & 1,086 & 182.29 s. & 0.15 & 686 & 46.36 s. & 0.07 & 431 & 60.32 s. & 0.06 & 851 & 194.74 s. & 0.10 & 256 & 324.63 s. & 0.01 \\
Agent (829) & 43 & 14.93 s. & 0.46 & 39 & 6.28 s. & 0.49 & 10 & 22.94 s. & 0.09 & 23 & 318.42 s. & 0.17 & 12 & 394.34 s. & 0.13 \\
Instrument (348) & 61 & 3.36 s. & 0.33 & 61 & 45.61 s. & 0.23 & 25 & 64.65 s. & 0.10 & 26 & 404.10 s. & 0.02 & 2 & 381.84 s. & 0.01 \\
Result (788) & 118 & 4.46 s. & 0.35 & 94 & 11.04 s. & 0.29 & 52 & 74.86 s. & 0.09 & 64 & 294.91 s. & 0.12 & 39 & 418.10 s. & 0.01 \\
\hdashline[2.5pt/2.5pt]
Truth-tests (4,969) & 1,556 & 141.43 s. & 0.19 & 1,117 & 49.53 s. & 0.10 & 609 & 67.80 s. & 0.12 & 1,126 & 196.18 s. & 0.11 & 499 & 256.66 s. & 0.05 \\
\hdashline[2.5pt/2.5pt]
Multiple Mapping (151) & 3 & 271.55 s. & 0.52 & 2 & 230.74 s. & 0.59 & 2 & 55.36 s. & 0.02 & 3 & 85.13 s. & 0.02 & 0 & - s. & - \\
Event \#1 (24) & 5 & 128.66 s. & 0.35 & 7 & 42.40 s. & 0.02 & 3 & 388.45 s. & 0.00 & 4 & 250.19 s. & 0.01 & 5 & 357.09 s. & 0.00 \\
Event \#2 (350) & 117 & 58.43 s. & 0.15 & 131 & 36.53 s. & 0.14 & 38 & 173.34 s. & 0.04 & 41 & 88.12 s. & 0.09 & 52 & 306.44 s. & 0.00 \\
Event \#3 (2,011) & 646 & 35.05 s. & 0.73 & 646 & 22.27 s. & 0.53 & 104 & 120.23 s. & 0.09 & 83 & 62.00 s. & 0.02 & 190 & 271.18 s. & 0.00 \\
\hdashline[2.5pt/2.5pt]
Falsity-tests (2,536) & 771 & 40.12 s. & 0.63 & 786 & 25.36 s. & 0.46 & 147 & 136.56 s. & 0.08 & 131 & 76.45 s. & 0.05 & 247 & 280.34 s. & 0.00 \\
\hline
{\bf Total (7,505)} & {\bf 2,327} & {\bf 107.86 s.} & {\bf 0.30} & {\bf 1,903} & {\bf 39.54 s.} & {\bf 0.19} & {\bf 756} & {\bf 81.56 s.} & {\bf 0.11} & {\bf 1,257} & {\bf 183.70 s.} & {\bf 0.11} & {\bf 746} & {\bf 264.50 s.} & {\bf 0.03} \\
\hline
\end{tabular}
}
\caption{\label{table:ATPComparison} Evaluating the performance of FOL ATPs}
\end{sidewaystable}

Regarding efficiency, \ADIMENSUMO{} v2.2 is the most efficient ontology and \TPTPSUMO{} the least efficient one according the efficiency values: 3.64 (\ADIMENSUMO{} v2.2), 0.94 (\ADIMENSUMO{} v2.6) and 0.61 (\TPTPSUMO{}). The fact that \ADIMENSUMO{} v2.2 outperforms \ADIMENSUMO{} v2.6 in terms of efficiency is not surprising since \ADIMENSUMO{} v2.2 encodes less knowledge and is less competent that \ADIMENSUMO{} v2.6. Similarly, the average run times with \ADIMENSUMO{} are in general shorter than those with \TPTPSUMO{}, especially in the truth-test division: 14.68 s. (\ADIMENSUMO{} v2.2) and 32.25 s. (\ADIMENSUMO{} v2.6) against 44.19 s. (\TPTPSUMO{}). At the same time, the average run times with \ADIMENSUMO{} v2.6 are longer than the ones with \ADIMENSUMO{} v2.2. These facts lead us to think that the problems that are only solved using \ADIMENSUMO{} v2.6 require complex and long proofs and, additionally, it also confirms the improvement of \ADIMENSUMO{} v2.6 in terms of competency.

\subsection{Evaluating the performance of FOL ATPs}

According to the results reported in the previous section, most of the truth-tests belonging to the {\it Mapping} categories are solved in less than 2 seconds: more concretely, all the proved truth-tests belonging the Event category. Furthermore, in additional preliminary experiments using the remaining ATPs, we check that most of the proofs were trivial ---just involving 2 or 3 axioms--- and that no significant differences between the considered ATPs. Consequently, we conclude that those CQs do not enable a suitable evaluation of ATPs. Additionally, very few falsity-tests belonging to the {\it Competency} categories are proved (less than 1\% of CQs). Further, the remaining ones are not likely to be entailed by \ADIMENSUMO{}, as confirmed in Subsection \ref{subsection:evaluatingAdimenSUMO}. Therefore, we concentrate on the {\it Competency} categories of the truth-test division and the {\it Mapping} categories of the falsity-test division.

In Table \ref{table:ATPComparison}, we summarize some figures from the evaluation of the different versions of Vampire (VP)\footnote{Using the following parameters: \tt{--proof tptp --output\_axiom\_names on --mode casc -t 600 -m 2048}.} and E (EP)\footnote{Using the following parameters: \tt{--auto --proof-object -s --cpu-limit=600 --memory-limit=2048 }.} introduced in Subsection \ref{subsection:ATPs} using \ADIMENSUMO{} v2.6. For each ATP, we provide the number of proofs (\# column), the average run times (T column) and the CASC efficiency measure (E column) in each problem subcategory.

Globally, Vampire v2.6 is the most effective ATP according to the total number of proofs (2,327 proofs) with a difference of 424 and 1,070 proofs to Vampire v3.0 (second place) and Vampire v4.1 (third place) respectively. This result is different from our preliminary evaluation of ATPs reported in \cite{ALR16}. In that evaluation, Vampire v3.0 was the most effective ATP and Vampire v2.6 obtained nearly the same number of proofs for the set of CQs proposed in \cite{ALR15}, which is different from the set of CQs introduced in this work. With respect to the remaining ATPs (Vampire v4.0 and E v2.0), the number of proofs is clearly smaller.

Regarding each division and problem subcategory, Vampire v2.6 is the winner in the truth-test division (1,556 proofs) and in all the truth-test problem subcategories. On the contrary, Vampire v3.0 is the winner in the falsity-test division (786 proofs) and in all the Antonym problem subcategories. In the case of the falsity-test division, the differences between the two most effective ATPs (Vampire v3.0 and v2.6) are smaller than the differences between the two most effective ATPs in the truth-test division (Vampire v2.6 and v4.1), but the difference between the two most effective ATPs and the remaining ones is clearly larger.

The analysis of efficiency is more disparate:
\begin{itemize}
\item According to the CASC efficiency measure, Vampire v2.6 is also the most efficient in all the divisions and (sub)categories followed by Vampire v3.0, although Vampire v4.0 and Vampire v4.1 outperform Vampire v3.0 in the truth-test division. On the contrary, E is the less efficient ATP in both the truth- and the falsity-test divisions.
\item Vampire v3.0 is the ATP with the lowest average run time (39.54 s.) followed by Vampire v4.0 (81.56 s.) and Vampire v2.6 (107.86 s.). However, Vampire v4.0 proves fewer CQs in comparison with Vampire v2.6, which in general is faster than Vampire v4.0 except for the third problem subcategory of Antonym (182.29 s. compared to 60.32 s.) and the Multiple Mapping category (271.55 s. compared to 55.36 s.).
\item Vampire v4.0 is more efficient (lower average run time and higher efficiency value) than Vampire v4.1 and E v2.0, although Vampire v4.1 and E v2.0 are the two fastest systems in the first subcategory of Antonym problems (Antonym \#1), and Vampire v4.1 also performs faster than Vampire v4.0 in all the Event subcategories of the falsity-test division.
\end{itemize}
To sum up, we can conclude that our set of proposed CQs is really heterogeneous, enabling the evaluation of a wide range of features of state-of-the-art ATPs.

\subsection{Evaluating \ADIMENSUMO{} v2.6 and its Mapping from \WORDNET{}} \label{subsection:evaluatingAdimenSUMO}

\begin{sidewaystable}
\centering
\resizebox{\columnwidth}{!}{
\begin{tabular}{ll;{2.5pt/2.5pt}rrrr;{2.5pt/2.5pt}rrrrr;{2.5pt/2.5pt}rrrr}
\hline
\multicolumn{2}{c;{2.5pt/2.5pt}}{\multirow{2}{*}{{\bf Problem category}}} & \multicolumn{4}{c;{2.5pt/2.5pt}}{{\bf Proofs}} & \multicolumn{5}{c;{2.5pt/2.5pt}}{{\bf Coverage}} & \multicolumn{4}{c}{{\bf Difficulty}} \\
\multicolumn{2}{c;{2.5pt/2.5pt}}{\multirow{2}{*}{}} & \multicolumn{1}{c}{\#} & \multicolumn{1}{c}{\%} & \multicolumn{1}{c}{T} & \multicolumn{1}{c;{2.5pt/2.5pt}}{E} & \multicolumn{1}{c}{{\bf N}} & \multicolumn{1}{c}{{\bf P}} & \multicolumn{1}{c}{{\bf S}} & \multicolumn{1}{c}{{\bf C}} & \multicolumn{1}{c;{2.5pt/2.5pt}}{{\bf F}} & \multicolumn{1}{c}{{\bf D}} & \multicolumn{1}{c}{{\bf N}} & \multicolumn{1}{c}{{\bf C}} & \multicolumn{1}{c}{{\bf F}} \\
\hline
\multicolumn{2}{l;{2.5pt/2.5pt}}{{\bf Truth-tests}} & & & & & & & & & & & & & \\
 & Multiple Mapping (151) & 27 & 17.88\% & 134.69 s. & 0.17 & 131 & 1.76\% & 23 & 106 & 25 & 0.48 & 14.44 & 8.19 & 6.26 \\
 & Event \#1 (24) & 0 & 0.00\% & - s. & - & - & -~~~ & - & - & - & - & - & - & - \\
 & Event \#2 (350) & 108 & 30.86\% & 0.26 s. & 4.21 & 196 & 2.64\% & 7 & 176 & 20 & 0.00 & 5.32 & 3.27 & 2.06 \\
 & Event \#3 (2,011) & 582 & 28.94\% & 0.30 s. & 3.69 & 380 & 5.11\% & 88 & 378 & 2 & 0.00 & 1.78 & 1.32 & 0.45 \\
\hdashline[2.5pt/2.5pt]
\multicolumn{2}{l;{2.5pt/2.5pt}}{Mapping (2,536)} & 717 & 28.27\% & 5.35 s. & 3.61 & 606 & 8.15\% & 118 & 565 & 41 & 0.02 & 2.79 & 1.88 & 0.91 \\
\hdashline[2.5pt/2.5pt]
 & Antonym \#1 (71) & 44 & 61.97\% & 20.06 s. & 2.19 & 94 & 1.26\% & 26 & 67 & 27 & 0.05 & 4.14 & 1.77 & 2.36 \\
 & Antonym \#2 (489) & 233 & 47.65\% & 65.27 s. & 0.13 & 601 & 8.08\% & 40 & 432 & 169 & 0.45 & 11.39 & 5.79 & 5.60 \\
 & Antonym \#3 (2,444) & 1,167 & 47.75\% & 110.16 s. & 0.20 & 1,601 & 21.53\% & 652 & 1,121 & 480 & 0.44 & 13.91 & 8.86 & 5.05 \\
 & Agent (829) & 43 & 5.19\% & 14.90 s. & 0.57 & 144 & 1.94\% & 14 & 102 & 42 & 0.42 & 11.46 & 6.53 & 4.93 \\
 & Instrument (348) & 61 & 17.53\% & 3.34 s. & 0.39 & 161 & 2.16\% & 39 & 120 & 41 & 0.42 & 12.23 & 7.25 & 4.98 \\
 & Result (788) & 118 & 14.97\% & 3.47 s. & 0.44 & 281 & 3.78\% & 32 & 200 & 81 & 0.43 & 11.49 & 6.30 & 5.19 \\
\hdashline[2.5pt/2.5pt]
\multicolumn{2}{l;{2.5pt/2.5pt}}{Competency (4,969)} & 1,666 & 33.53\% & 87.58 s. & 0.22 & 1,822 & 24.50\% & 803 & 1,266 & 556 & 0.43 & 13.00 & 7.94 & 5.06 \\
\hdashline[2.5pt/2.5pt]
\multicolumn{2}{l;{2.5pt/2.5pt}}{Total (7,505)} & 2,383 & 31.75\% & 63.47 s. & 1.24 & 1,989 & 26.74\% & 921 & 1,431 & 558 & 0.31 & 9.93 & 6.12 & 3.81 \\
\hline
\multicolumn{2}{l;{2.5pt/2.5pt}}{{\bf Falsity-tests}} & & & & & & & & & & & & \\
 & Multiple Mapping (151) & 3 & 1.99\% & 69.47 s. & 0.54 & 31 & 0.42\% & 1 & 20 & 11 & 0.33 & 11.33 & 6.67 & 4.67 \\
 & Event \#1 (24) & 8 & 33.33\% & 176.04 s. & 0.27 & 93 & 1.25\% & 5 & 70 & 23 & 0.42 & 16.25 & 10.13 & 6.13 \\
 & Event \#2 (350) & 131 & 37.43\% & 47.73 s. & 0.20 & 381 & 5.12\% & 45 & 305 & 76 & 0.41 & 14.35 & 8.16 & 6.19 \\
 & Event \#3 (2,011) & 646 & 32.12\% & 31.22 s. & 0.76 & 466 & 6.27\% & 49 & 401 & 65 & 0.47 & 14.99 & 8.83 & 6.16 \\
\hdashline[2.5pt/2.5pt]
\multicolumn{2}{l;{2.5pt/2.5pt}}{Mapping (2,536)} & 788 & 25.45\% & 35.44 s. & 0.66 & 659 & 8.86\% & 100 & 541 & 118 & 0.46 & 14.89 & 8.73 & 6.16 \\
\hdashline[2.5pt/2.5pt]
 & Antonym \#1 (71) & 4 & 5.63\% & 1.88 s. & 0.77 & 30 & 0.40\% & 0 & 18 & 12 & 0.40 & 8.25 & 5.00 & 3.25 \\
 & Antonym \#2 (489) & 23 & 4.70\% & 50.94 s. & 3.59 & 33 & 0.44\% & 0 & 20 & 13 & 0.43 & 2.96 & 1.57 & 1.39 \\
 & Antonym \#3 (2,444) & 4 & 0.16\% & 401.56 s. & 1.79 & 47 & 0.63\% & 6 & 24 & 23 & 0.80 & 23.50 & 11.50 & 12.00 \\
 & Agent (829) & 6 & 0.72\% & 273.96 s. & 0.01 & 53 & 0.71\% & 3 & 21 & 32 & 0.57 & 14.67 & 5.00 & 9.67 \\
 & Instrument (348) & 1 & 0.29\% & 385.10 s. & 0.00 & 15 & 0.20\% & 4 & 5 & 10 & 0.60 & 15.00 & 5.00 & 10.00 \\
 & Result (788) & 21 & 2.66\% & 286.12 s. & 0.16 & 121 & 1.63\% & 10 & 62 & 59 & 0.60 & 18.57 & 6.29 & 12.29 \\
\hdashline[2.5pt/2.5pt]
\multicolumn{2}{l;{2.5pt/2.5pt}}{Competency (4,969)} & 59 & 1.19\% & 183.44 s. & 1.68 & 199 & 2.68\% & 23 & 111 & 88 & 0.53 & 11.66 & 4.56 & 7.10 \\
\hdashline[2.5pt/2.5pt]
\multicolumn{2}{l;{2.5pt/2.5pt}}{Total (7,505)} & 847 & 11.29\% & 45.88 s. & 0.75 & 764 & 10.27\% & 123 & 581 & 183 & 0.47 & 14.66 & 8.44 & 6.22 \\
\hline
\multicolumn{2}{l;{2.5pt/2.5pt}}{{\bf Total (15,010)}} & {\bf 3,230} & {\bf 21.52\%} & {\bf 58.39 s.} & {\bf 1.11} & {\bf 2,149} & {\bf 28.90\%} & {\bf 1,044} & {\bf 1,542} & {\bf 607} & {\bf 0.35} & {\bf 11.16} & {\bf 6.72} & {\bf 4.44} \\
\hline
\end{tabular}
}
\caption{\label{table:AdimenSUMOEvaluation} Evaluating the competency of \ADIMENSUMO{} v2.6}
\end{sidewaystable}

Finally, we evaluate the competency of \ADIMENSUMO{} v2.6 ---which consists of 7,437 axioms: 4,638 unit clauses (atomic formulas) and 2,799 formulae (non-atomic formulas)--- and its mapping from \WORDNET{}. For this purpose, all the ATP systems introduced in the above subsection have been used individually to experiment with the entire set of 15,010 CQs and then the outputs obtained from them have been jointly analyzed. 

In Table \ref{table:AdimenSUMOEvaluation}, we report the results of this experimentation and our joint analysis. These results are organized in three main parts ---Proofs, Coverage and Difficulty---, each of them consisting of three columns. In the first part (Proofs), we provide the number (\# column) of CQs that are proved by some of the ATPs, together with its percentage (\% column) and the average run time (T column). In the second part (Coverage), we provide the following figures about the axioms that are used in some of the proofs provided by ATPs:
\begin{itemize}
\item The number (N column) and percentage (P column) of axioms that are used in some proofs.
\item The number of axioms that are exclusively used in proofs of the corresponding problem subcategory (S column).
\item The number of used unit clauses (C column) and formulae (F column).
\end{itemize}
In the last part (Difficulty), we provide some measures of how difficult it is to prove the CQs of each (sub)category:
\begin{itemize}
\item On one hand, we use the {\it problem difficulty rating} introduced in \cite{SuS01}, which is calculated as the ratio between the number of ATPs that fail to solve a conjecture ({\it failing rating contributors}) and the total number of ATPs that have been tried ({\it rating contributors}). Thus, this rating provides a value between 0 ({\it easy problems}, 0 failing contributors) and 1 (unknown of {\it difficult problems}, all the rating contributors are failing) for each CQ. In column D, we provide the average of the difficulty problem ranking for the CQs that are successfully solved by at least one ATP among the five ranking contributors. Consequently, the highest possible value in column D is 0.80, since the number of rating contributors is 5.

\item On the other hand, we report the average number of axioms (N column) that are used in each proof and the average number of unit clauses (C column) and formulae (F column). These values provide a measure about the amount and nature of knowledge that is required for solving a CQ: more concretely, the amount of explicit (unit clauses) and implicit knowledge (formulae) of the ontology that is used by ATPs.
\end{itemize}

Regarding the {\it Mapping} categories, we have proposed 2,536 problems, from which 1,505 have been successfully solved by ATPs (59.35\%). According to the results reported in Table \ref{table:AdimenSUMOEvaluation}, most of the truth-tests are easily solved by ATPs: in the second and third Event subcategories, the difficulty ratio is 0.0 and the average runtime is 0.30 seconds or smaller. On the contrary, the difficulty of the falsity-tests is comparable to the difficulty of the CQs belonging to the {\it Competency} categories. Among the solved problems, the mapping information is validated by 717 truth-tests and 788 falsity-tests enable the detection of some defects. For example, the synsets \synset{affirm}{3}{v} and \synset{affirmation}{2}{n} are related by \textPredicate{event} and respectively connected to \subsumptionMappingOfConcept{\SUMOClass{Communication}} and \equivalenceMappingOfConcept{\SUMOClass{Stating}}, from which we obtain the following truth-test:

\vspace{-\baselineskip}
\begin{footnotesize}
\begin{flalign} \label{CQ:StatingCommunication}
%
\doubletab & ( \predicate{\$subclass} \; \constant{Stating} \; \constant{Communication} ) & 
\end{flalign}
\end{footnotesize}
\hspace{-5pt}The above CQ is entailed by \ADIMENSUMO{} v2.6 since \SUMOClass{Stating} is a subclass of \SUMOClass{LinguisticCommunication}, which is in turn a subclass of \SUMOClass{Communication}. Therefore, the problem is decided to be {\it solved} and {\it entailed}, and we can conclude that the mapping of the synsets in the pair \pair{event}{\synset{affirm}{3}{v}}{\synset{affirmation}{2}{n}} is validated according to our criteria. On the contrary, we detect some defects in the mapping information of the synsets in \pair{event}{\synset{represent}{14}{v}}{\synset{representation}{1}{n}} as follows. Since \synset{represent}{14}{v} and \synset{representation}{1}{n} are connected to \equivalenceMappingOfConcept{\SUMOClass{Stating}} and \equivalenceMappingOfConcept{\SUMOClass{Imagining}} respectively, we propose the following falsity-test stating that its mapping is wrong:

\vspace{-\baselineskip}
\begin{footnotesize}
\begin{flalign} \label{CQ:StatingImagining}
%
\doubletab & ( \connective{not} \\
 & \tab ( \predicate{equal} \; \constant{Stating} \; \constant{Imagining} ) ) & \nonumber
\end{flalign}
\end{footnotesize}
\hspace{-5pt}ATPs can prove that the above CQ is entailed by \ADIMENSUMO{} v2.6 since \SUMOClass{Stating} is a subclass of \SUMOClass{IntentionalProcess} and \SUMOClass{Dreaming} is a subclass of \SUMOClass{Imagining}, with \SUMOClass{IntentionalProcess} and \SUMOClass{Dreaming} being disjoint classes. Thus, the problem is decided to be {\it solved} and {\it incompatible} and it enables the detection of that the mapping information in the pair \pair{event}{\synset{represent}{14}{v}}{ \synset{representation}{1}{n}} is incorrect.

With respect to the {\it Competency} categories (4,969 problems), we can conclude that:
\begin{itemize}
\item The knowledge of \ADIMENSUMO{} and \WORDNET{} seems to be well-aligned for 33.53\% of problems (1,666 truth-tests from the Antonym and Relation categories are solved).
\item A quarter of the ontology (1,822 axioms, 24.50\% of total) has been used in the proof of the truth-tests.
\item Only 1.19\% of problems (59 falsity-tests from the Antonym and Relation categories are proved) enable the detection of some failure or misalignment in the knowledge of \ADIMENSUMO{} and \WORDNET{}, which involve a total of 199 axioms (2.68\% of the total).
\item The knowledge about antonym concepts in \WORDNET{} is better covered by \ADIMENSUMO{} than the knowledge about roles in events: 48.07\% of truth-tests from the Antonym category (1,444 proofs from 3,004 CQs) are proved against 11.30\% of truth-tests from the Relation category (222 proofs from 1,965 CQs).
\end{itemize}

In addition to evaluating the competency, {\it incompatible} (its falsity-test is classified as non-passing) and {\it unsolved} problems (both tests are classified as unknown) provide useful information to improve the ontology. For example, ATPs do not find a proof for the CQ in (\ref{CQ:BirthDeath}) (truth-test) and its negation (falsity-test). By inspecting the ontology, it is easy to check that the \SUMO{} classes \SUMOClass{Birth} and \SUMOClass{Death} are not axiomatized to be disjoint, as one would naturally expect. Thus, the problem consisting of (\ref{CQ:BirthDeath}) (truth-test) and its negation (falsity-test) enables the detection of missing knowledge in the ontology.

Further, the quality of the problems belonging to the {\it Competency} categories can be measured through the following three indicators:
\begin{itemize}
\item The average difficulty rating (D column) of all the problem subcategories in the truth-test division is at least 0.40 except for the first Antonym subcategory. In addition, the average difficulty rating is much higher (0.60 or around) in all the Result subcategories of the falsity-test division, and further the maximum possible (0.80) in the last subcategory of Antonym.
\item The average number of axioms that are used in each proof (N column of Difficulty part) is higher than 11 except for the case of the first Antonym subcategory (both divisions) and the second Antonym subcategory of falsity-test division. This implies that substantial portions of the knowledge in the ontology are required for proving these CQs.
\item The number of axioms that are used in proofs (N column of Coverage part) grows with (and it is always greater than) the number of proofs (\# column). In addition, among the 921 axioms that are exclusively used in a single problem subcategory (S column), 803 axioms correspond to the {\it Competency} categories of the truth-test division.
\end{itemize}
We conclude that the two first indicators lead us to confirm that the proofs for the problems in the Antonym and Relation categories are not trivial, while the last reveals that ATPs are not repeatedly using an small subset of axioms of the ontology for constructing the proofs.

Finally, from the results reported in Tables \ref{table:ATPComparison} and \ref{table:AdimenSUMOEvaluation} we can conclude that ATPs are able to prove different subsets of CQs. In this sense, the number of truth-tests belonging to the {\it Competency} categories that are proved by at least one of the ATPs (1,666 truth-tests) is 7,07\% larger than the number of truth-tests that are proved by Vampire v2.6 (1,556 truth-tests), which is the most effective ATP. In particular, 233 truth-tests belonging to the second Antonym subcategory are proved by some of the ATPs while each ATP at most proves 204 truth-tests. Therefore, the number of CQs entailed by \ADIMENSUMO{} could be larger than the number of proofs reported in Table \ref{table:AdimenSUMOEvaluation}. We could increase the number of proofs in our experiments by increasing the execution time and memory limit settings, by tuning the ATPs to \ADIMENSUMO{} or by trying other ATP systems.

\begin{table}[t]
\centering
\resizebox{\columnwidth}{!}{
\begin{tabular}{l;{2.5pt/2.5pt}r;{2.5pt/2.5pt}rr;{2.5pt/2.5pt}rr;{2.5pt/2.5pt}rr;{2.5pt/2.5pt}rr;{2.5pt/2.5pt}rr}
\hline
\multicolumn{1}{c;{2.5pt/2.5pt}}{\multirow{2}{*}{{\bf Problem category}}} & \multicolumn{1}{c;{2.5pt/2.5pt}}{\multirow{2}{*}{{\bf Problems}}} & \multicolumn{2}{c;{2.5pt/2.5pt}}{{\bf Mapping}} & \multicolumn{6}{c;{2.5pt/2.5pt}}{{\bf Solutions}} & \multicolumn{2}{c}{{\bf Missing solutions}} \\
\multicolumn{1}{c;{2.5pt/2.5pt}}{\multirow{2}{*}{}} & \multicolumn{1}{c;{2.5pt/2.5pt}}{\multirow{2}{*}{}} & \multicolumn{1}{c}{Correct} & \multicolumn{1}{c;{2.5pt/2.5pt}}{Incorrect} & \multicolumn{1}{c}{TT} & \multicolumn{1}{c}{FT} & \multicolumn{1}{c}{CM} & \multicolumn{1}{c}{IM} & \multicolumn{1}{c}{CK} & \multicolumn{1}{c;{2.5pt/2.5pt}}{IK} & \multicolumn{1}{c}{Knowledge} & \multicolumn{1}{c}{ATP} \\
\hline
Multiple Mapping (151) & 1 & 1 (0) & 0 & 0 & 0 & - & - & - & - & 1 & 0 \\
Event \#1 (24) & 0 & - (-) & - & - & - & - & - & - & - & - & - \\
Event \#2 (350) & 1 & 1 (0) & 0 & 0 & 0 & - & - & - & - & 1 & 0 \\
Event \#3 (2,011) & 22 & 16 (6) & 6 & 7 & 7 & 8 & 6 & 14 & 0 & 7 & 1 \\
\hdashline[2.5pt/2.5pt]
Mapping (2,536) & 24 & 18 (6) & 6 & 7 & 7 & 8 & 6 & 14 & 0 & 9 & 1 \\
\hdashline[2.5pt/2.5pt]
Antonym \#1 (71) & 2 & 2 (2) & 0 & 2 & 0 & 2 & - & 1 & 0 & - & - \\
Antonym \#2 (489) & 3 & 2 (0) & 1 & 1 & 0 & 1 & 0 & 1 & 0 & 2 & 0 \\
Antonym \#3 (2,444) & 27 & 8 (2) & 19 & 14 & 0 & 5 & 9 & 14 & 0 & 3 & 0 \\
Agent (829) & 5 & 4 (1) & 1 & 1 & 0 & 1 & 0 & 1 & 0 & 3 & 0 \\
Instrument (348) & 2 & 2 (2) & 0 & 0 & 0 & - & - & - & - & 0 & 2 \\
Result (788) & 12 & 7 (4) & 5 & 1 & 0 & 1 & 0 & 1 & 0 & 6 & 0 \\
\hdashline[2.5pt/2.5pt]
Competency (4,969) & 51 & 25 (11) & 26 & 19 & 0 & 10 & 9 & 19 & 0 & 13 & 2 \\
\hline
{\bf Total (7,505)} & {\bf 75} & {\bf 43 (17)} & {\bf 32} & {\bf 26} & {\bf 7} & {\bf 18} & {\bf 15} & {\bf 33} & {\bf 0} & {\bf 22} & {\bf 3} \\
\hline
\end{tabular}
}
\caption{\label{table:CQEvaluation} Detailed analysis of problems}
\end{table}

\subsection{A Complete Analysis of a Small Set of Problems}

As we have already described in the above subsections, the proposed set of CQs is suitable for evaluating the competency of \ADIMENSUMO{} and for detecting some mapping misalignments. These are some good indicators of the quality of the proposed CQs. However, a more detailed analysis requires a manual inspection of the conjectures, the mapping of the involved synsets, and the proofs obtained by ATPs. Thus, we have randomly selected a sample of 75 problems (1\%) following a uniform distribution.

In Table \ref{table:CQEvaluation}, we summarize some figures of our detailed analysis in four main parts ---Problems, Mapping, Solutions and Missing solutions---. In the first part (Problems), we provide the number of problems of each subcategory that have been randomly chosen. In the second part (Mapping, two columns), we provide the result of our quality analysis of the mapping between \WORDNET{} and \ADIMENSUMO{}: the number of problems where both synsets are correctly connected to \ADIMENSUMO{} (Correct column) and the number of problems such that at least one of the synsets is incorrectly connected (Incorrect column). In addition, we also provide the number of mappings where the two synsets are both correctly and {\it precisely} connected (Correct column, between brackets). Our criteria for classifying a mapping as {\it only correct} or as {\it correct and precise} are the following: on one hand, we consider a mapping as {\it correct} if the semantics associated with the \ADIMENSUMO{} concept and with the synset are compatible, and a correct mapping is also considered as {\it precise} if the semantics of the synset and the \SUMO{} concept are equivalent. For example, the semantics of the noun synset \synset{machine}{1}{n} is {\it ``Any mechanical or electrical device that transmits or modifies energy to perform or assist in the performance of human tasks''} and the semantics of the \SUMO{} class \SUMOClass{Machine} is {\it ``Machines are Devices that that have a well-defined resource and result and that automatically convert the resource into the result''}. Thus, the mapping of \synset{machine}{1}{n} to \equivalenceMapping{\SUMOClass{Machine}} is classified as correct and precise. On the contrary, the semantics of the adjective synset \synset{homemade}{1}{a} is {\it ``made or produced in the home or by yourself''} and the semantics of the \SUMO{} class \SUMOClass{Making} is {\it ``The subclass of Creation in which an individual Artifact or a type of Artifact is made''}. Hence, we classify the mapping of \synset{homemade}{1}{a} to \subsumptionMapping{\SUMOClass{Making}} as incorrect. On the other hand, we consider a mapping as {\it only correct} (that is, correct but not precise) when the semantics of the \ADIMENSUMO{} concept is more general than the semantics of the synset. For example, the semantics of the verb synset \synset{machine}{1}{v} is {\it ``Turn, shape, mold, or otherwise finish by machinery''}. Hence, the mapping of \synset{machine}{1}{v} to \subsumptionMapping{\SUMOClass{Making}} is classified as only correct. In the third part (Solutions, six columns), we provide the number of solutions classified according to two different criteria:
\begin{itemize}
\item In the first two columns (TT and FT columns), we provide the number of truth-and falsity-tests that are proved.
\item In the next two columns, we provide the number CQs where the mapping is correct ---both only correct or correct and precise--- (CM column) and incorrect (IM column) from the truth-and falsity-tests that are proved.
\item In the last two columns (CK and IK columns), we provide the number of truth-and falsity-tests that are proved on the basis of knowledge that is classified as correct (CK column) and incorrect (IK columns).
\end{itemize}
Finally, in the last part (Missing solutions, two columns) we sum up the results of our analysis of unsolved problems with a correct mapping ---either {\it only correct} or {\it correct and precise}---: the number of problems that cannot be solved because of lack of knowledge in \ADIMENSUMO{} or due to a misalignment in the knowledge of \WORDNET{} and \ADIMENSUMO{} (Knowledge column), and the number of problems that are entailed by the ontology although the ATPs do not find a proof within the given resource limits (ATP column).

As reported at the bottom of Table \ref{table:CQEvaluation}, the synsets in 43 problems are decided to be correctly connected to \ADIMENSUMO{} and, among them, the synsets in 17 problems are decided to be precisely connected to \ADIMENSUMO{} (Correct column). Thus, some of the synsets are not correctly connected to \ADIMENSUMO{} in 32 problems (Incorrect column). In total, 33 problems are solved: 26 problems are classified as entailed (TT column) and 7 problems are classified as incompatible (FT column). The knowledge of the ontology that is used in the proof of those 33 solved problems is decided to be correct (CK column) and, among them, the synsets in 18 problems are decided to be correctly connected (CM column). Consequently, from the 43 problems where the synsets are decided to be correctly connected, there are 25 unsolved problems and \ADIMENSUMO{} lacks sufficient knowledge in 22 problems (Knowledge column). Thus, 3 problems that are entailed by \ADIMENSUMO{} remain unsolved (ATP column). With respect to the 32 problems where some of the synsets are not correctly connected to \ADIMENSUMO{}, 15 problems are solved (IM column) and the remaining unsolved problems (17) have not been analyzed since the resulting conjectures are senseless.

Next, we summarize the main conclusions drawn from our detailed analysis:
\begin{itemize}
\item More than two thirds of the problems with an incorrect mapping (20 of 32 problems) belong to the Antonym category, especially to the third Antonym subcategory (19 problems). This is mainly due to the poor mapping of \WORDNET{} adjectives and satellites. More concretely, many \WORDNET{} adjectives and satellites are connected to \SUMO{} processes instead of \SUMO{} attributes.
\item Among the problems with a correct mapping, the number of problems with a precise mapping is very low (17 of 43 problems). However, this is not surprising due to the large difference between the number of concepts defined in \ADIMENSUMO{} (3,407 concepts) and \WORDNET{} (117,659 synsets).
\item Our evaluation results (i.e. number of solved problems) are penalized by the poor mapping of \WORDNET{} adjectives and satellites, especially in the third Antonym subcategory: 62.5\% of problems with a correct mapping are solved (5 of 8 problems) against 47.37\% of problems with an incorrect mapping (9 of 19 problems).
\item The solutions of all the problems that have been solved (33 problems) are based on correct knowledge of the ontology (CK column), for problems with both a correct and incorrect mapping. This means that we have not discovered incorrect knowledge in the ontology by inspecting the proofs provided by the ATPs. 
\item Most of the unsolved problems with a correct mapping (22 of 25 problems) are due to the lack of information in the ontology. However, we have also discovered 3 problems for which either the truth- or the falsity-test are entailed by \ADIMENSUMO{} although it cannot be proved by ATPs within the given resources of time and memory.
\end{itemize}

\section{Conclusions and Future Work} \label{section:conclusions}

Artificial Intelligence aims to provide computer programs with commonsense knowledge to reason about our world \cite{mccarthy1986applications}. This work offers a new practical approach towards automated commonsense reasoning with \SUMO{}-based first-order logic (FOL) ontologies. Next, we review the main contributions and results reported in this paper and discuss future work.

First, we have introduced a novel black-box testing methodology for FOL ontologies ---which is an evolved version of the methodology introduced in \cite{ALR15}--- that exploits \WORDNET{} and its mapping into \SUMO{}. For this purpose, we have considered different interpretations of the mapping and selected the most productive option for our purposes. By following our proposal, we have obtained more than 7,500 problems (thus, more than 15,000 CQs). To the best of our knowledge, this is the largest set of problems proposed for \SUMO{}-based ontologies. Secondly, we have experimentally evaluated the competency of various translations of \SUMO{} into FOL ontologies ---\TPTPSUMO{}, \ADIMENSUMO{} v.2.2 and \ADIMENSUMO{} v2.6---, the mapping between \SUMO{} and \WORDNET{}, and the efficiency of several FOL ATPs. In our experimentation, we have checked the coverage of our set of problems by analyzing the axioms that are used in the proofs provided by the ATPs. Additionally, we have also demonstrated that the proposed set of problems enables the evaluation of different features of the ATPs since each system is able to solve a different subset of problems using the same time and memory resources. Finally, we have manually evaluated the quality of a subset of the proposed problems when testing \ADIMENSUMO{} v2.6. From our manual evaluation, we have detected a) some defects in the mapping of synsets (especially in the case of adjectives) and b) some solvable problems for which ATPs find no solution. We plan to propose the inclusion of our set of problems in the CSR domain of the TPTP problem library and in the set of eligible problems for the LTB division of CASC.

All the resources that have been used and developed during this work are available in a single package, including:\footnote{The package is available at \url{http://adimen.si.ehu.es/web/AdimenSUMO}.} a) the ontologies; b) tools for the creation of tests, its experimentation and the analysis of results; and c) the resulting tests for each ontology and the output obtained from different ATPs.

Regarding future work, our plan is to enlarge the proposed set of problems by following different strategies. Amongst others:
\begin{itemize}
\item By considering alternative proposals for the translation of the \WORDNET{}-\SUMO{} mapping.
\item By exploiting additional \SUMO{} relations, such as meronymy, hyponymy, etc. Some preliminary works have been already introduced in \cite{AlR18,AGR18}.
\item By exploiting other resources of knowledge such as \TCO{} \cite{AAC08}, \FRAMENET{} \cite{REP06}, \PREDICATEMATRIX{} \cite{LLA16}, \CONCEPTNET{} \cite{SpH12} or \VISUALGENOME{} \cite{KZG16}.
\item By following white-box testing strategies that focus on the particular representation of the knowledge \cite{AHL17}.
\end{itemize}
Furthermore, we also aim to exploit unsolved problems in order to improve \ADIMENSUMO{}. For this purpose, we will have to analyze whether the classification of problems as unsolved is due to the lack of knowledge in \ADIMENSUMO{}. If so, we would consider the possibility of enriching \ADIMENSUMO{} by adding knowledge from \WORDNET{} or other resources. Additionally, \WORDNET{} itself and its mapping can be evaluated. For example, by detecting synsets that are frequently involved in problems classified as incompatible. Finally, we plan to evaluate the knowledge in the Multilingual Central Repository (MCR) \cite{GLR12} and to check the utility of \ADIMENSUMO{} v2.6 in Natural Language Processing (NLP) tasks that involve reasoning on commonsense knowledge \cite{Bos09}, such as Recognizing Textual Entailment (RTE) \cite{BoM06,DRS13,Abz17}, Natural Language Inference (NLI) \cite{BAP15} or Interpretable Semantic Textual Similarity (ISTS) \cite{LMG17}.

\section*{Acknowledgements}

We thank the anonymous reviewers for their valuable comments and suggestions.

This work has been partially funded by the Spanish Projects TUNER (TIN2015-65308-C5-1-R) and COMMAS (TIN2013-46181-C2-2-R), the Basque Project LoRea (GIU15/30) and grant BAILab (UFI11/45).

\section*{Bibliography}

\bibliographystyle{abbrv} 
\bibliography{black}

\end{document}

%% file: commands.tex
\newcommand{\function}[2]{#1_{/#2}}


\newcommand{\tab}{\hspace{20pt}}
\newcommand{\doubletab}{\tab\tab}
\newcommand{\connective}[1]{\bf #1 \;}
\newcommand{\predicate}[1]{\rm #1}
\newcommand{\constant}[1]{\rm #1}
\newcommand{\variable}[1]{\tt ?#1}
\newcommand{\rowVariable}[1]{\tt @#1}
\newcommand{\true}{\it true}
\newcommand{\false}{\it false}

\newcommand{\TT}[1]{#1}
\newcommand{\FT}[1]{\neg #1}


\newcommand{\textVariable}[1]{{\it{?#1}}}
\newcommand{\textConstant}[1]{{\it{#1}}}
\newcommand{\textFunction}[1]{{\it{#1}}}
\newcommand{\textPredicate}[1]{{\it{#1}}}
\newcommand{\metaConstant}[1]{{\it{\dollar #1}}}
\newcommand{\metaPredicate}[1]{{\dollar #1}}
\newcommand{\dollar}{\$}


\newcommand{\SUMOObjectSymbol}{o}
\newcommand{\SUMOClassSymbol}{c}
\newcommand{\SUMOIndividualRelationSymbol}{r}
\newcommand{\SUMOClassOfRelationsSymbol}{R}
\newcommand{\SUMOIndividualAttributeSymbol}{a}
\newcommand{\SUMOClassOfAttributesSymbol}{A}

\newcommand{\SUMOObject}[1]{{\textConstant{#1}$_\SUMOObjectSymbol$}}
\newcommand{\SUMOClass}[1]{{\textConstant{#1}$_\SUMOClassSymbol$}}
\newcommand{\SUMOIndividualRelation}[1]{{\textConstant{#1}$_\SUMOIndividualRelationSymbol$}}
\newcommand{\HoldsIndividualRelation}[2]{{\textConstant{#1}$^{#2}_{\SUMOIndividualRelationSymbol}$}}
\newcommand{\SUMOClassOfRelations}[1]{{\textConstant{#1}$_\SUMOClassOfRelationsSymbol$}}
\newcommand{\SUMOIndividualAttribute}[1]{{\textConstant{#1}$_\SUMOIndividualAttributeSymbol$}}
\newcommand{\SUMOClassOfAttributes}[1]{{\textConstant{#1}$_\SUMOClassOfAttributesSymbol$}}

\newcommand{\SUMOObjectTikZ}[1]{{\constant{#1}_\SUMOObjectSymbol}}
\newcommand{\SUMOClassTikZ}[1]{{\constant{#1}_\SUMOClassSymbol}}
\newcommand{\SUMOIndividualRelationTikZ}[1]{{\constant{\it #1}_\SUMOIndividualRelationSymbol}}
\newcommand{\SUMOClassOfRelationsTikZ}[1]{{\constant{\it #1}_\SUMOClassofRelationsSymbol}}
\newcommand{\SUMOIndividualAttributeTikZ}[1]{{\constant{#1}_\SUMOIndividualAttributeSymbol}}
\newcommand{\SUMOClassOfAttributesTikZ}[1]{{\constant{#1}_\SUMOClassOfAttributesSymbol}}


\newcommand{\synset}[3]{{\it{#1}$_{#3}^{#2}$}}
\newcommand{\pair}[3]{{\it{#1}(#2,#3)}}

\newcommand{\equivalenceMappingSymbol}{=}
\newcommand{\subsumptionMappingSymbol}{+}
\newcommand{\instanceMappingSymbol}{@}
\newcommand{\negatedEquivalenceMappingSymbol}{\widehat{\equivalenceMappingSymbol}}
\newcommand{\negatedSubsumptionMappingSymbol}{\widehat{\subsumptionMappingSymbol}}

\newcommand{\equivalenceMapping}[1]{{\textConstant{#1}$\equivalenceMappingSymbol$}}
\newcommand{\subsumptionMapping}[1]{{\textConstant{#1}$\subsumptionMappingSymbol$}}
\newcommand{\instanceMapping}[1]{{\textConstant{#1}$\instanceMappingSymbol$}}
\newcommand{\negatedEquivalenceMapping}[1]{{\textConstant{#1}$\negatedEquivalenceMappingSymbol$}}
\newcommand{\negatedSubsumptionMapping}[1]{{\textConstant{#1}$\negatedSubsumptionMappingSymbol$}}

\newcommand{\equivalenceMappingOfConcept}[1]{{#1}$\equivalenceMappingSymbol$}
\newcommand{\subsumptionMappingOfConcept}[1]{{#1}$\subsumptionMappingSymbol$}
\newcommand{\instanceMappingOfConcept}[1]{{#1}$\instanceMappingSymbol$}
\newcommand{\negatedEquivalenceMappingOfConcept}[1]{{#1}$\negatedEquivalenceMappingSymbol$}
\newcommand{\negatedSubsumptionMappingOfConcept}[1]{{#1}$\negatedSubsumptionMappingSymbol$}

\newcommand{\equivalenceMappingRelation}{{\it{equivalence}}}
\newcommand{\subsumptionMappingRelation}{{\it{subsumption}}}
\newcommand{\instanceMappingRelation}{{\it{instance}}}

\newcommand{\synsetTikZ}[3]{{\it{#1}_{#3}^{#2}}}

\newcommand{\equivalenceMappingTikZ}[1]{{{\constant{#1}}\hspace{-4pt}\equivalenceMappingSymbol}}
\newcommand{\subsumptionMappingTikZ}[1]{{{\constant{#1}}\subsumptionMappingSymbol}}
\newcommand{\instanceMappingTikZ}[1]{{{\constant{#1}}\instanceMappingSymbol}}
\newcommand{\negatedEquivalenceMappingTikZ}[1]{{{\constant{#1}}\negatedEquivalenceMappingSymbol}}
\newcommand{\negatedSubsumptionMappingTikZ}[1]{{{\constant{#1}}\negatedSubsumptionMappingSymbol}}

\newcommand{\equivalenceMappingTikZOfConcept}[1]{{#1}\hspace{-4pt}\equivalenceMappingSymbol}
\newcommand{\subsumptionMappingTikZOfConcept}[1]{{#1}\subsumptionMappingSymbol}
\newcommand{\instanceMappingTikZOfConcept}[1]{{#1}\instanceMappingSymbol}
\newcommand{\negatedEquivalenceMappingTikZOfConcept}[1]{{#1}\negatedEquivalenceMappingSymbol}
\newcommand{\negatedSubsumptionMappingTikZOfConcept}[1]{{#1}\negatedSubsumptionMappingSymbol}


\newcommand{\WORDNET}{{WordNet}}
\newcommand{\SUMO}{SUMO}
\newcommand{\Cyc}{Cyc}
\newcommand{\DOLCE}{DOLCE}
\newcommand{\YAGO}{YAGO}
\newcommand{\TPTPSUMO}{TPTP-SUMO}
\newcommand{\ADIMENSUMO}{Adimen-SUMO}
\newcommand{\adimensumo}{Adimen-SUMO}
\newcommand{\FRAMENET}{FrameNet}
\newcommand{\PREDICATEMATRIX}{Predicate Matrix}
\newcommand{\TCO}{EuroWordNet Top Ontology}
\newcommand{\CONCEPTNET}{ConceptNet}\newcommand{\VISUALGENOME}{VisualGenome}

%% file: black.revision01.bbl
\begin{thebibliography}{10}

\bibitem{Abz17}
L.~Abzianidze.
\newblock {LangPro}: Natural language theorem prover.
\newblock In L.~Specia, M.~Post, and M.~Paul, editors, {\em {Proc. of the 2017
  Conference on Empirical Methods in Natural Language Processing: System
  Demonstrations (EMNLP 2017)}}, pages 115--120. Association for Computational
  Linguistics, 2017.

\bibitem{AAC08}
J.~{\'A}lvez, J.~Atserias, J.~Carrera, S.~Climent, E.~Laparra, A.~Oliver, and
  G.~Rigau.
\newblock Complete and consistent annotation of {WordNet} using the {Top
  Concept Ontology}.
\newblock In N.~Calzolari, K.~Choukri, B.~Maegaard, J.~Mariani, J.~Odjik,
  S.~Piperidis, and D.~Tapias, editors, {\em {Proc. of the 6$^{th}$ Int. Conf.
  on Language Resources and Evaluation (LREC 2008)}}, pages 1529--1534.
  European Language Resources Association (ELRA), may 2008.

\bibitem{AGR18}
J.~{\'Alvez}, I.~Gonzalez{-}Dios, and G.~Rigau.
\newblock Cross-checking {\WORDNET{}} and {\SUMO{}} using meronymy.
\newblock In N.~Calzolari, editor, {\em Proc. of {the 11$^{th}$ Int. Conf. on
  Language Resources and Evaluation (LREC 2018)}}. European Language Resources
  Association (ELRA), 2018.

\bibitem{AHL17}
J.~{\'{A}}lvez, M.~Hermo, P.~Lucio, and G.~Rigau.
\newblock Automatic white-box testing of first-order logic ontologies.
\newblock {\em CoRR}, abs/1705.10219, 2017.

\bibitem{ALR12}
J.~{\'A}lvez, P.~Lucio, and G.~Rigau.
\newblock {\ADIMENSUMO{}}: Reengineering an ontology for first-order reasoning.
\newblock {\em Int. J. Semantic Web Inf. Syst.}, 8(4):80--116, 2012.

\bibitem{ALR15}
J.~{\'A}lvez, P.~Lucio, and G.~Rigau.
\newblock Improving the competency of first-order ontologies.
\newblock In K.~Barker and J.~M. G\'{o}mez{-}P\'{e}rez, editors, {\em Proc. of
  the 8$^{th}$ {Int. Conf. on Knowledge Capture (K-CAP 2015)}}, pages
  15:1--15:8. ACM, 2015.

\bibitem{ALR16}
J.~{\'A}lvez, P.~Lucio, and G.~Rigau.
\newblock Evaluating automated theorem provers using {\ADIMENSUMO{}}.
\newblock In L.~Kov\'{a}cs and A.~Voronkov, editors, {\em Proc. of the 3$^{rd}$
  {Vampire Workshop (Vampire 2016)}}, volume~44 of {\em EPiC Series in
  Computing}, pages 74--82. EasyChair, 2017.

\bibitem{AlR18}
J.~{\'A}lvez and G.~Rigau.
\newblock Towards cross-checking {\WORDNET{}} and {\SUMO{}} using meronymy.
\newblock In P.~Vossen, C.~Fellbaum, and F.~Bond, editors, {\em Proc. of the
  9$^{th}$ {Global WordNet Conference (GWC 2018)}}, 2018.

\bibitem{BlB05}
P.~Blackburn and J.~Bos.
\newblock {\em Representation and Inference for Natural Language. {A} First
  Course in Computational Semantics}.
\newblock {CSLI} Studies in Computational Linguistics. {CSLI} Publications,
  2005.

\bibitem{BBK01}
P.~Blackburn, J.~Bos, M.~Kohlhase, and H.~D. Nivelle.
\newblock {\em Inference and Computational Semantics}, pages 11--28.
\newblock Springer Netherlands, Dordrecht, 2001.

\bibitem{Bos09}
J.~Bos.
\newblock Applying automated deduction to natural language understanding.
\newblock {\em J. of Applied Logic}, 7(1):100--112, 2009.
\newblock Special Issue: Empirically Successful Computerized Reasoning.

\bibitem{BoM06}
J.~Bos and K.~Markert.
\newblock Recognising textual entailment with robust logical inference.
\newblock In J.~Qui{\~{n}}onero-Candela, I.~Dagan, B.~Magnini, and
  F.~d'Alch{\'e} Buc, editors, {\em Machine Learning Challenges. Evaluating
  Predictive Uncertainty, Visual Object Classification, and Recognising Tectual
  Entailment}, pages 404--426, Berlin, Heidelberg, 2006. Springer Berlin
  Heidelberg.

\bibitem{BAP15}
S.~R. Bowman, G.~Angeli, C.~Potts, and C.~D. Manning.
\newblock A large annotated corpus for learning natural language inference.
\newblock In {\em Proc. of the {Conf. on Empirical Methods in Natural Language
  Processing (EMNLP 2015)}}. Association for Computational Linguistics, 2015.

\bibitem{DRS13}
I.~Dagan, D.~Roth, M.~Sammons, and F.~M. Zanzotto.
\newblock Recognizing textual entailment: Models and applications.
\newblock {\em Synthesis Lectures on Human Language Technologies}, 6(4):1--220,
  2013.

\bibitem{DaM15}
E.~Davis and G.~Marcus.
\newblock Commonsense reasoning and commonsense knowledge in artificial
  intelligence.
\newblock {\em Commun. ACM}, 58(9):92--103, aug 2015.

\bibitem{LLA16}
M.~L. de~Lacalle, E.~Laparra, I.~Aldabe, and G.~Rigau.
\newblock Predicate matrix: automatically extending the semantic
  interoperability between predicate resources.
\newblock {\em Language Resources and Evaluation}, 50(2):263--289, 2016.

\bibitem{Fellbaum'98}
C.~Fellbaum, editor.
\newblock {\em {WordNet:} {A}n Electronic Lexical Database}.
\newblock MIT Press, 1998.

\bibitem{FOC09}
C.~Fellbaum, A.~Osherson, and P.~Clark.
\newblock Putting semantics into {WordNet}'s ``{M}orphosemantic'' {L}inks.
\newblock In Z.~Vetulani and H.~Uszkoreit, editors, {\em Human Language
  Technology. Challenges of the Information Society}, LNCS 5603, pages
  350--358. Springer, 2009.

\bibitem{FGS13}
M.~Fern{\'a}ndez-L{\'o}pez, A.~G{\'o}mez-P{\'e}rez, and M.~C.
  Su{\'a}rez-Figueroa.
\newblock Methodological guidelines for reusing general ontologies.
\newblock {\em Data \& Knowledge Engineering}, 86:242--275, 2013.

\bibitem{GCC06}
A.~Gangemi, C.~Catenacci, M.~Ciaramita, and J.~Lehmann.
\newblock Modelling ontology evaluation and validation.
\newblock In Y.~Sure and J.~Domingue, editors, {\em The Semantic Web: Research
  and Applications}, LNCS 4011, pages 140--154. Springer, 2006.

\bibitem{Richard+'92}
M.~R. Genesereth, R.~E. Fikes, D.~Brobow, R.~Brachman, T.~Gruber, P.~Hayes,
  R.~Letsinger, V.~Lifschitz, R.~Macgregor, J.~Mccarthy, P.~Norvig, R.~Patil,
  and L.~Schubert.
\newblock {Knowledge Interchange Format} version 3.0 reference manual.
\newblock Technical Report Logic-92-1, Stanford University, Computer Science
  Department, Logic Group, 1992.

\bibitem{GFC04}
A.~G{\'o}mez-P{\'e}rez, M.~Fern{\'a}ndez-L{\'o}pez, and O.~Corcho-Garc{\'i}a.
\newblock Ontological engineering.
\newblock {\em Computing Reviews}, 45(8):478--479, 2004.

\bibitem{GLR12}
A.~Gonzalez{-}Agirre, E.~Laparra, and G.~Rigau.
\newblock {Multilingual Central Repository} version 3.0.
\newblock In N.~Calzolari, K.~Choukri, T.~Declerck, M.~U. Dogan, B.~Maegaard,
  J.~Mariani, J.Odijk, and S.~Piperidis, editors, {\em Proc. of {the 8$^{th}$
  Int. Conf. on Language Resources and Evaluation (LREC 2012)}}, pages
  2525--2529. European Language Resources Association (ELRA), 2012.

\bibitem{Gru09}
T.~Gruber.
\newblock Ontology.
\newblock In L.~Liu and M.~T. {\"O}zsu, editors, {\em Encyclopedia of Database
  Systems}, pages 1963--1965. Springer US, 2009.

\bibitem{GrF95}
M.~Gr{\"u}ninger and M.~S. Fox.
\newblock Methodology for the design and evaluation of ontologies.
\newblock In {\em Proc. of {the Workshop on Basic Ontological Issues in
  Knowledge Sharing (IJCAI 1995)}}, 1995.

\bibitem{HoV06}
I.~Horrocks and A.~Voronkov.
\newblock Reasoning support for expressive ontology languages using a theorem
  prover.
\newblock In {J. Dix et al.}, editor, {\em Foundations of Information and
  Knowledge Systems}, LNCS 3861, pages 201--218. Springer, 2006.

\bibitem{KoV13}
L.~Kov\'{a}cs and A.~Voronkov.
\newblock First-order theorem proving and {Vampire}.
\newblock In N.~Sharygina and H.~Veith, editors, {\em {Computer Aided
  Verification}}, LNCS 8044, pages 1--35. Springer, 2013.

\bibitem{KZG16}
R.~Krishna, Y.~Zhu, O.~Groth, J.~Johnson, K.~Hata, J.~Kravitz, S.~Chen,
  Y.~Kalantidis, L.~J. Li, D.~A. Shamma, M.~Bernstein, and L.~Fei-Fei.
\newblock {Visual Genome}: Connecting language and vision using crowdsourced
  dense image annotations.
\newblock {\em CoRR}, abs/1602.07332, 2016.

\bibitem{LMG17}
I.~Lopez-Gazpio, M.~Maritxalar, A.~Gonzalez-Agirre, G.~Rigau, L.~Uria, and
  E.~Agirre.
\newblock Interpretable semantic textual similarity: Finding and explaining
  differences between sentences.
\newblock {\em Knowledge-Based Systems}, 119:186 -- 199, 2017.

\bibitem{mccarthy1986applications}
J.~McCarthy.
\newblock Applications of circumscription to formalizing common-sense
  knowledge.
\newblock {\em Artificial Intelligence}, 28(1):89--116, 1986.

\bibitem{mccarthy1989artificial}
J.~McCarthy.
\newblock Artificial intelligence, logic and formalizing common sense.
\newblock In R.~H. Thomason, editor, {\em Philosophical Logic and Artificial
  Intelligence}, pages 161--190. Springer, 1989.

\bibitem{minsky2007emotion}
M.~Minsky.
\newblock {\em The Emotion Machine: Commonsense Thinking, Artificial
  Intelligence, and the Future of the Human Mind}.
\newblock Simon \& Schuster, 2007.

\bibitem{MSB12}
G.~J. Myers, C.~Sandler, and T.~Badgett.
\newblock {\em The art of software testing}.
\newblock John Wiley \& Sons, Inc., Hoboken, N.J., 2012.

\bibitem{Niles+Pease'01}
I.~Niles and A.~Pease.
\newblock Towards a standard upper ontology.
\newblock In {Guarino N. et al.}, editor, {\em Proc. of the 2$^{nd}$ {Int.
  Conf. on Formal Ontology in Information Systems (FOIS 2001)}}, pages 2--9.
  ACM, 2001.

\bibitem{Niles+Pease'03}
I.~Niles and A.~Pease.
\newblock Linking lexicons and ontologies: Mapping {WordNet} to the {Suggested
  Upper Merged Ontology}.
\newblock In H.~R. Arabnia, editor, {\em Proc. of the {IEEE Int. Conf. on Inf.
  and Knowledge Engin. (IKE 2003)}}, volume~2, pages 412--416. CSREA Press,
  2003.

\bibitem{noy2001ontology}
N.~F. Noy and D.~L. McGuinness.
\newblock Ontology development 101: A guide to creating your first ontology.
\newblock Technical Report KSL-01-05 and SMI-2001-0880, Stanford Knowledge
  Systems Laboratory and Stanford Medical Informatics, 2001.

\bibitem{Pea09}
A.~Pease.
\newblock {Standard Upper Ontology Knowledge Interchange Format}.
\newblock Retrieved June 18, 2009, from
  \url{http://sigmakee.cvs.sourceforge.net/sigmakee/sigma/suo-kif.pdf}, 2009.

\bibitem{PeB13}
A.~Pease and C.~Benzm{\"u}ller.
\newblock Sigma: An integrated development environment for formal ontology.
\newblock {\em AI Communications (Special Issue on Intelligent Engineering
  Techniques for Knowledge Bases)}, 26(1):79--97, 2013.

\bibitem{PeS07}
A.~Pease and G.~Sutcliffe.
\newblock First-order reasoning on a large ontology.
\newblock In {Sutcliffe G. et al.}, editor, {\em Proc. of the {Workshop on
  Empirically Successful Automated Reasoning in Large Theories (CADE-21)}},
  CEUR Workshop Proceedings 257. CEUR-WS.org, 2007.

\bibitem{PSS02}
F.~Pelletier, G.~Sutcliffe, and C.~Suttner.
\newblock {The Development of CASC}.
\newblock {\em AI Communications}, 15(2-3):79--90, 2002.

\bibitem{RiV02}
A.~Riazanov and A.~Voronkov.
\newblock The design and implementation of {V}ampire.
\newblock {\em AI Communications}, 15(2-3):91--110, 2002.

\bibitem{REP06}
J.~Ruppenhofer, M.~Ellsworth, M.~R.~L. Petruck, C.~R. Johnson, and
  J.~Scheffczyk.
\newblock {\em {FrameNet} II: Extended Theory and Practice}.
\newblock International Computer Science Institute, Berkeley, California, 2006.
\newblock Distributed with the FrameNet data.

\bibitem{Sch02}
S.~Schulz.
\newblock {E - A} brainiac theorem prover.
\newblock {\em AI Communications}, 15(2-3):111--126, 2002.

\bibitem{SpH12}
R.~Speer and C.~Havasi.
\newblock Representing general relational knowledge in {\CONCEPTNET{}} 5.
\newblock In N.~Calzolari, K.~Choukri, T.~Declerck, M.~U. Dogan, B.~Maegaard,
  J.~Mariani, J.Odijk, and S.~Piperidis, editors, {\em Proc. of {the 8$^{th}$
  Int. Conf. on Language Resources and Evaluation (LREC 2012)}}, pages
  3679--3686. European Language Resources Association (ELRA), 2012.

\bibitem{StS09}
S.~Staab and R.~Studer.
\newblock {\em Handbook on Ontologies}.
\newblock Springer Publishing Company, Incorporated, 2$^{nd}$ edition, 2009.

\bibitem{Sut09}
G.~Sutcliffe.
\newblock The {TPTP} problem library and associated infrastructure.
\newblock {\em J. Automated Reasoning}, 43(4):337--362, 2009.

\bibitem{SuS01}
G.~Sutcliffe and C.~Suttner.
\newblock Evaluating general purpose automated theorem proving systems.
\newblock {\em Artificial Intelligence}, 131(1):39 -- 54, 2001.

\bibitem{SuS06}
G.~Sutcliffe and C.~Suttner.
\newblock {The State of CASC}.
\newblock {\em AI Communications}, 19(1):35--48, 2006.

\end{thebibliography}
